\documentclass[letterpaper]{article} %
\usepackage{aaai2026}  %
\usepackage{times}  %
\usepackage{helvet}  %
\usepackage{courier}  %
\usepackage[hyphens]{url}  %
\usepackage{graphicx} %
\usepackage{natbib}  %
\usepackage{caption} %
\usepackage{algorithm}
\usepackage{algorithmic}

\usepackage{newfloat}
\usepackage{listings}
\DeclareCaptionStyle{ruled}{labelfont=normalfont,labelsep=colon,strut=off} %
\floatstyle{ruled}
\newfloat{listing}{tb}{lst}{}
\floatname{listing}{Listing}
\usepackage{booktabs}
\usepackage{multirow}
\usepackage{amsmath}
\usepackage{enumitem}
\usepackage{tcolorbox}

\usepackage{comment}

\newif\ifshowappendix
\showappendixtrue

\usepackage[capitalize]{cleveref}
\crefname{section}{Sec.}{Secs.}
\Crefname{section}{Section}{Sections}
\Crefname{table*}{table*}{Tables}
\crefname{table*}{Tab.}{Tabs.}
\Crefname{figure*}{figure*}{Figures}
\crefname{figure*}{Fig.}{Figs.}

\newcommand{\objbox}[1]{\textcolor{blue!80!black}{{#1}}}
\newcommand{\atrbox}[1]{\textcolor{purple!85!black}{{#1}}}
\newcommand{\relbox}[1]{\textcolor{green!60!black}{{#1}}}

\title{Multigranular Evaluation for Brain Visual Decoding}
\author{
    Weihao Xia\thanks{Corresponding author: wx258@cam.ac.uk}, Cengiz Oztireli
}
\affiliations{
    University of Cambridge %
}

\begin{document}

\maketitle

\begin{abstract}
Existing evaluation protocols for brain visual decoding predominantly rely on coarse metrics that obscure inter-model differences, lack neuroscientific foundation, and fail to capture fine-grained visual distinctions. To address these limitations, we introduce BASIC, a unified, multigranular evaluation framework that jointly quantifies structural fidelity, inferential alignment, and contextual coherence between decoded and ground-truth images. For the structural level, we introduce a hierarchical suite of segmentation-based metrics, including foreground, semantic, instance, and component masks, anchored in granularity-aware correspondence across mask structures. For the semantic level, we extract structured scene representations encompassing objects, attributes, and relationships using multimodal large language models, enabling detailed, scalable, and context-rich comparisons with ground-truth stimuli. We benchmark a diverse set of visual decoding methods across multiple stimulus-neuroimaging datasets within this unified evaluation framework. Together, these criteria provide a more discriminative, interpretable, and comprehensive foundation for evaluating brain visual decoding methods.
\end{abstract}

\begin{links}
    \link{Code}{https://github.com/weihaox/BASIC}
\end{links}

\section{Introduction}

Recent advances in brain visual decoding~\cite{takagi2023high,ozcelik2023brain,scotti2023reconstructing,xia2024umbrae} have achieved remarkable success in reconstructing visual stimuli from the neural activations. However, the evaluation protocols commonly used in this field remain limited in several critical aspects. 
First, current metrics often saturate across state-of-the-art models, limiting their discriminative capacity and obscuring substantive differences in decoded results.
Second, these metrics often lack a neuroscientific foundation, failing to capture the perceptual validity of decoded outputs and their alignment with human-like perception.
Third, prevailing evaluation strategies typically fail to reflect the multilevel and structured nature of visual perception, neglecting key components such as object semantics, scene understanding, and  contextual reasoning.
These limitations hinder rigorous benchmarking of brain decoding models and obscure the specific dimensions along which reconstructions succeed or fall short.

\paragraph{What should brain decoding recover.} %

Brain visual decoding aims to reconstruct visual experiences from neural activations, recovering not only the appearance of stimuli but also their structure, semantics, and perceptual salience.
A well-decoded reconstruction should reflect what the subject consciously perceived, preserving salient objects, their attributes, spatial configuration, and overall scene coherence. Since human visual perception is shaped by attention, context, and prior knowledge, brain decoding must align with the hierarchical nature of vision, spanning from low-level pixel patterns to high-level semantic understanding. Therefore, effective brain visual decoding requires both perceptual accuracy and semantic integrity. It should faithfully capture salient elements and spatial context in line with the subject’s attention, while maintaining consistent inter-object relationships and scene-level coherence.

\paragraph{What should we measure in brain visual decoding.} %

Current evaluation protocols, such as pixel-wise correlation or feature-based similarity, often fall short in capturing the full complexity of brain visual decoding. Low-level metrics tend to overlook scene semantics and perceptual plausibility, while high-level black-box measures conflate multiple alignment aspects into a single score, offering limited diagnostic insight. 
They struggle to determine whether the ``decoded'' details truly originates from brain signals, or if they are instead hallucinated constructs based on prototypical co-occurrences from pretrained generative models conditioned on scene or object labels.
Moreover, existing metrics are often \textit{saturated}, assigning uniformly high scores across diverse methods and thus failing to capture fine-grained distinctions in reconstruction quality.
We argue that an effective brain visual decoding metric should be neuroscientifically interpretable, grounded in principles of human visual perception, and meet three key desiderata. First, it should be \textit{multigranular}, capturing perceptual alignment across multiple abstraction levels -- from basic object identification and segmentation to semantic and spatial reasoning about attributes and interactions. Second, it should be \textit{semantically aligned} with human perception, reflecting the way humans interpret scenes, objects, and their relationships in a coherent and meaningful manner. Third, it should be \textit{diagnostically informative}, offering interpretable feedback on what is correct, what is missing, and where the reconstruction fails. Specifically, it should localize and characterize semantic and structural errors in decoded outputs, such as object misidentification, incorrect attributes, or implausible interactions.

\paragraph{Our method: BASIC.}  

To bridge these gaps, we introduce BASIC (Brain-Aligned Structural, Inferential, and Contextual similarity), a unified, multigranular evaluation framework for brain visual decoding. 
BASIC integrates structure matching and semantic reasoning to systematically quantify structural, inferential, and contextual alignment between decoded results and reference stimuli across diverse brain-to-vision tasks in a structured and interpretable manner.

BASIC decomposes evaluation into three complementary perspectives, each reflecting a core aspect of perceptual alignment in brain visual decoding:
(a) \textit{structural similarity} quantifies the reconstructed visual structures, capturing spatial organization and categorical boundaries. This consistency with reference stimuli is operationalized through a granularity-aware mask correspondence across the foreground, semantic, instance, and component levels;
(b) \textit{inferential similarity} measures semantic accuracy, evaluating whether the decoded image conveys the same entities and conceptual content as the reference. This is computed via structured comparisons across object categories, attributes, and inter-object relational graphs extracted using captions from multimodal large language models;
(c) \textit{contextual similarity} assesses perceptual and cognitive plausibility, examining whether the reconstructed scene forms an internally coherent and contextually appropriate whole. This is evaluated using MLLM-based scene reasoning to quantify narrative consistency and global scene coherence.

BASIC offers a comprehensive view of decoding performance across modalities (image, video, 3D) and neuroimaging types (fMRI, EEG). Our framework facilitates both quantitative comparisons and qualitative diagnostics, allowing for fine-grained benchmarking of brain decoding models across datasets.
BASIC aims to (1) offer a more detailed and interpretable evaluation of brain decoding results, (2) quantify semantic plausibility in terms more closely aligned with human cognitive processes, and (3) facilitate comparison of decoding methods within a unified evaluation framework applicable across diverse stimulus-neuroimaging datasets. We hope our method contributes to establishing a more systematic foundation for brain visual decoding evaluation.

\begin{figure*}[t!]
    \centering
    \includegraphics[width=0.98\linewidth]{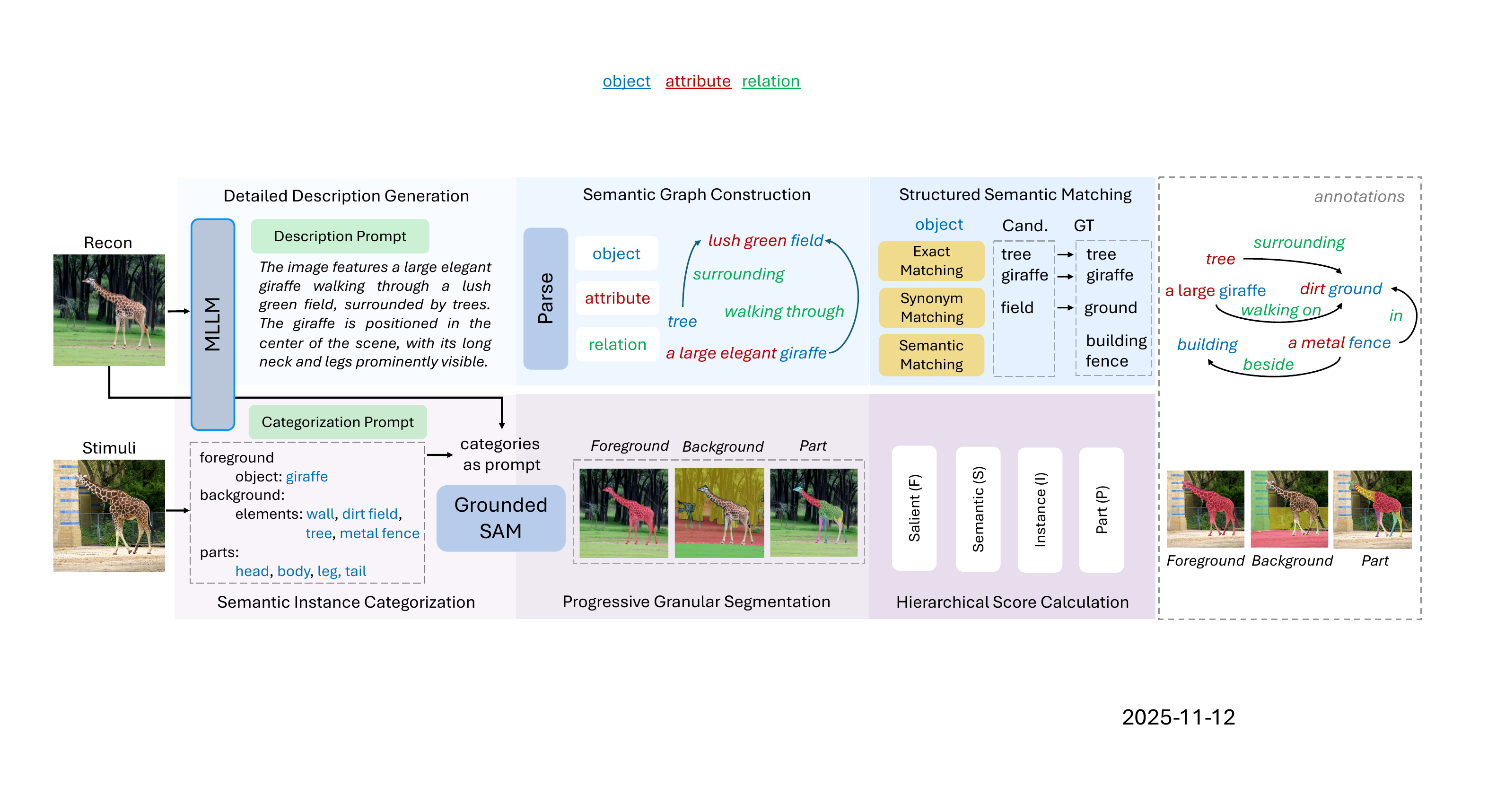}
    \vspace{-5pt}
    \caption{BASIC evaluates decoded reconstructions along two axes: high-level semantic (BASIC-H) and low-level structural (BASIC-L) similarities. For the semantic axis (inferential and contextual), we extract and compare structured representations from reconstructed and ground-truth images. For the structural axis, we compute mask-based matching across fine-grained segmentation types of identified scenes and objects: salient, semantic, instance, and parts.}
    \label{fig:overview}
\end{figure*}

\section{Related Work}
\label{sec:related_work}

\paragraph{Brain visual decoding.} The task of brain visual decoding aims to reconstruct perceived visual stimuli, such as images, videos, or 3D shapes, from recorded neural activations. Recent progress in this domain has been closely tied to advancements in computational modeling frameworks. Early methods primarily involved training neural networks from scratch to learn mappings between brain activity and visual features. However, these models often suffered from limited fidelity and exhibited artifacts and poor visual quality in the reconstructed outputs. Recent developments have led to significant improvements due to the emergence of multimodal generative models~\cite{radford2021learning,rombach2022high,xu2022versatile} and large-scale brain-stimulus datasets~\cite{wen2018neural,allen2022massive,gao2024mind3d,guo2025neuro3d}. 
These resources have facilitated a new generation of decoding paradigms that leverage intermediate representations from pretrained generative models and align them with neural responses through various strategies, including linear regression~\cite{ozcelik2023brain,takagi2023high}, diffusion priors~\cite{scotti2023reconstructing,scotti2024mindeye2}, or feature-wise reconstruction~\cite{xia2024umbrae,xia2025mevox,xia2025vindex}. Recent efforts have also focused on eliminating dependence on subject-specific encoders~\cite{scotti2024mindeye2,wang2024mindbridge,xia2024umbrae,tian2025brainguard,gong2025mindtuner}, demonstrating promising progress toward subject-independent brain decoding.
Besides the commonly used fMRI-image Natural Scenes Dataset (NSD)~\cite{allen2022massive}, a growing body of research has explored alternative combinations of neuroimaging modalities and stimuli, such as fMRI-video~\cite{wen2018neural}, fMRI-3D~\cite{gao2024mind3d}, EEG-image~\cite{grootswagers2022human}, EEG-video~\cite{liu2024eeg2video}, and EEG-3D~\cite{guo2025neuro3d} decoding. These explorations broaden the scope of brain decoding and offer new insights into the neural representation of dynamic and immersive visual experiences.

\paragraph{Brain decoding evaluation.}
The evaluation metrics for visual brain decoding lack a universally accepted standard. Different stimulus-neuroimaging combinations employ varying evaluation protocols. For instance, the following eight metrics are commonly used on NSD: PixCorr, SSIM~\cite{wang2004image}, AlexNet~\cite{krizhevsky2017imagenet}-2/5, Inception~\cite{szegedy2016rethinking}, CLIP~\cite{radford2021learning}, EffNet~\cite{tan2019efficientnet}, and SwAV~\cite{caron2020unsupervised}. The first four metrics are considered low-level, focusing on perceptual similarity and pixel-wise or structural correspondence. In contrast, the latter four are used to evaluate semantic decoding or high-level representations, emphasizing the overall theme or content of the reconstructed images. For other stimulus-neuroimaging setups, evaluation protocols remain inconsistent. Metrics such as $n$-way classification accuracy~\cite{guo2025neuro3d}, CLIP-based Pearson correlation~\cite{gong2024neuroclips}, mask-matching ratios~\cite{li2025decofuse}, and modality-specific evaluation metrics~\cite{gao2024mind3d++} have all been reported in the recent literature. 
However, prior metrics overlook the hierarchical nature of perception and struggle to distinguish models with subtle differences. In contrast, our BASIC framework offers a unified, multigranular evaluation across structural, inferential, and contextual dimensions, enabling more nuanced and interpretable comparisons.

\section{BASIC: Evaluating Brain Visual Decoding}
\label{sec:basic}

BASIC provides a versatile evaluation framework for brain visual decoding across diverse combinations of visual stimuli and neuroimaging modalities. This section details
the evaluation dimensions and the two complementary modules of our BASIC metric: 
BASIC-H  integrates the inferential and contextual dimensions of BASIC into a unified measure of high-level semantic correspondence, while BASIC-L quantifies low-level structural alignment.

\subsection{Evaluation Dimension}
\label{subsec:evaluation_dimension}

To gain deeper insight into how brain decoding models represent visual semantics across different levels of abstraction, we develop a structured evaluation protocol grounded in key conceptual dimensions of human visual perception: scenes, objects, attributes, and relations.
Each dimension is further divided into subcategories as outlined in~\cref{tab:evaluation_dim}. %

\textbf{Scene} reflects global properties, including overall layout, geometric structure, event context, and stylistic tone. This probes
the holistic configuration and contextual coherence.

\textbf{Object} evaluates recognition and differentiation of objects in the scene, including both categorical accuracy (e.g., ``cat'' vs. ``dog'') and semantic granularity—from object universality (e.g., identifying any ``car'') to specificity (e.g., distinguishing an ``ambulance'' from a ``sedan'').

\textbf{Attribute} covers appearance cues (e.g., color, texture, material) and spatial properties (e.g., location, count). We also extract symbolic visual information such as text using optical character recognition. %

\textbf{Relation} assesses interactions between entities, including inter-object and object-scene relationships, as well as spatial and part-whole relations (e.g., ``wheel is part of car''), physical interactions (e.g., ``man holding umbrella''), and dynamic cues such as posture, expression, and motion.

\textbf{Camera} captures the photographer's perspective and natural conditions during the shot, including viewpoint (e.g., frontal, top-down), motion (e.g., zoom, pan), and lighting.

These dimensions are motivated by principles in visual neuroscience and cognitive psychology~\cite{desimone1984stimulus,puce1996differential}, and aligned with the structure of scene understanding in large multimodal models~\cite{dong2024benchmarking,lu2024benchmarking,liu2023visual}. 
These dimensions are then organically integrated into structural, inferential, and contextual similarity components within our framework, converted into two sub-indicators, as outlined below.

\begin{table}[t!]
    \centering
    \caption{Evaluation dimension and details. 
    }
    \vspace{-5pt}
    \label{tab:evaluation_dim}
    \resizebox{\linewidth}{!}{
    \begin{tabular}{rl}
    \toprule
    Dims. & Details \\
    \midrule
    scene & layout, geometry, event, action, style  \\ %
    object &	category, commonality (universality), specificity (differentiation) \\
    attribute &	appearance (color, texture, material), position, quantity, symbols and text \\
    relation &	spatial or part-whole relation, interaction, kinematics (motion, speed) \\ %
    camera	& lighting, camera angle, camera motion \\
    \bottomrule
    \end{tabular}
    }
\end{table}

\begin{table*}[th!]
\centering
\small
\caption{BASIC-H scores across stimulus-neuroimaging datasets. 
This metric, with sub-indicators for precision, recall, and F1 -- evaluated on objects, attributes, and relations -- quantifies high-level semantic correspondence between reconstructions and stimuli. \textbf{Best} and \underline{second best} are highlighted.
}
\vspace{-5pt}
\label{tab:basic_h}
\resizebox{0.85\linewidth}{!}
{
\begin{tabular}{l|ccc|ccc|ccc|c}
\toprule
\multirow{2}{*}{Method} & \multicolumn{3}{c|}{Object}  & \multicolumn{3}{c|}{Attribute}  & \multicolumn{3}{c|}{Relation} & \multirow{2}{*}{BASIC-H} \\
& P & R & F1 & P & R & F1 & P & R & F1 & \\
\midrule
\multicolumn{11}{c}{{NSD~\cite{allen2022massive}}} \\
\midrule
SDRecon~\cite{takagi2023high} & 55.59 & 53.05 & 53.79 & 10.12 & 38.73 & 14.96 & 40.15 & 38.71 & 39.06 & 35.31 \\
BrainDiffuser~\cite{ozcelik2023brain} & 57.87 & 59.11 & 58.09 & 13.35 & 45.82 & 19.43 & 43.20 & 44.42 & 43.50 & 39.71 \\
MindEye~\cite{scotti2023reconstructing}   & 62.94 & 60.64 & 61.26 & 18.14 & 51.17 & 25.06 & 49.98 & 48.42 & 48.84 & 44.30 \\
DREAM~\cite{xia2024dream} & \textbf{65.63} & \underline{63.06} & \underline{63.56} & 18.97 & 50.68 & 25.92 & \underline{53.45} & \textbf{53.21} & \underline{52.91} & \underline{46.37} \\
MindEye2~\cite{scotti2024mindeye2}  & 62.57 & 62.12 & 61.72 & 17.86 & 50.16 & 24.71 & 49.72 & 49.17 & 49.07 & 44.39 \\
MindBridge~\cite{wang2024mindbridge} & 59.00 & 58.35 & 58.19 & 13.70 & 45.69 & 19.49 & 45.75 & 45.78 & 45.43 & 40.16 \\
UMBRAE~\cite{xia2024umbrae}    & 62.00 & 61.86 & 61.44 & 17.51 & 51.32 & 24.49 & 48.91 & 48.71 & 48.45 & 44.06 \\
NeuroPictor~\cite{huo2024neuropictor}   & 63.00 & 61.05 & 61.38 & 17.92 & 50.13 & 24.66 & 49.62 & 49.08 & 48.98 & 44.21 \\
NeuroVLA~\cite{shen2024neuro} & \underline{65.36} & \textbf{65.03} & \textbf{64.57} & \textbf{21.27} & \textbf{53.73} & \textbf{28.65} & \textbf{53.80} & \underline{52.86} & \textbf{52.95} & \textbf{47.88} \\
SepBrain~\cite{wang2024unibrain}  & 62.10 & 60.19 & 60.57 & 16.62 & 48.71 & 23.31 & 48.03 & 47.55 & 47.44 & 43.04 \\
UniBrain~\cite{wang2024unibrain}  & 59.03 & 58.21 & 58.07 & 13.27 & 43.89 & 19.02 & 45.55 & 45.58 & 45.25 & 39.89 \\
STTM~\cite{liu2025see}  & 64.50 & 62.50 & 62.88 & \underline{19.53} & 51.70 & \underline{26.64} & 51.17 & 50.28 & 50.36 & 45.88 \\
MindTuner~\cite{gong2025mindtuner} & 62.55 & 62.64 & 61.95 & 18.06 & 49.18 & 24.73 & 50.22 & 50.10 & 49.80 & 44.63 \\
BrainGuard~\cite{tian2025brainguard}    & 63.63 & 62.36 & 62.43 & 18.66 & \underline{52.40} & 25.84 & 51.25 & 50.72 & 50.60 & 45.43 \\
\midrule
\multicolumn{11}{c}{{EEG-Things ~\cite{grootswagers2022human}}} \\
\midrule
ATM~\cite{li2024visual}       & \textbf{44.54} & \textbf{44.51} & \textbf{43.77} & \textbf{9.65}  & \textbf{38.48} & \textbf{14.46} & \textbf{36.74} & \textbf{36.63} & \textbf{36.28} & \textbf{30.55} \\
CognitionCapturer~\cite{zhang2025cognitioncapturer} & 42.20 & 43.99 & 42.59 & 7.56  & 34.56 & 11.67 & 34.31 & 35.74 & 34.70 & 28.64 \\
\midrule
\multicolumn{11}{c}{{CC2017~\cite{wen2018neural}}} \\
\midrule
MinD-Video~\cite{chen2023cinematic}  & \underline{47.56} & 45.05 & 45.34 & \underline{6.95}  & \underline{29.27} & \underline{10.57} & 32.42 & 31.10 & 31.32 & 28.63 \\
NeuroClips~\cite{gong2024neuroclips}  & \textbf{62.82} & \textbf{61.68} & \textbf{61.28} & \textbf{17.12} & \textbf{50.82} & \textbf{24.30} & \textbf{53.91} & \textbf{56.33} & \textbf{54.42} & \textbf{45.12} \\
DecoFuse~\cite{li2025decofuse}   & 47.35 & \underline{47.12} & \underline{46.74} & 6.26  & 28.12 & 9.77  & \underline{32.75} & \underline{32.18} & \underline{32.11} & \underline{29.03} \\
\midrule
\multicolumn{11}{c}{{SEED-DV~\cite{liu2024eeg2video}}} \\
\midrule
EEG2Video~\cite{liu2024eeg2video}   & 66.05 & 65.75 & 64.36 & 25.43 & 51.85 & 32.20 & 54.63 & 56.02 & 54.59 & 49.54 \\
\midrule
\multicolumn{11}{c}{fMRI-Shape~\cite{gao2024mind3d}} \\
\midrule
MinD-3D~\cite{gao2024mind3d}    & \textbf{41.52} & 36.42 & 37.59 & 13.52 & 44.00 & 19.15 & 42.49 & 41.12 & 41.26 & 30.95 \\
MinD-3D++~\cite{gao2024mind3d++}    & 41.35 & \textbf{37.92} & \textbf{38.12} & \textbf{20.03} & \textbf{61.89} & \textbf{27.80} & \textbf{44.82} & \textbf{43.39} & \textbf{43.57} & \textbf{35.08} \\
\midrule
\multicolumn{11}{c}{EEG-3D~\cite{guo2025neuro3d}} \\
\midrule
Neuro-3D~\cite{guo2025neuro3d} & 35.79 & 32.86 & 34.26 & 6.32 & 27.92 & 10.32 & 23.41 & 29.87 & 26.24 & 23.08 \\
\bottomrule
\end{tabular}
}
\end{table*}

\begin{table*}[t!]
\centering
\small
\caption{BASIC-L scores across datasets. This metric evaluates low-level structural correspondence between reconstructed and reference images at four granularities: foreground saliency, semantic consistency, instance separation, and part-level delineation. 
}
\vspace{-5pt}
\label{tab:basic_l}
\resizebox{0.85\linewidth}{!}
{
\begin{tabular}{l|cc|cc|cc|cc|cc|c}
\toprule
\multirow{2}{*}{Method} & \multicolumn{2}{c|}{F (Foreground)} & \multicolumn{2}{c|}{B (Binary)} & \multicolumn{2}{c|}{S (Semantic)} & \multicolumn{2}{c|}{I (Instance)} & \multicolumn{2}{c|}{P (Part)} & \multirow{2}{*}{BASIC-L} \\
 & IoU & AP & IoU & AP & IoU & AP & IoU & AP & IoU & AP & \\
\midrule
\multicolumn{12}{c}{{NSD~\cite{allen2022massive}}} \\
\midrule
SDRecon~\cite{takagi2023improving}                  & 9.03 & 13.06 & 38.90 & 49.97 & 15.08 & 21.43 & 17.84 & 1.07 & 4.38 & 6.79 & 11.81 \\
BrainDiffuser~\cite{ozcelik2023brain} & 17.96 & 20.21 & 38.98 & 45.85 & 18.66 & 20.78 & 20.09 & 1.94 & 7.86 & 7.82 & 16.65 \\
MindEye~\cite{scotti2023reconstructing}   & 19.20 & 22.79 & 45.53 & 55.17 & 18.73 & 21.94 & 20.36 & 1.99 & 7.49 & 8.26 & 17.03 \\
DREAM~\cite{xia2024dream}     & 23.62 & 26.10 & 46.03 & 57.10 & 21.15 & 24.13 & 21.41 & 2.32 & 9.22 & 8.79 & 19.57 \\
MindEye2~\cite{scotti2024mindeye2}    & 25.29 & 26.27 & 47.93 & 57.52 & 24.33 & 25.68 & 24.09 & 3.45 & 12.32 & 11.03 & 22.16 \\
MindBridge~\cite{wang2024mindbridge} & 16.24 & 19.04 & 40.51 & 48.95 & 16.87 & 19.81 & 18.61 & 1.54 & 6.30 & 6.78 & 15.00 \\
UMBRAE~\cite{xia2024umbrae}   & 21.33 & 24.62 & 40.96 & 48.53 & 18.94 & 21.16 & 20.29 & 2.15 & 8.43 & 8.47 & 17.89 \\
NeuroPictor~\cite{huo2024neuropictor} & \textbf{29.45} & \textbf{31.29} & 47.79 & 56.48 & \textbf{27.97} & \textbf{29.57} & \textbf{26.47} & \textbf{4.08} & \textbf{17.17} & \textbf{15.84} & \textbf{25.88} \\
NeuroVLA~\cite{shen2024neuro} & 16.03 & 20.85 & 44.09 & 54.42 & 13.84 & 17.54 & 17.57 & 1.34 & 4.38 & 5.57 & 13.54 \\
SepBrain~\cite{wang2024unibrain}  & 21.22 & 23.72 & 45.06 & 53.47 & 20.88 & 23.32 & 21.98 & 2.51 & 8.79 & 8.98 & 18.84 \\
UniBrain~\cite{wang2024unibrain}  & 15.24 & 18.02 & 37.48 & 45.09 & 14.99 & 17.75 & 17.45 & 1.06 & 5.54 & 6.07 & 13.79 \\
STTM~\cite{liu2025see}   & \underline{27.31} & \underline{29.44} & \textbf{49.35} & \textbf{59.05} & \underline{24.61} & \underline{26.37} & \underline{24.19} & \underline{3.50} & \underline{12.55} & \underline{11.65} & \underline{22.90} \\
MindTuner~\cite{gong2025mindtuner}   & 15.38 & 16.66 & 40.74 & 49.02 & 21.46 & 24.50 & 21.49 & 1.97 & 8.16 & 8.13 & 16.98 \\
BrainGuard~\cite{tian2025brainguard}  & 25.90 & 28.07 & \underline{49.22} & \underline{58.99} & 23.34 & 25.21 & 23.98 & 3.29 & 10.82 & 10.32 & 21.76 \\
\midrule
\multicolumn{12}{c}{{EEG-Things ~\cite{grootswagers2022human}}} \\
\midrule
ATM~\cite{li2024visual}       & \textbf{13.87} & \textbf{18.77} & \textbf{39.46} & \textbf{50.14} & \textbf{22.85} & \textbf{34.69} & \textbf{22.02} & \textbf{2.15} & 11.13 & 17.07 & \textbf{17.60} \\
CognitionCapturer~\cite{zhang2025cognitioncapturer} & 10.26 & 12.71 & 33.31 & 40.67 & 21.37 & 29.84 & 20.75 & 1.37 & \textbf{13.08} & \textbf{17.26} & 16.22 \\
\midrule
\multicolumn{12}{c}{{CC2017~\cite{wen2018neural}}} \\
\midrule
MinD-Video~\cite{chen2023cinematic}   & \underline{12.82} & \underline{24.20} & 44.89 & 53.50 & 20.29 & 33.57 & 19.87 & \underline{3.53} & 6.83 & \textbf{14.69} & \underline{15.25} \\
NeuroClips~\cite{gong2024neuroclips}  & \textbf{24.37} & \textbf{31.59} & \textbf{65.51} & \textbf{73.39} & \textbf{28.23} & \textbf{36.00} & \textbf{28.74} & \textbf{7.73} & \textbf{9.82} & \underline{13.14} & \textbf{23.52} \\
DecoFuse~\cite{li2025decofuse}    & 12.32 & 18.07 & \underline{49.87} & \underline{59.46} & \underline{22.33} & \underline{34.63} & \underline{20.87} & 2.74 & 4.06 & 9.80 & 15.31 \\
\midrule
\multicolumn{12}{c}{{SEED-DV~\cite{liu2024eeg2video}}} \\
\midrule
EEG2Video~\cite{liu2024eeg2video} & 27.77 & 32.82 & 57.26 & 69.97 & 22.93 & 31.96 & 23.41 & 3.51 & 3.10 & 7.10 & 20.54 \\
\midrule
\multicolumn{12}{c}{fMRI-Shape~\cite{gao2024mind3d}} \\
\midrule
MinD-3D~\cite{gao2024mind3d}  & \textbf{5.69} & \textbf{8.10} & \textbf{5.69} & \textbf{8.10} & \textbf{19.91} & {29.35} & \textbf{19.38} & {1.45} & \textbf{15.96} & {25.30} & \textbf{14.72} \\
MinD-3D++~\cite{gao2024mind3d++}  & 3.80 & 5.23 & 3.80 & 5.23 & 16.03 & \textbf{35.83} & 17.96 & \textbf{1.83} & 14.94 & \textbf{36.19} & 12.62 \\
\midrule
\multicolumn{12}{c}{EEG-3D~\cite{guo2025neuro3d}} \\
\midrule
Neuro-3D~\cite{guo2025neuro3d} & 2.39 & 3.65 & 2.39 & 3.65 & 13.77 & 25.92 & 12.25 & 1.02 & 12.33 & 18.56 & 9.69 \\
\bottomrule
\end{tabular}
}
\end{table*}

\subsection{BASIC-H}
\label{subsec:basic_h}

BASIC-H quantifies high-level semantic correspondence by integrating inferential and contextual similarities, which respectively capture complementary aspects of conceptual accuracy and narrative coherence, for a holistic evaluation of reconstructed scene semantics.
Specifically, BASIC-H operates in an automated pipeline with three steps: (1) detailed description generation, where state-of-the-art MLLMs produce semantic-rich captions for both reconstructed and reference images; (2) semantic graph construction, which parses captions into semantic graphs representing objects, attributes, and relations; and (3) structured semantic matching, which computes semantic correspondace using symbolic concept matching and embedding-based similarities. 
See~\cref{fig:overview} for the method overview.

\paragraph{Detailed description generation.} To support multigranular evaluation, we first prompt MLLMs to produce semantically rich descriptions from images. This \textit{description prompt} extracts detailed information across key semantic components corresponding to the semantic dimensions outlined in~\cref{tab:evaluation_dim}, including scene context, objects, object attributes, inter-object relations, and camera viewpoint (when applicable). Recent studies indicate that state-of-the-art MLLMs can generate captions on par with human experts, albeit with occasional hallucinations~\cite{lu2024benchmarking,dong2024benchmarking}. 

\paragraph{Semantic graph construction.} Given detailed captions, we first segment them into individual sentences using NLTK~\cite{loper2002nltk}. We then extract key visual elements -- objects, attributes, and relations -- using a T5-based factual parser~\cite{li2023factual}.
To handle objects mentioned across multiple sentences in the caption, we consolidate references to the same object into subcaptions and bind attributes to the correct instances, avoiding indiscriminate merging. For directional relations (e.g., subject-object pairs), we concatenate relevant sentences to maintain relational coherence~\cite{lu2024benchmarking,dong2024benchmarking}.

\paragraph{Structured semantic matching.} The matching process aligns the extracted objects, attributes, and relations in three steps: (a) \textit{exact matching}, which identifies direct lexical matches between visual elements from the reconstruction and the reference stimuli; (b) \textit{synonym matching}, which aligns semantically equivalent terms, such as ``building'' and ``edifice'' for objects; ``bright blue'' and ``azure'' for attributes; ``next to'' and ``beside'' for relations; and (c) \textit{semantic matching}, which computes cosine similarity for any remaining unmatched elements based on contextual meaning. %
This multi-step process ensures that all visual elements, whether directly mentioned or linguistically varied, are appropriately matched across decoded and reference images. %

Finally, we compute semantic alignment scores based on the matching results. To mitigate the effects of omissions and hallucinations in MLLM outputs, we evaluate precision, recall, and F1 scores for each type of visual element (object, attribute, relation), based on the matching outcomes. The overall BASIC-H score is computed as a weighted sum of the three F1 scores, with weights of $\alpha_1$, $\alpha_2$, $\alpha_3$ assigned to objects, attributes, and relations, respectively. 
Precision penalizes hallucinations by measuring the proportion of correctly matched elements among all generated ones, whereas recall captures omissions by evaluating how completely the reference content is recovered. Their combination in F1 score enables a balanced assessment of both over- and under-generation issues common in MLLM outputs.

\subsection{BASIC-L}
\label{subsec:basic_l}

BASIC-L evaluates the structural correspondence between the reconstructed and reference images across four granularities: foreground saliency, semantic consistency, instance separation, and part-level delineation. 
The process first extracts key hierarchical visual components and organizes them into natural language expressions in a progressively granular, whole-to-part format using structured prompts. It then applies referring expression comprehension to localize specific objects in the scene based on these expressions.
The evaluation follows two steps:

\paragraph{Semantic instance categorization.} We prompt MLLMs to generate structured semantic annotations from images.
These annotations include object-centric identifications that specify spatial roles (e.g., foreground or background), semantic categories (e.g., ``dog'', ``tree''), and part-level component decompositions (e.g., ``head'', ``leg”).
This process incorporates a multi-level decomposition of visual elements, 
capturing a hierarchical view of the scene components
and facilitating further grounded segmentation and multigranular structural alignment. 

\paragraph{Progressive granular segmentation.}

The categories from these structured annotations are then used as prompts for referring expression comprehension methods~\cite{ren2024grounded,kirillov2023segment}, which produces segmentation masks for each identified object and component. 
The segmentation results are decided by the text and box thresholds, where only the highest-similarity boxes exceeding the box threshold and words with similarity scores above the text threshold are considered as predicted labels~\cite{liu2024grounding}.

\paragraph{Hierarchical score calculation.}
We compare predicted masks against those from the reference image across four granularities: salient (foreground categories, F), binary (foreground and background categories, B), semantic (all distinct categories, S), instance (individual instances of all identified object categories, I), and part (subobject components, mostly salient foreground objects, P), computing intersection-over-union (IoU) and average precision (AP). %

We aggregate saliency, semantic, instance, and part-level segmentation scores into the overall BASIC-L metric via a weighted sum of respective IoUs using weights of $\beta_1$, $\beta_2$, $\beta_3$, $\beta_4$ for each granularity. 
This scheme prioritizes global layout and object-level coherence while accounting for fine-grained details, reflecting the limited granularity of information captured during the brain data acquisition experiments.

\section{Experiments}
\label{sec:exp}

\subsection{Experimental Setup}
\label{subsec:experimental_details}

\begin{figure}[t!]
    \centering
    \includegraphics[width=0.8\linewidth]{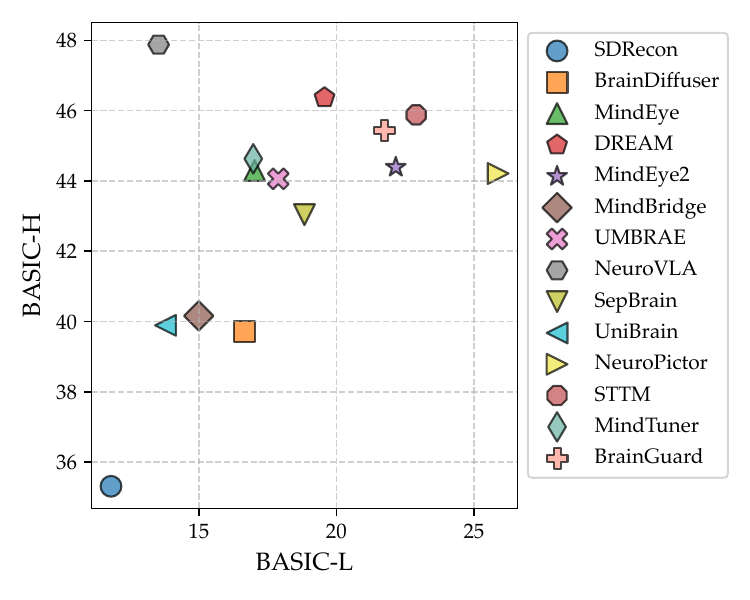}
    \vspace{-5pt}
    \caption{BASIC performance.}
    \label{fig:nsd_scatterplot}
\end{figure}

\paragraph{Implementation details.}

We use LLaVA-1.6-13B~\cite{liu2023visual} as the default MLLM for detailed description generation and semantic instance categorization. 
Ground truth captions are obtained from human experts through error correction, missing element addition, and hallucination removal, based on GPT-4o-generated captions, for better alignment with human judgment.
For multigranular segmentation, we employ Grounded-SAM2~\cite{ren2024grounded}, an open-source referring expression comprehension tool.
The default weights 
for BASIC-H
and BASIC-L are set to 4:4:2 and 3:2.5:2.5:2, respectively, reflecting the empirically informed importance of each sub-indicator in capturing semantic and structural alignment.
Experiments are conducted on an NVIDIA A100 GPU. 
To ensure broad applicability and methodological consistency, we refrain from using MLLMs or segmentation methods specifically designed for video or 3D data. 
Instead, we retain an image-based framework with automatic selection of representative video frames and rendered 3D views that best capture key semantics and structure from stimuli and decoded reconstructions.%

\subsection{Experimental Results}
\label{subsec:experimental_results}

While previous metrics suffer from score saturation and fail to effectively distinguish between method performances (see supplementary material), our metrics enable more nuanced evaluation. They offer clearer differentiation across objects, attributes, and relations for high-level semantics, as well as across salient, semantic, instance, and part-level granularity for low-level structure. Detailed descriptions of each dimension are provided in~\cref{tab:basic_h} and~\cref{tab:basic_l}.

\paragraph{Semantics:~object, attribute, and relation.} \cref{tab:basic_h} presents a comparison of object, attribute, and relation prediction performance across several datasets. 
While prior metrics often produce saturated or undifferentiated scores across models,
our proposed multigranular evaluation reveals meaningful variations in object coherence, attribute accuracy, and relational plausibility, offering a more fine-grained diagnostic lens.
Here, for methods on NSD~\cite{allen2022massive}, NeuroVLA~\cite{shen2024neuro}, DREAM~\cite{xia2024dream}, and STTM~\cite{liu2025see} achieve the highest BASIC-H scores,
reflecting their strength in modeling rich visual semantics. This likely stems from detailed caption generation, complex visual semantics modeling, cross-subject training, or their combinations. A closer comparison between NeuroVLA and DREAM shows that while both models reconstruct correct instances (achieving higher precision) across the three semantic dimensions, 
NeuroVLA misses fewer relevant instances (better recall).
In contrast, SDRecon~\cite{takagi2023improving}, BrainDiffuser~\cite{ozcelik2023brain}, and UniBrain~\cite{wang2024unibrain} lag behind with BASIC-H scores.
These methods struggle notably with the identification of object categories, attributes, and relations, indicating limitations in handling semantics or generalizing across visual categories.

\paragraph{Structure: salient, semantic, instance, and part.} 
\cref{tab:basic_l} reports performance across salient, binary, semantic, instance, and part segmentation matching scores. For decoding methods on NSD,
NeuroPictor~\cite{huo2024neuropictor} leads with the highest BASIC-L score,
driven by its superior performance in instance and part segmentation. This reflects its ability to reconstruct fine-grained visual details and delineate object boundaries from neural signals.
STTM~\cite{liu2025see} and MindEye2~\cite{scotti2024mindeye2} perform competitively.
SDRecon~\cite{takagi2023improving} and NeuroVLA~\cite{shen2024neuro}
show relatively poor spatial structural fidelity.
BrainGuard~\cite{tian2025brainguard} achieves competitive scores in salient and semantic categories but performs less satisfactorily in instance and part segmentation. This indicates that while the model captures the overall structure of salient objects and categories, it faces challenges in distinguishing between multiple instances within the same category and identifying sub-component relationships.

\cref{fig:nsd_scatterplot} provides performance of visual decoding models on NSD in terms of structural alignment (BASIC-L) and semantic alignment (BASIC-H). Each point represents a model, revealing trade-offs between spatial fidelity and semantic coherence. Models closer to the top-right demonstrate stronger performance across both dimensions.
For structural alignment, NeuroPictor~\cite{huo2024neuropictor}, STTM~\cite{liu2025see}, and MindEye2~\cite{scotti2024mindeye2} achieve the highest BASIC-L scores, reflecting better spatial reconstruction. In contrast, semantic alignment (BASIC-H) is best captured by NeuroVLA~\cite{shen2024neuro}, DREAM~\cite{xia2024dream}, and STTM~\cite{liu2025see}, indicating superior semantic preservation.

\begin{figure*}[t!]
    \centering
    \includegraphics[width=0.92\linewidth]{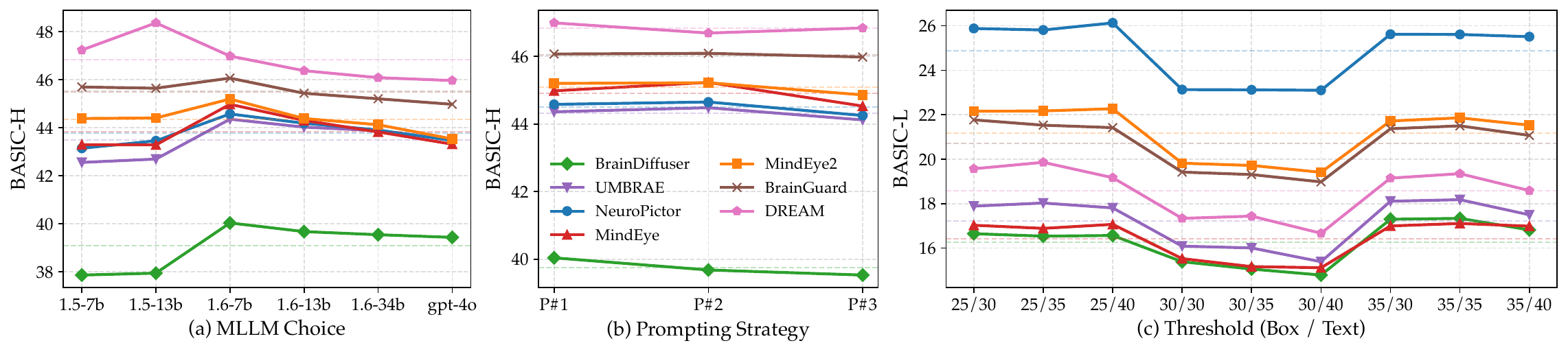}
    \vspace{-5pt}
    \caption{BASIC demonstrates stable and consistent performance in method evaluation across variations in (a) MLLMs~\cite{liu2023visual}, (b) prompting strategies, and (c) thresholds for box and text~\cite{ren2024grounded}.
    }
    \label{fig:ablation}
\end{figure*}

\begin{figure}[t!]
    \centering
    \includegraphics[width=0.98\linewidth]{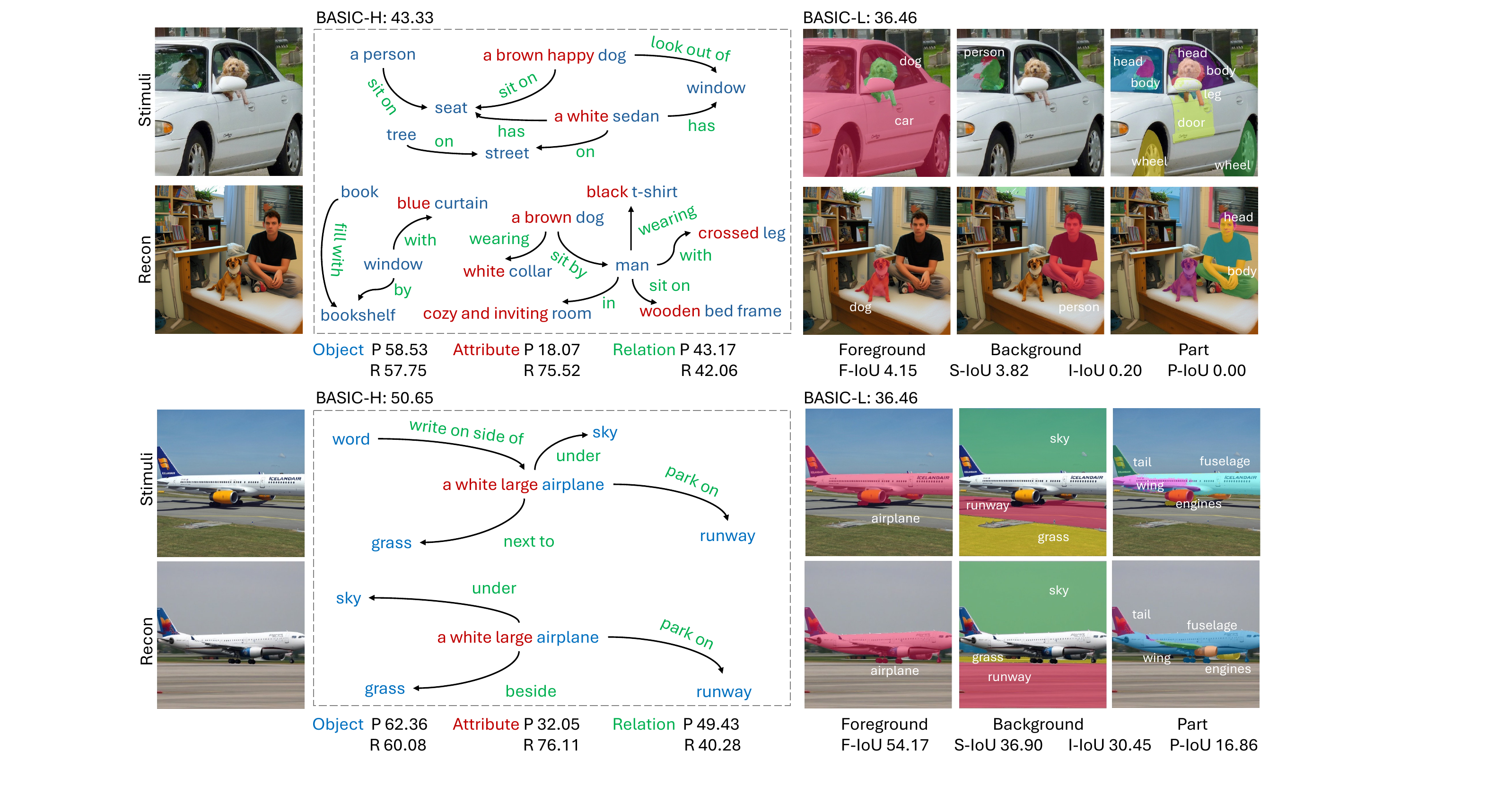}
    \vspace{-5pt}
    \caption{Qualitative examples with BASIC-H and BASIC-L scores, including sub-indicators.
    }
    \label{fig:qualitative_example}
\end{figure}

\subsection{Discussion}
\label{subsec:discussion}

\paragraph{Toward open, stable, and versatile evaluation.} 
\textbf{Open:} Our evaluation pipeline is designed to be model-agnostic, avoiding reliance on proprietary or task-specific components to ensure broad applicability and reproducibility. While models like GPT-4o are shown to have stronger multimodal performance in captioning benchmarks~\cite{lu2024benchmarking,dong2024benchmarking}, they are not suitable for open, large-scale benchmarking due to API restrictions and cost. We use LLaVA-1.6-13B, which balances captioning accuracy and computational efficiency. %
\textbf{Stable:} The methods under our metrics demonstrate stable and consistent performance despite variations in (a) MLLMs, (b) prompting strategies, and (c) thresholds for box and text, as shown in \cref{fig:ablation}. The relative ranking and discriminative power across decoding methods remain consistent, although absolute scores may vary slightly. The stable BASIC-H results across different MLLMs and prompts indicate that our metric reliably captures semantic performance differences among methods. Similarly, the BASIC-L scores provide consistent structural evaluation across varying text and box threshold settings, where only the highest similarity boxes exceeding the box threshold and words with similarity scores above the text threshold are considered predicted labels.
\textbf{Versatile:}  
For video and 3D data evaluation, instead of using modality-specific tools,
we maintain the same image-based pipeline, with automatic selection of representative video frames and 3D-rendered views to ensure compatibility across formats without compromising evaluation quality.

\paragraph{Toward informative diagnostic insight.}

This multigranular, interpretable feedback provides fine-grained diagnostic insights into brain-based visual decoding and can help uncover blind spots. 
Decoding methods that rely on pretrained models benefit from powerful generative capabilities but are also susceptible to systematic biases, often introducing hallucinated details such as prototypical co-occurrence patterns. For instance, \textit{savannah} may be infered when the decoded scene label is \textit{giraffe} even though the original stimulus was a \textit{zoo}. These hallucinations tend to affect high-level contextual information—such as scene types or typical object-environment associations—whereas low-level attributes (e.g., blue) and spatial relationships (e.g., to the left) are less likely to be inferred without support from brain-derived signals. 
BASIC-H mitigates this confound and allows us to better isolate the contribution of genuine brain-derived information and more accurately assess the true decoding performance.
BASIC-H penalizes misidentifications, such as hallucinated objects, incorrect attributes, and implausible relationships. 
These semantic-level discrepancies are directly reflected in the precision and recall of object categories, attributes, and relations, offering interpretable signals beyond opaque scores from pretrained networks.
The segmentation-based BASIC-L scores also serve diagnostic purposes, measuring structure correspondance across four granularities: salient, semantic, instance and part.
\cref{fig:qualitative_example} presents qualitative examples illustrating how BASIC-H and BASIC-L enable fine-grained breakdowns of semantic and structural alignment. In the top reconstruction example from NeuroVLA~\cite{shen2024neuro}, the concepts of ``dog'' and ``man'' are correctly predicted but placed in an incorrect scene. The bottom example from MindEye2 in the airplane scenario correctly predicts both semantic and structural aspects, therefore achieving a high score at both dimensions.

\paragraph{Towards cross-dataset performance dissection.}

Our evaluation follows a unified protocol across visual modalities, enabling cross-dataset dissection. The fMRI-based image  reconstructions~\cite{ozcelik2023brain,scotti2023reconstructing} consistently outperform EEG-based counterparts~\cite{li2024visual,zhang2025cognitioncapturer}, attributable to intrinsic limitations of EEG in spatial resolution and information.
The fMRI-to-video reconstructions~\cite{gong2024neuroclips,chen2023cinematic,li2025decofuse} currently struggles to preserve the structure of salient objects, but performs comparably to image modalities in terms of instance semantic identification and structural preservation.
The high performance in EEG-to-video decoding~\cite{liu2024eeg2video} appears to reflect the simplicity of stimuli -- short clips with prominent foregrounds and minimal background clutter. Future work explore more semantically rich and visually complex video scenarios.
For 3D reconstruction, both fMRI-based and EEG-based visual decoding methods~\cite{gao2024mind3d,gao2024mind3d++,guo2025neuro3d} struggle to recover even basic semantics, despite simple, canonical structures of the target object categories. The incapability in semantic identification and structure preservation highlights the need for decoding models with stronger semantic and geometric priors. 

\section{Conclusion}
\label{sec:conclusion}

We present a brain-based visual decoding evaluation framework that captures the multigranular nature of human visual perception. By integrating mask-based segmentation alignment with structured object-attribute-relation similarity, our approach enhances performance discriminability, neuroscientific validity, and semantic interpretability. The resulting BASIC metrics provide a comprehensive assessment across structural precision, inferential accuracy, and contextual coherence.
We benchmark diverse brain visual decoding models across six major stimulus-neuroimaging datasets under a unified evaluation protocol, establishing the first standardized, interpretable, and extensible benchmark for this task.

\noindent{}{%
\textbf{Acknowledgements.} This work was 
supported by a UKRI Future Leaders Fellowship [grant number G104084].%
}

\bibliography{reference}

\ifshowappendix

\clearpage
\appendix
\onecolumn

\begin{center}
{\LARGE \textbf{Appendix}\\[2pt]
\vspace*{0.3em}
\large Supplemental Material for ``Multigranular Evaluation for Brain Visual Decoding''\par}
\end{center}

\setcounter{figure}{0}
\renewcommand{\thefigure}{S\arabic{figure}}
\setcounter{table}{0}
\renewcommand{\thetable}{S\arabic{table}}

\section{Details on Stimulus-Neuroimaging Datasets}
\label{sec:supmat_dataset}

This section provides details on stimulus-neuroimaging datasets.
\cref{tab:supmat_stimulus_neuroimaging_data} summarizes the core properties of these stimulus-neuroimaging combinations. 
The Natural Scenes Dataset~\cite{allen2022massive} has recently garnered significant attention, with many methods proposed for its analysis, and most of our experiments focus on methods applied to this dataset for comprehensive and detailed analysis. 

\paragraph{NSD.} NSD~\cite{allen2022massive} is the largest public 7 Tesla fMRI scans, featuring brain recordings from eight participants who passively viewed images from the Common Objects in Context (COCO) dataset~\cite{lin2014microsoft} for up to 40 hours in an MRI machine. Each image was shown for three seconds and repeated three times across 30-40 scanning sessions, resulting in 22,000-30,000 fMRI response trials per participant.

\paragraph{EEG-Things.} EEG-Things~\cite{grootswagers2022human} provides neuroimaging recordings for a systematic collection of objects and concepts, featuring electroencephalography (EEG) responses from 50 subjects to a stimulus set of 1,854 object concepts and 22,248 images.

\paragraph{CC2017.} CC2017~\cite{wen2018neural} is a public benchmark consisting of video clips and corresponding fMRI recordings collected from three subjects using a 3T MRI scanner. 
The training set includes 18 segments of 8-minute video clips (totaling 2.4 hours), yielding 4,320 paired fMRI-video examples. The test set includes 5 segments (totaling 40 minutes), providing 1,200 samples. %

\paragraph{SEED-DV.} SEED-DV~\cite{liu2024eeg2video} includes EEG recordings from 20 subjects using a 62-channel system as they viewed 1,400 video clips covering 40 object concepts. These concepts were grouped into 9 broader categories, offering dynamic visual stimuli paired with corresponding neural responses.

\paragraph{fMRI-3D.} The fMRI-3D dataset comprises two successive components for neural decoding of 3D visual perception. fMRI-Shape~\cite{gao2024mind3d} collects fMRI responses from 14 participants viewing 360-degree videos of 1,624 3D objects spanning 55 categories, totaling 123,200 frames. The subsequent fMRI-Objaverse~\cite{gao2024mind3d++} further expands category diversity and object coverage, with data from 5 participants exposed to 3,142 3D objects spanning 117 categories, each accompanied by descriptive text captions.

\paragraph{EEG-3D.} EEG-3D~\cite{guo2025neuro3d} provides extensive EEG recordings from 12 participants who viewed 3D objects across 72 categories, rendered as both videos and images. Each participant underwent approximately 5.5 hours of recording. The dataset includes 10 objects per category, selected for shape diversity, with each object paired with a descriptive text caption.

\begin{table*}[thbp]
    \centering
    \caption{Details on stimulus-neuroimaging datasets.}
    \label{tab:supmat_stimulus_neuroimaging_data}
    \resizebox{\linewidth}{!}{
    \begin{tabular}{llccl}
        \toprule
         Dataset & Venue & Neuroimaging & Stimulus & Method \\
        \midrule
         NSD~\cite{allen2022massive} & Nature Neuroscience 2022 & fMRI & Image & \citet{takagi2023improving,scotti2023reconstructing,scotti2024mindeye2} %
         \\
         EEG-Things~\cite{grootswagers2022human} & Scientific Data 2022 & EEG  & Image & \citet{li2024visual,zhang2025cognitioncapturer} \\
         CC2017~\cite{wen2018neural} & Cerebral Cortex 2018 & fMRI & Video & \citet{li2025decofuse,gong2024neuroclips} %
         \\
         SEED-DV~\cite{liu2024eeg2video} & NeurIPS 2024 & EEG &  Video & \citet{liu2024eeg2video} \\
         fMRI-3D~\cite{gao2024mind3d} & ECCV 2024  & fMRI & 3D Shape & \citet{gao2024mind3d,gao2024mind3d++} \\
         EEG-3D~\cite{guo2025neuro3d} & CVPR 2025 & EEG & 3D Shape & \citet{guo2025neuro3d} \\
        \bottomrule
    \end{tabular}
    } 
\end{table*}

The latter three datasets, SEED-DV~\cite{liu2024eeg2video}, fMRI-3D~\cite{gao2024mind3d}, EEG-3D~\cite{guo2025neuro3d}, are the most recently proposed datasets, with limited comparisons available. We share their results to demonstrate that the proposed evaluation framework can be extended to different stimulus-neuroimaging datasets.

\section{Details on Prior Evaluation Procedure}
\label{sec:supmat_evaluation}

Brain visual decoding lacks consistent evaluation standards, with different datasets often employing distinct metrics for performance assessment. This section introduces the common evaluation metrics used in the field and outlines how these metrics assess model performance.

\cref{tab:quantitative_nsd_eeg_things} presents results on NSD and EEG-Things using eight commonly used metrics: PixCorr, SSIM~\cite{wang2004image}, AlexNet~\cite{krizhevsky2017imagenet}-2/5, Inception~\cite{szegedy2016rethinking}, CLIP~\cite{radford2021learning}, EffNet~\cite{tan2019efficientnet}, and SwAV~\cite{caron2020unsupervised}.
PixCorr measures the Pearson correlation between the pixel values of the ground-truth (GT) image and the reconstruction, quantifying the pixel-level similarity. SSIM (Structural Similarity Index)~\cite{wang2004image} compares the structural and textural features of the GT and the reconstruction.
The AlexNet-2/5 metrics refer to the similarity between the embeddings of the GT and the reconstruction, based on the deep features extracted from the second and fifth layers of the pretrained AlexNet~\cite{krizhevsky2017imagenet}, respectively.
Similarly, Inception refers to the comparison of the final pooling layer in InceptionV3~\cite{szegedy2016rethinking}, and CLIP corresponds to the comparison of embeddings from the final layer of CLIP-Vision~\cite{radford2021learning}. 
EffNet-B and SwAV metrics refer to the distance or similarity between the GT embedding and the reconstructed embedding using EfficientNet-B1~\cite{tan2019efficientnet} and SwAV-ResNet50~\cite{caron2020unsupervised}, respectively. 

\cref{tab:supmat_cc2017} presents a quantitative evaluation of video reconstruction on CC2017~\cite{wen2018neural}, using 2-way and 50-way semantic accuracy as the semantic metrics, and the ratio of foreground-background matching (mask matching ratio, MMR) between the ground truth and decoded images as the spatial-level metric.
\cref{tab:supmat_eeg2video}, \cref{tab:supmat_neuro3d}, and \cref{tab:supmat_mind3d} shows quantitative evaluation of visual reconstruction on SEED-DV~\cite{liu2024eeg2video}, EEG-3D~\cite{guo2025neuro3d}, and fMRI-3D~\cite{gao2024mind3d}, with results excerpted from the corresponding papers.
For ease of comparison, we also include BASIC-L and BASIC-H evaluation results. As shown, while NSD metrics saturates in scores and fails to distinguish method performance, our metrics, BASIC-L and BASIC-H, provide clearer differentiation and more discriminative information across objects, attributes, and relations for high-level semantics, as well as salient, semantic, instance, and part granularity for low-level structure. Details for each dimension are in~\cref{tab:basic_h} and~\cref{tab:basic_l}.

Despite slight variations in specific definitions, these evaluations utilize n-way identification metrics, which measure the percentage of cases where the ground truth embedding or category is closer to its corresponding reconstruction than to other reconstructions.
This n-way identification metric is problematic for providing consistent and nuanced assessments in cross-model comparisons, as the simplistic criterion only requires the reconstruction to be closer to the ground truth than a randomly selected alternative and each model is evaluated against different sets of reconstructions.

Other modality-specific metrics are also introduced. For instance, fMRI-3D evaluation uses common 3D reconstruction metrics, including Fréchet Point Cloud Distance (FPD) (scaled by $\times10^{-1}$), Chamfer Distance (CD) (scaled by $\times10^{2}$), and Earth Mover’s Distance (EMD) (scaled by $\times10^{2}$). These metrics are computed by sampling point clouds from both the ground truth and the generated meshes.

\setlength{\tabcolsep}{4pt}
\setlength{\fboxrule}{0pt} 
\setlength{\fboxsep}{2pt}
\begin{table*}[t!]
	\centering
	\caption{Quantitative evaluation for fMRI-to-image reconstruction on NSD~\cite{allen2022massive} and EEG-to-image reconstruction on EEG-Things~\cite{grootswagers2022human} following standard NSD and our BASIC metrics. While NSD metrics saturates in scores and fails to distinguish method performance, our metrics provide clearer differentiation and more discriminative information. See detailed sub-scores in~\cref{tab:basic_h} and~\cref{tab:basic_l}.
    }
	\label{tab:quantitative_nsd_eeg_things}
	\resizebox{\linewidth}{!}{
		\begin{tabular}{l|ccccc|ccccc}
			\toprule
			\multirow{2}{*}{\textsc{Method}}  & \multicolumn{5}{c|}{Low-Level} & \multicolumn{5}{c}{High-Level} \\
			~ & PixCorr $\uparrow$ & SSIM $\uparrow$ & Alex(2) $\uparrow$ & Alex(5) $\uparrow$ & BASIC-L $\uparrow$ & Incep $\uparrow$ & CLIP $\uparrow$  & EffNet $\downarrow$ & SwAV $\downarrow$ & BASIC-H $\uparrow$ \\
			\midrule
            \multicolumn{11}{c}{{NSD ~\cite{allen2022massive}}} \\
            \midrule
            SDRecon~\cite{takagi2023improving} & 0.246 & 0.410  & 78.9\% & 85.6\% & 11.81 & 83.8\% & 82.1\% & 0.811  & 0.504 & 35.31\\
            BrainDiffuser~\cite{ozcelik2023brain} & 0.273 & 0.365 & 94.4\% & 96.6\% & 16.65 & 91.3\% & 90.9\% & 0.728 & 0.422 & 39.71 \\
            MindEye~\cite{scotti2023reconstructing} & 0.319 & 0.360 & 92.8\% & 96.9\% & 17.03 & 94.6\% & 93.3\% & 0.648 & 0.377 & 44.30 \\
    		DREAM~\cite{xia2024dream} & 0.274 & 0.328 & 93.9\% & 96.7\% & 19.57 & 93.4\% & 94.1\% & 0.645 & 0.418 & \underline{46.37} \\
            MindEye2~\cite{scotti2024mindeye2} & \underline{0.322} & \textbf{0.431} & \underline{96.1\%} & \underline{98.6\%} & 22.16 & 95.4\% & 93.0\% & 0.619 & 0.344 & 44.39 \\
            MindBridge~\cite{wang2024mindbridge} & 0.151 & 0.263 & 87.7\% & 95.5\% & 15.00 & 92.4\% & 94.7\% & 0.712 & 0.418 & 40.16 \\ 
            UMBRAE~\cite{xia2024umbrae} & 0.283 & 0.341 & 95.5\% & 97.0\% & 17.89 & 91.7\% & 93.5\% & 0.700 & 0.393 & 44.06 \\
            NeuroPictor~\cite{huo2024neuropictor} & 0.229 & 0.375 & \textbf{96.5\%} & 98.4\% & \textbf{25.88} & 94.5\% & 93.3\% & 0.639 & 0.350 & 44.21 \\
            NeuroVLA~\cite{shen2024neuro} & 0.265 & 0.357 & 93.1\% & 97.1\% & 13.54 & \textbf{96.8\%}  & \textbf{97.5\%} & 0.633 & \textbf{0.321} &  \textbf{47.88} \\
            SepBrain~\cite{wang2024unibrain} & 0.309 & 0.317 & 94.2\% & 97.4\% & 18.84 & 94.5\%  & 95.3\% & 0.656 & 0.374 & 43.04 \\
            UniBrain~\cite{wang2024unibrain} &  0.155 & 0.259 & 87.8\% & 95.9\% & 13.79 & 92.4\% & 94.0\% & 0.691 & 0.407 & 39.89 \\
            STTM~\cite{liu2025see} & \textbf{0.333} & 0.334 & 95.7\% & 98.5\% & \underline{22.90} & 95.8\% & 95.7\% & \textbf{0.611} & \underline{0.338} & 45.88 \\
            MindTuner~\cite{gong2025mindtuner} & \underline{0.322} & \underline{0.421} & 95.8\% & \textbf{98.8\%} & 16.98 & 95.6\% & 93.8\% & \underline{0.612} & 0.340 & 44.63  \\
            BrainGuard~\cite{tian2025brainguard} & 0.313 & 0.330 & 94.7\% & 97.8\% & 21.76 & \underline{96.1\%} & \underline{96.4\%} & 0.624 & 0.353 & 45.43 \\
            \midrule
            \multicolumn{11}{c}{{EEG-Things ~\cite{grootswagers2022human}}} \\
            \midrule
            ATM~\cite{li2024visual} & {0.160} & {0.345} & \textbf{0.776}  & \textbf{0.866} & \textbf{17.60} & \textbf{0.734} & \textbf{0.786} & - & {0.582} & \textbf{30.55} \\
            CogCapturer~\cite{zhang2025cognitioncapturer} & \textbf{0.175} & \textbf{0.366} & 0.760 & 0.610 & 16.22 & 0.721 & 0.744 & - &  \textbf{0.577} & 28.64 \\ %
			\bottomrule
		\end{tabular}
	}
\end{table*}

\setlength{\tabcolsep}{4pt}
\setlength{\fboxrule}{0pt} 
\setlength{\fboxsep}{2pt}
\begin{table*}[h]
\centering
\caption{Quantitative evaulation for fMRI-to-video reconstruction on CC2017~\cite{wen2018neural}.}
\label{tab:supmat_cc2017}
\setlength{\tabcolsep}{4pt}
\resizebox{0.6\textwidth}{!}
{%
\begin{tabular}{lccccccc} 
\toprule
\multirow{2}{*}{\textsc{Method}} & \multicolumn{3}{c}{High-Level} & \multicolumn{2}{c}{Low-Level} \\
\cmidrule(r){2-4} \cmidrule(l){4-6} 
& 2-way & 50-way & BASIC-L & MMR & BASIC-H \\
\midrule
MinD-Video~\cite{chen2023cinematic} & 0.792 & 0.172 & 28.63 & 0.660 & 15.25 \\ 
NeuroClips~\cite{gong2024neuroclips} & \underline{0.808}  & \underline{0.195} & \textbf{45.12} & \underline{0.687} & \textbf{23.52} \\
DecoFuse~\cite{li2025decofuse} & \textbf{0.824} & \textbf{0.208} & \underline{29.03} & \textbf{0.706} & \underline{15.31} \\
\bottomrule
\end{tabular}
}
\end{table*}

\setlength{\tabcolsep}{4pt}
\setlength{\fboxrule}{0pt} 
\setlength{\fboxsep}{2pt}
\begin{table*}[t!]
\centering
\begin{minipage}{0.45\textwidth}
    \centering
    \caption{Quantitative evaluation for EEG-to-video reconstruction on SEED-DV~\cite{liu2024eeg2video}.}
    \label{tab:supmat_eeg2video}
    \setlength{\tabcolsep}{4pt} %
    \resizebox{1.0\textwidth}{!}{
    \begin{tabular}{lccccccc} 
    \toprule
    \multirow{2}{*}{\textsc{Method}} & \multicolumn{3}{c}{High-level} & \multicolumn{4}{c}{Low-level}  \\
    \cmidrule(r){2-4} \cmidrule(l){5-8} 
 & 2-way & 40-way & BASIC-H & 2-way & 40-way & SSIM & BASIC-L \\
    \midrule
    EEG2Video~\cite{liu2024eeg2video} & 0.798 & 0.159 & 49.54 & 0.774 & 0.138 & 0.256 & 20.54 \\
    \bottomrule
    \end{tabular}
    }
\end{minipage}%
\hfill
\begin{minipage}{0.53\textwidth}
    \centering
    \caption{Quantitative evaluation for EEG-to-3D reconstruction on EEG-3D~\cite{guo2025neuro3d}.}
    \label{tab:supmat_neuro3d}
    \setlength{\tabcolsep}{4pt} %
    \resizebox{1.0\textwidth}{!}{
    \begin{tabular}{lccccccc} 
    \toprule
    \multirow{2}{*}{\textsc{Method}} & \multicolumn{2}{c}{Average} & \multicolumn{2}{c}{Top-1 of 5 Samples} & \multicolumn{2}{c}{BASIC}  \\
    \cmidrule(r){2-3} \cmidrule(l){4-5} \cmidrule(l){6-7}
 & 2-way, Top-1 & 10-way, Top-3 & 2-way, Top-1 & 10-way, Top-3 & BASIC-L & BASIC-H \\
    \midrule
    Neuro-3D~\cite{guo2025neuro3d} & 55.81 & 35.89 & 72.08 & 57.64 & 9.69 & 23.08 \\
    \bottomrule
    \end{tabular}
    }
\end{minipage}
\end{table*}

\begin{table*}[t!]
\caption{Quantitative evaluation for fMRI-to-3D reconstruction on fMRI-3D~\cite{gao2024mind3d,gao2024mind3d++}.
}
\label{tab:supmat_mind3d}
\centering{
\setlength{\tabcolsep}{4pt}
\setlength{\fboxrule}{0pt} 
\setlength{\fboxsep}{2pt}
{
\resizebox{0.85\textwidth}{!}
{
    \begin{tabular}{l|c|cc|ccc|ccc|cc}
    \toprule
    \multicolumn{1}{c|}{\multirow{2}{*}{\textsc{Methods}}} & \multicolumn{1}{c|}{\multirow{2}{*}{\textsc{Dataset}}}  & \multicolumn{2}{c|}{Semantic-Level} & \multicolumn{3}{c|}{Structure-Level} & \multicolumn{3}{c}{Textural-Level} & \multicolumn{2}{c}{BASIC}\\
   & & 2-way$\uparrow$  &  10-way$\uparrow$   & FPD$\downarrow$ & CD$\downarrow$  & EMD$\downarrow$ & LPIPS$\downarrow$ &PSNR$\uparrow$ & SSIM$\uparrow$  & BASIC-L $\uparrow$ & BASIC-H $\uparrow$ \\
    \midrule
    MinD-3D~\cite{gao2024mind3d}   & \multirow{2}{*}{fMRI-Shape} & 0.828 & 0.459  & 3.157  & 1.742  & 3.833  &  0.306 & 32.81 & 0.674 & \textbf{14.72} & 30.95 \\ 
    {MinD-3D++}~\cite{gao2024mind3d++} & ~ & \textbf{0.887} & \textbf{0.616} & \textbf{3.025}  & \textbf{1.635} & \textbf{3.672} & \textbf{0.234} & \textbf{34.09} & \textbf{0.763}  & 12.62 & \textbf{35.08} \\
    \midrule
    MinD-3D~\cite{gao2024mind3d}   &\multirow{2}{*}{fMRI-Objaverse} &  0.793  & 0.427 & {4.304}  & {2.142} & {5.323} & {0.544} & 31.09 & 0.724 & N/A & N/A \\ 
    {MinD-3D++}~\cite{gao2024mind3d++}  & ~ & \textbf{0.894}  & \textbf{0.618}  & \textbf{3.325}  & \textbf{1.779} & \textbf{4.073} & \textbf{0.343} & \textbf{33.64} & \textbf{0.808} & N/A & N/A \\ 
    \bottomrule
    \end{tabular}}
    }
    }\\
    {\tiny N/A means reconstruction results on fMRI-Objaverse dataset are unavailable.}
\end{table*}

\section{Details on BASIC}
\label{sec:supmat_basic_implementation}

This section complements the method section in the main paer
with additional implementation details and qualitative results.

\subsection{Structured Entity Extraction from Captions} %
\label{subsec:supmat_basic_h}

\paragraph{Structured Descriptive Captioning.} We adopt LLaVA-1.6-13B~\cite{liu2023visual} as our default MLLM, considering the trade-off between computational efficiency and captioning accuracy. Recent studies on detailed captioning evaluation~\cite{lu2024benchmarking,dong2024benchmarking} suggest that this configuration achieves top-tier performance in long image description, comparable to human judgments. 
For prompting, we adopt a simple instruction—
``Describe the image in detail''—instead of crafting elaborate prompts that explicitly reflect evaluation dimensions outlined in~\cref{tab:evaluation_dim}. Preliminary experiments indicate that this simple prompt is sufficient to elicit detailed captions covering the desired evaluation dimensions when using LLaVA~\cite{liu2023visual} . This may be attributed to LLaVA’s training corpus, which likely includes similarly phrased instructions for detailed image captioning.
Further discussions on the qualitative and quantitative effects of different prompts and MLLMs are presented in the following sections. 

\paragraph{Semantic Parsing and Matching.} This step extracts and matches three core visual elements, objects, attributes, and relationships, from candidate captions. The extracted entities are then evaluated using precision, recall, and F1 score. To facilitate intuitive understanding, we further decompose this process and demonstrate each component in detail.
Consider the following two captions as the reference and candidate descriptions, respectively:
\begin{itemize}[leftmargin=17.5mm]
\setlength{\itemsep}{2pt}
\small{
\item[Reference] "A peaceful beach with soft white sand stretching along the coastline, where turquoise ocean waves gently roll onto the shore. Several people are sunbathing near the water while others are playing volleyball in the distance."
\item[Candidate] "A tropical island beach with lush palm trees swaying in the breeze, and the bright blue sea sparkling under the sun. Some people are lounging under umbrellas by the water, while a group of friends is playing volleyball near the edge of the beach."
}
\end{itemize}

The \objbox{objects}, \atrbox{attributes}, and \relbox{relationships} extracted from the captions are listed below. The attributes associated with the same object across different sentences, as well as spatial relationships - which are less intuitive and require further analysis to interpret the positional context - can also be extracted.

\begin{itemize}[leftmargin=17.5mm]
\setlength{\itemsep}{2pt}
\small{
\item[Reference] \objbox{objects}: volleyball, sand, water, beach, people, shore

\atrbox{attributes}: 
\objbox{beach}: peaceful; 
\objbox{sand}: soft, white; 
\objbox{people}: sunbathing;
\objbox{wave}: turquoise

\relbox{relations}: 
(\objbox{people}, play, \objbox{volleyball}); (\objbox{people}, sunbath, \objbox{water});
(\objbox{people}, play near, \objbox{water})

\item[Candidate] \objbox{objects}: water, people, island, beach, umbrella, sea

\atrbox{attributes}: 
\objbox{island}: tropical; 
\objbox{sea}: bright blue;
\objbox{palm tree}: lush

\relbox{relations}: 
(\objbox{umbrella}, by, \objbox{water}); 
(\objbox{people}, lounge under, \objbox{umbrella}); 
(\objbox{people}, on edge of, \objbox{beach}); 
(\objbox{beach}, on, \objbox{island}); (\objbox{people}, play, \objbox{volleyball}); (\objbox{sea}, under, \objbox{sun})

}
\end{itemize}

Given that substances extracted from ground truth and candidate captions often differ in wording, we apply a stepwise matching strategy: (i) exact matching, (ii) synonym matching, and (iii) cosine similarity of word embeddings to resolve remaining unmatched elements via semantic alignment. 

The evaluation scores for objects, attributes, and relations are computed as follows:
(a) \textit{Precision}, defined as the fraction of correctly predicted items among all items mentioned in the candidate description;
(b) \textit{Recall}, defined as the proportion of correctly predicted items among all items in the reference;
(c) \textit{F1} score, the harmonic mean of precision and recall.
\begin{equation}
  \text{Precision} = \frac{N(\text{Matched})}{N(\text{Candidate})}, \quad
\text{Recall} = \frac{N(\text{Matched})}{N(\text{Reference})}, \quad
\text{F1} = \frac{2 \cdot \text{Precision} \cdot \text{Recall}}{\text{Precision} + \text{Recall}},  
\end{equation}
where \( N(\text{Matched}) \) denotes the number of items in the candidate that correctly match those in the reference, \( N(\text{Candidate}) \) is the total number of items in the candidate, and \( N(\text{Reference}) \) is the total number of items in the reference.

\begin{figure*}[t]
\centering
\renewcommand{\arraystretch}{0.52}
\setkeys{Gin}{width=0.15\linewidth}
\setlength{\tabcolsep}{0.8pt}
\footnotesize
{
\resizebox{\textwidth}{!}
{
\begin{tabular}{cccccc}
\includegraphics{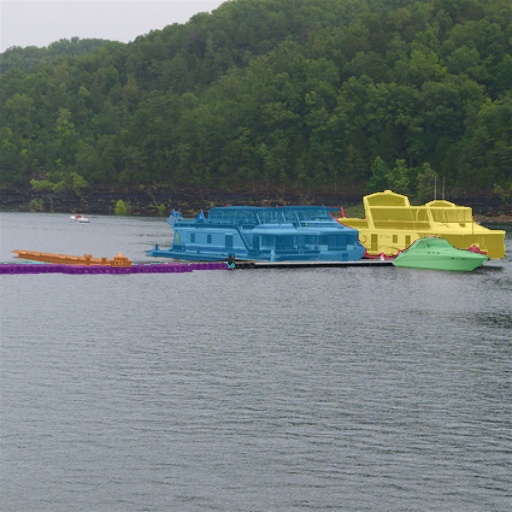} & 
\includegraphics{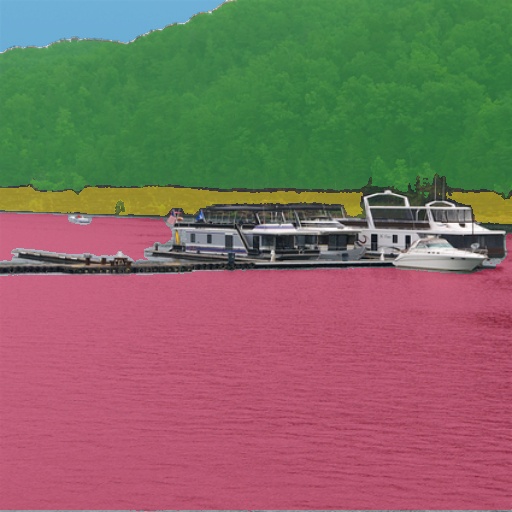} & 
\includegraphics{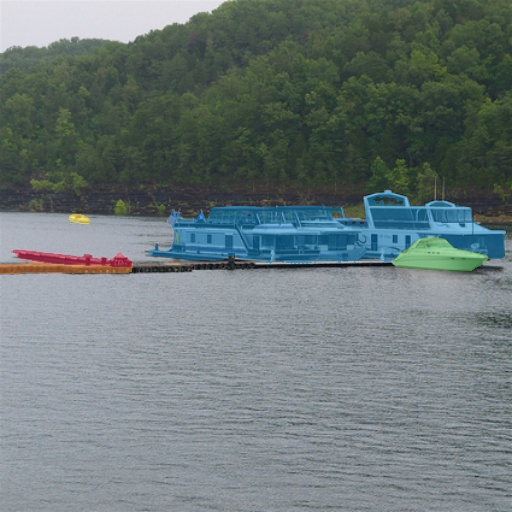} & 
\includegraphics{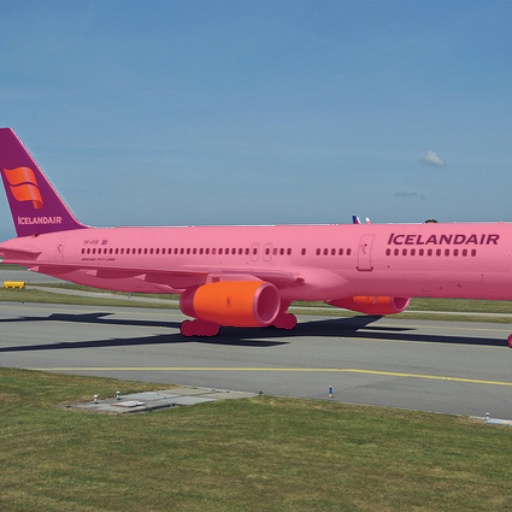} & 
\includegraphics{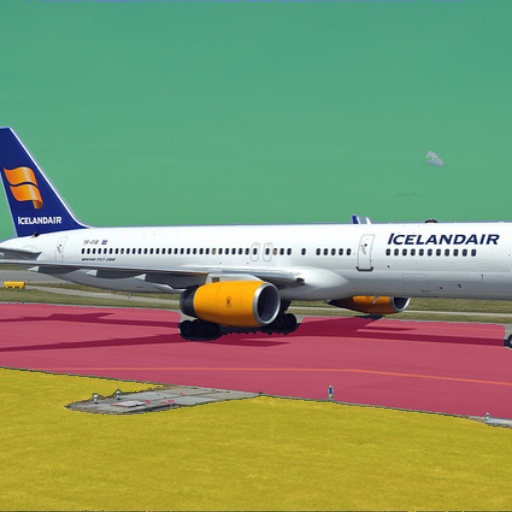} & 
\includegraphics{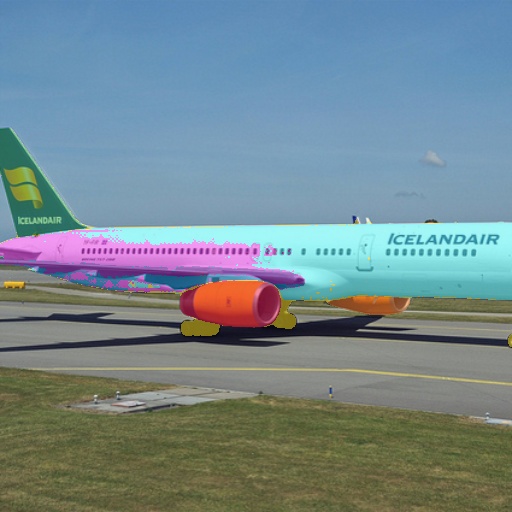}  \\
\includegraphics{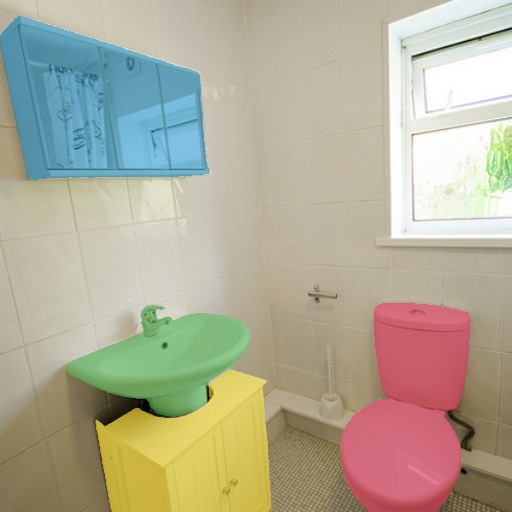} & 
\includegraphics{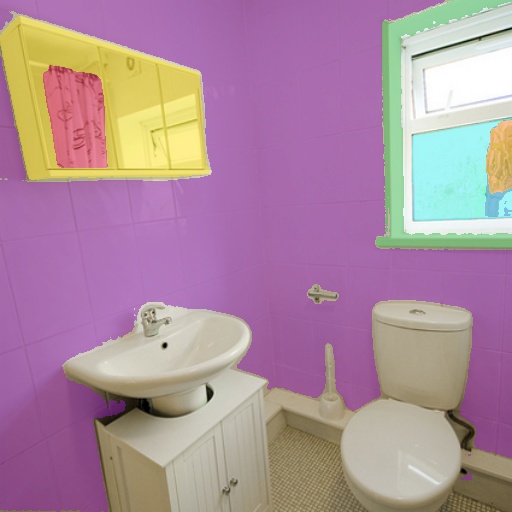} & 
\includegraphics{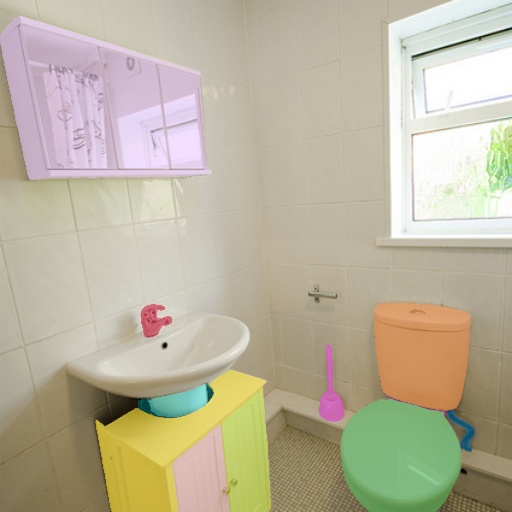} & 
\includegraphics{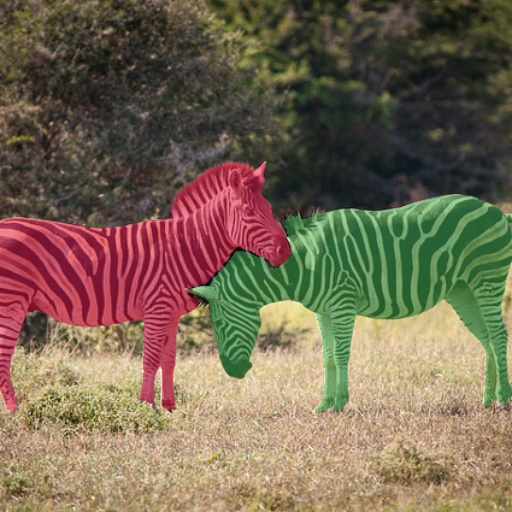} & 
\includegraphics{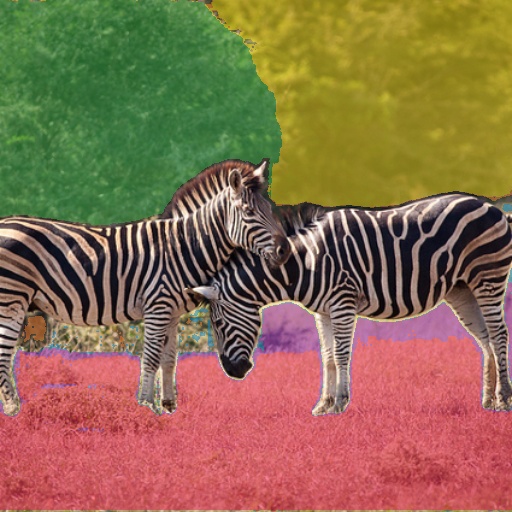} & 
\includegraphics{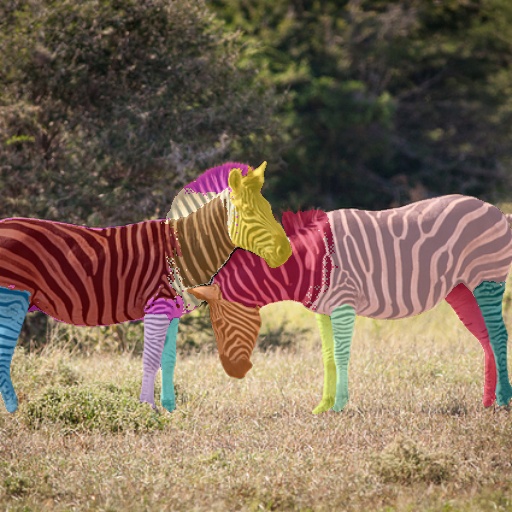}  \\
foreground & background & component & foreground & background & component \\
\end{tabular}
}
\caption{The visualization of multigranular segmentation. Reference images and structured annotations can be found in \cref{fig:supmat_nsd_recon} and  \cref{tab:supmat_sam_label}, respectively.
}
\label{fig:supmat_nsd_sam}
}
\end{figure*}

\subsection{Multigranular Image Segmentation}
\label{subsec:supmat_basic_l}

This section provides additional implementation details and qualitative results to complement the main paper.

\paragraph{Semantic Instance Categorization.}

Multigranular segmentation involves decomposing visual scenes into hierarchical levels of detail, from whole-object recognition to fine-grained part segmentation.  We first identify multigranular semantic instances that serve as textual instructions for subsequent segmentation using a multimodal large language model~\cite{liu2023visual}. The full categorization prompt used for  multigranular object identification
is provided in \cref{tab:supmat_sam_prompt}. 
As illustrated in \cref{tab:supmat_sam_label}, each image is annotated with structured information, including foreground objects, their constituent parts, and relevant contextual background elements. This hierarchical annotation schema allows for fine-to-coarse segmentation and supports comprehensive evaluation of vision models in terms of object-level recognition and detailed part-level understanding across diverse scene contexts. 

\paragraph{Progressive Granular Segmentation.}

Using the structured annotations described above as natural language prompts, we apply Grounded-SAM2~\cite{ren2024grounded} to produce multigranular segmentation masks. Prompts such as ``zebra'', ``tail'', or ``legs'' lead the model to localize and segment corresponding entities and component within an image. 
This enables 
model evaluation on fine-grained visual grounding, compositional understanding, and spatial reasoning. 
The resulting multigranular segmentation visualizations are shown in \cref{fig:supmat_nsd_sam}, with corresponding reference images displayed in \cref{fig:supmat_nsd_recon} and structured annotations provided in \cref{tab:supmat_sam_label}.
Taking the top-right three images of the airplane as an example, we present the salient foreground (airplane), background (sky, runway, grass, and trees), and the components of the airplane (fuselage, wings, engines, and tail), respectively.

The metrics are then calculated between the reconstruction and ground truth over a hierarchical structure: salient (foreground categories), binary (foreground and background), semantic (all distinct categories), instance (individual instances of all identified object categories), and part (sub-object components, mostly salient foreground objects). This structured evaluation facilitates systematic and fine-grained assessment of model performance across multiple levels of visual granularity.

Please note that \textit{semantic} and \textit{instance} in the context of BASIC-L evaluation are technical terms from the task of image segmentation, referring to semantic segmentation (labeling each pixel with a class) and instance segmentation (distinguishing individual object instances), and should not be confused with their broader conceptual meanings.

\begin{table*}[ht]
\centering
\caption{The \textit{categorization prompt} for multigranular object identification.
}
\label{tab:supmat_sam_prompt}
\begin{tcolorbox}[width=0.92\textwidth, colback=gray!5, colframe=black!40, boxrule=0.5pt, sharp corners]
\small
You are a visual understanding assistant. 
Given an input image, please analyze and describe it with a structured categorization. Each response should include:\\
1. Foreground Objects: List the main objects in the foreground and their semantic categories.\\
2. Background Elements: Describe elements in the background and their semantic categories.\\
3. Part-level Categories: For each foreground object, identify the visible parts in a general way (e.g., "car" $\rightarrow$ "wheels", "doors").\\
Provide your answer in a structured JSON format like this:
\begin{verbatim}
{
  "foreground_objects": [
    {
      "object": "dog",
      "semantic_category": "animal",
      "parts": ["head", "legs", "tail", "fur"]
    }
  ],
  "background_elements": [
    {
      "element": "tree",
      "semantic_category": "plant"
    },
    {
      "element": "sky",
      "semantic_category": "natural"
    }
  ]
}
\end{verbatim}
\end{tcolorbox}
\end{table*}

\setlength{\tabcolsep}{4pt}
\setlength{\fboxrule}{0pt} 
\setlength{\fboxsep}{2pt}
\begin{table*}[t!]
\centering
\caption{Structured scene annotations for four image instances. ID 0 to 3 corresponds to references in~\cref{fig:supmat_nsd_recon}. 
Each scene is decomposed into foreground (F) objects (e.g., boats, airplane, zebra), their associated part-level components, and background (B) elements (e.g., trees, sky, runway) that represent the environmental context. 
This is from a standardized JSON format.
}
\label{tab:supmat_sam_label}
\begin{tabular}{@{}ccp{12cm}@{}}
\toprule
\textbf{ID} & \textbf{Type} & \textbf{Details} \\
\midrule
\multirow{2}{*}{0} 
  & F & 
    \textbf{boats}: hull, deck, mast, sail; \textbf{dock}: pilings, decking, floating pontoons, fenders \\
  & B & water, trees, sky, shore \\
\midrule
\multirow{2}{*}{1} 
  & F & \textbf{airplane}: fuselage, wings, engines, tail \\
  & B & sky, runway, grass, trees \\
\midrule
\multirow{2}{*}{2} 
  & F & 
    \textbf{sink}: basin, faucet, pipes; %
    \textbf{toilet}: tank, bowl, seat, handle;
    \textbf{cabinet}: door, drawer, shelf \\
    & B & tiles, window, mirror, shower curtain, plants \\
\midrule
\multirow{2}{*}{3} 
  & F & 
    \textbf{zebra}: head, neck, torso, tail, legs \\
  & B & grass, trees, sky \\
\bottomrule
\end{tabular}
\end{table*}

\section{Qualitative Comparative Analysis}
\label{sec:supmat_qualitative_results}

This section presents additional qualitative results and analyses, including brain visual decoding reconstructions,
detailed captions for both reconstructions and stimuli, 
parsed caption annotations, 
and ablations on using different MLLMs and prompts.
The results are primarily based on methods applied to NSD~\cite{allen2022massive} for a comprehensive and in-depth examination.

\subsection{Brain Decoding Reconstruction}
\label{subsec:supmat_brain_reconstruction}

\cref{fig:supmat_nsd_recon} presents qualitative comparisons of brain visual decoding methods on the NSD dataset, including the original visual stimuli and reconstructions from SDRecon~\cite{takagi2023improving}, BrainDiffuser~\cite{ozcelik2023brain}, MindEye~\cite{scotti2023reconstructing}, DREAM~\cite{xia2024dream}, MindEye2~\cite{scotti2024mindeye2}, MindBridge~\cite{wang2024mindbridge}, UMBRAE~\cite{xia2024umbrae}, NeuroPictor~\cite{huo2024neuropictor}, NeuroVLA~\cite{shen2024neuro}, UniBrain~\cite{wang2024unibrain}, STTM~\cite{liu2025see}, MindTuner~\cite{gong2025mindtuner}, and BrainGuard~\cite{tian2025brainguard}.
These results encompass the majority of current brain visual decoding methods on NSD, providing a representative and comprehensive overview that reflects the latest advancements and trends in the field.

\begin{figure*}
\centering
\renewcommand{\arraystretch}{0.52}
\setkeys{Gin}{width=0.15\linewidth}
\setlength{\tabcolsep}{0.8pt}
\footnotesize
{
\resizebox{\textwidth}{!}
{
\begin{tabular}{ccccccc}
\includegraphics{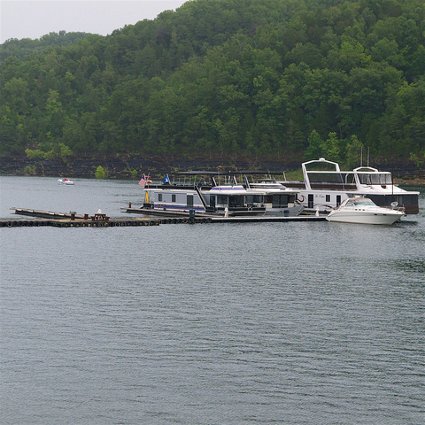} & 
\includegraphics{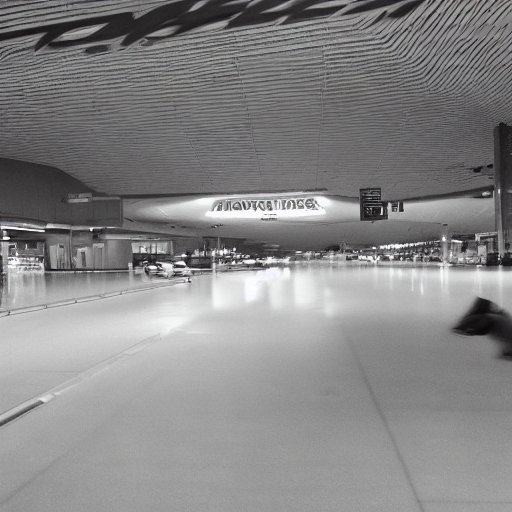} & 
\includegraphics{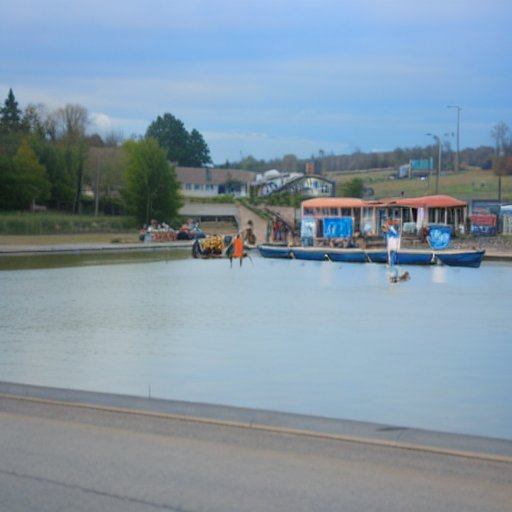} & 
\includegraphics{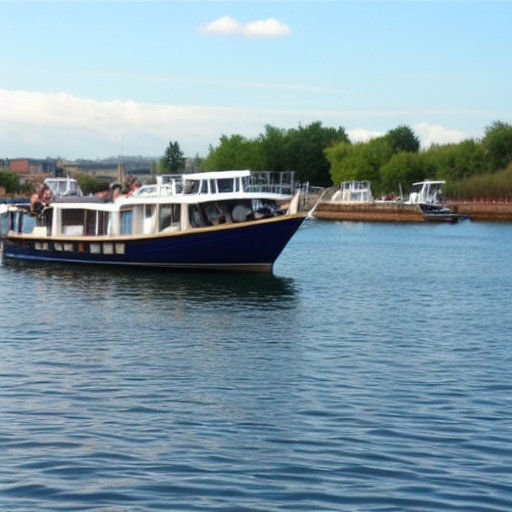} & 
\includegraphics{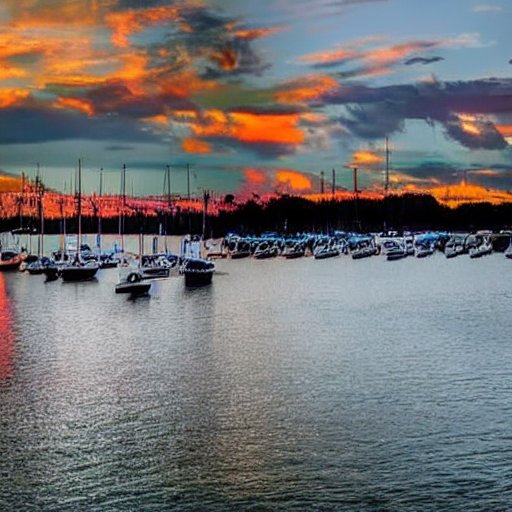} &
\includegraphics{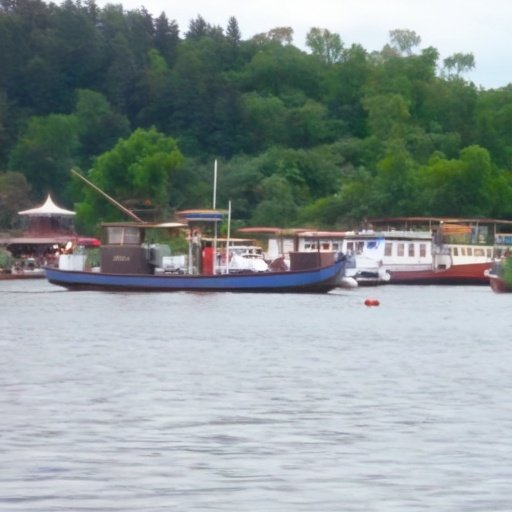} &
\includegraphics{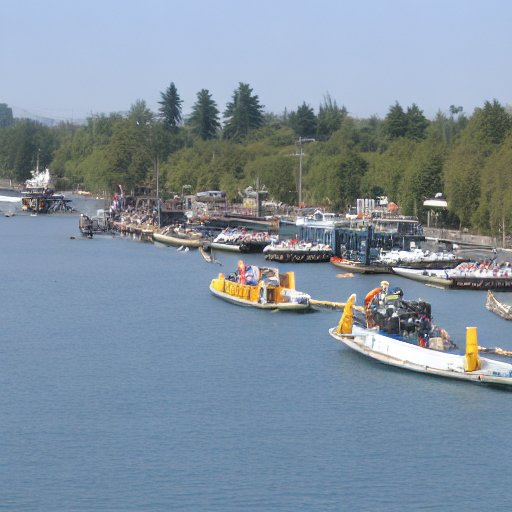}  \\
{Reference} & {SDRecon} & {BrainDiffuser} & {MindEye} & {DREAM} & {MindEye2} & {MindBridge} \\
\includegraphics{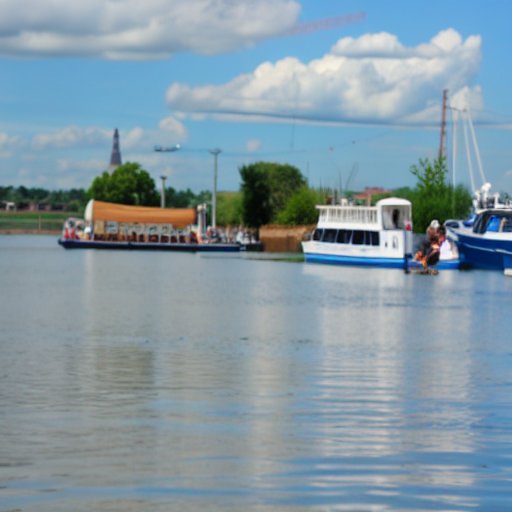}  & \includegraphics{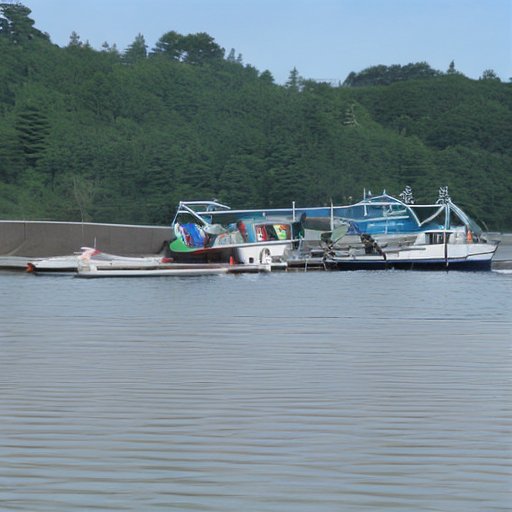} & 
\includegraphics{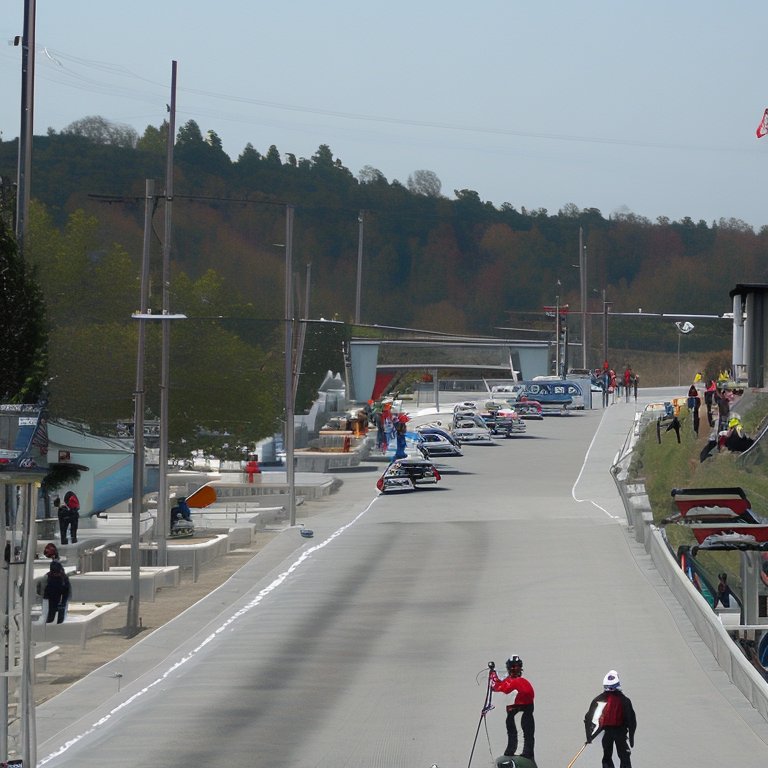} &
\includegraphics{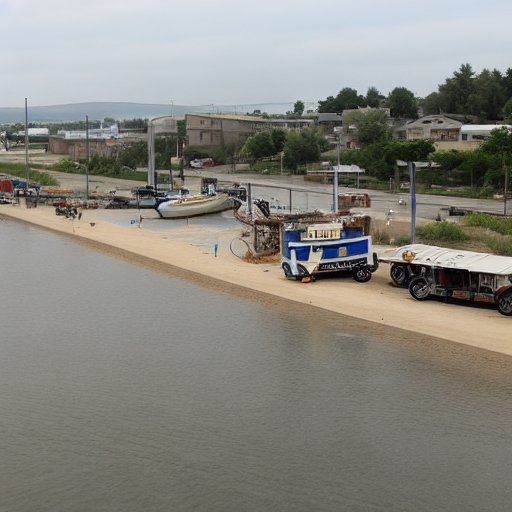} & 
\includegraphics{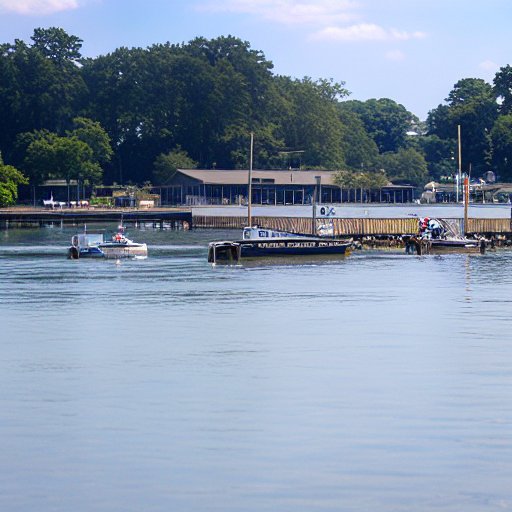}  & 
\includegraphics{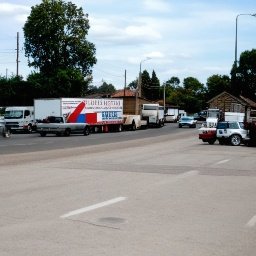} & 
\includegraphics{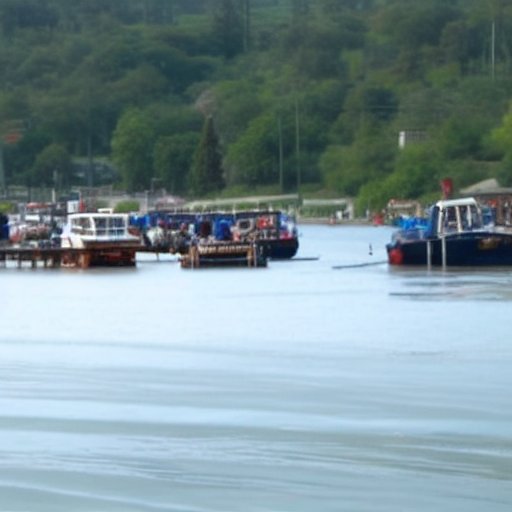} \\
{UMBRAE} & {NeuroPictor} & {NeuroVLA} & {UniBrain} & {STTM} & {MindTuner} & {BrainGuard} \\ [5pt]
\includegraphics{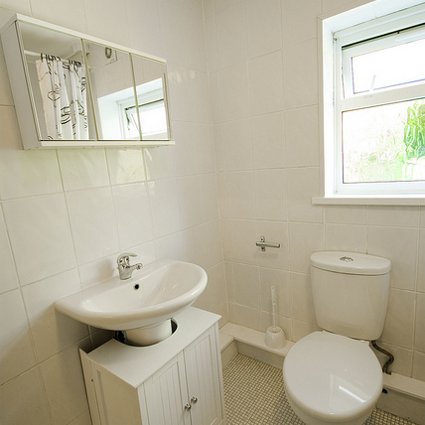} & 
\includegraphics{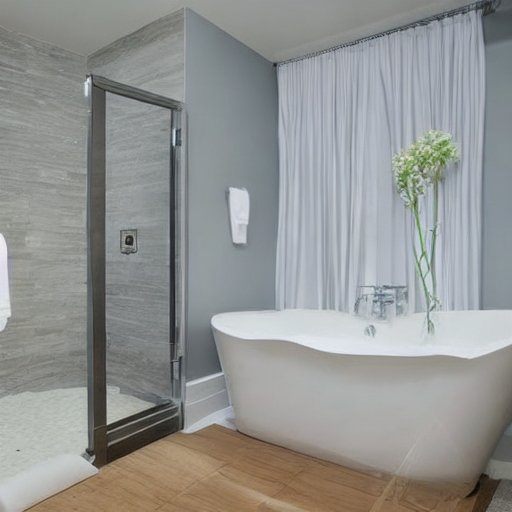} & 
\includegraphics{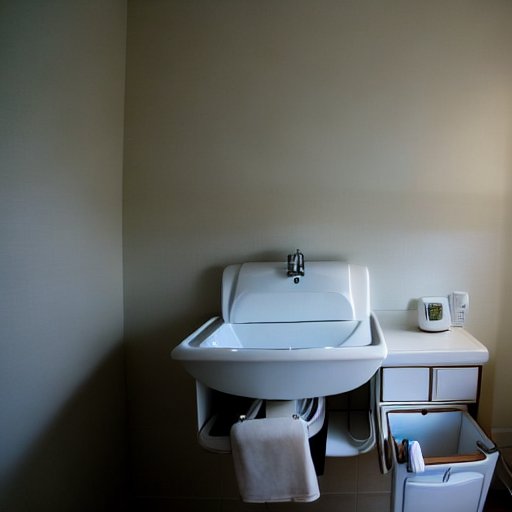} & 
\includegraphics{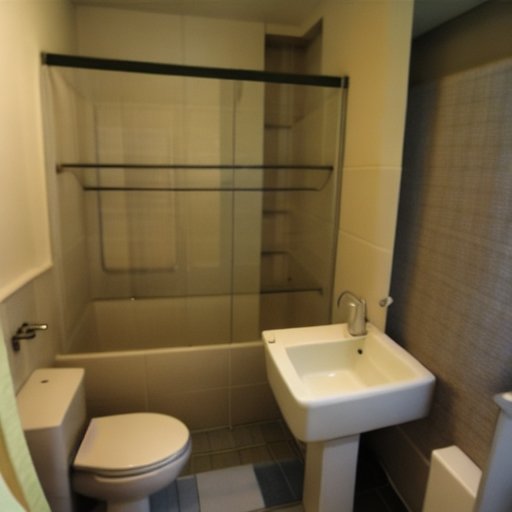} & 
\includegraphics{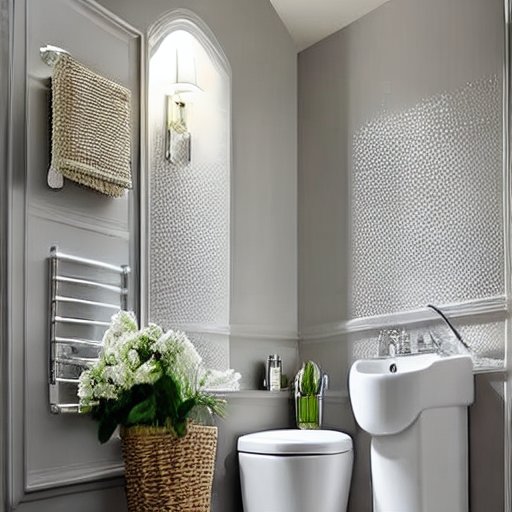} &
\includegraphics{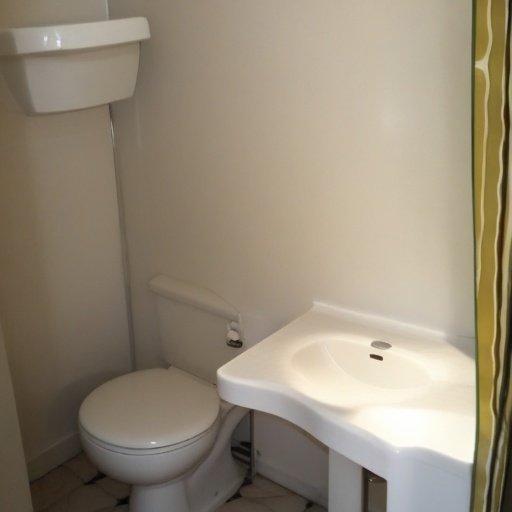} &
\includegraphics{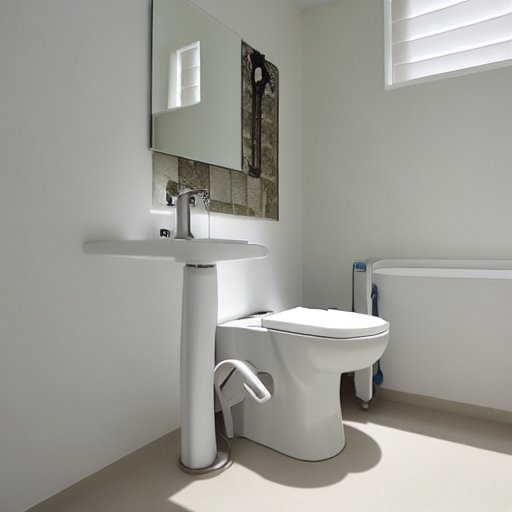}  \\
{Reference} & {SDRecon} & {BrainDiffuser} & {MindEye} & {DREAM} & {MindEye2} & {MindBridge} \\
\includegraphics{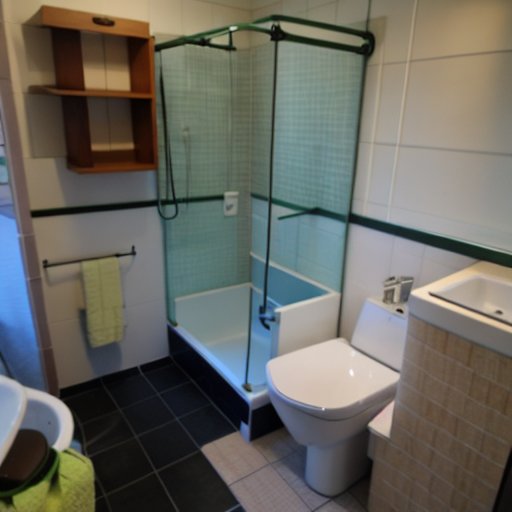} & \includegraphics{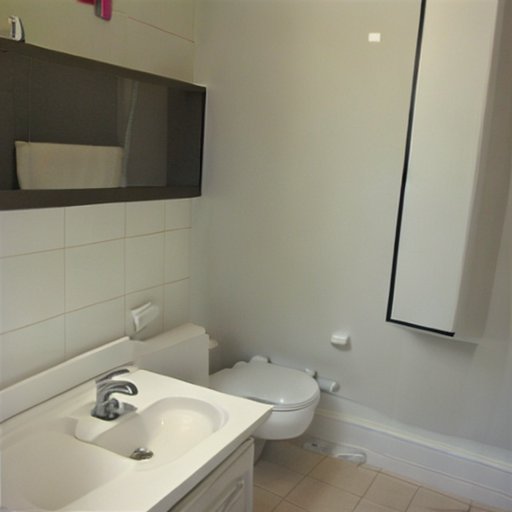} & 
\includegraphics{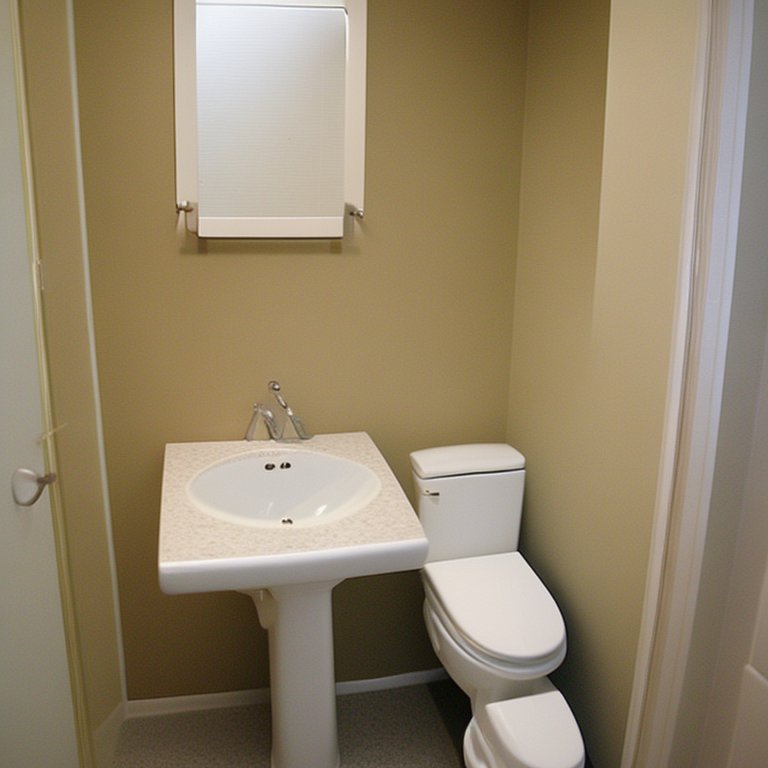} &
\includegraphics{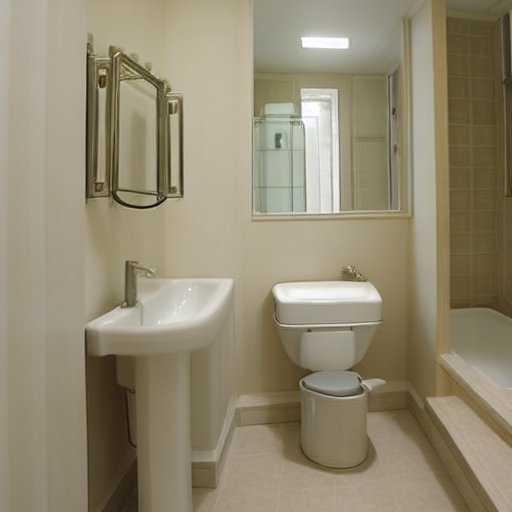} & 
\includegraphics{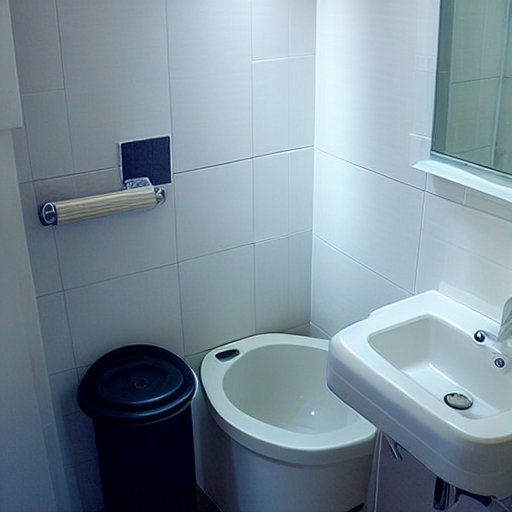}  & 
\includegraphics{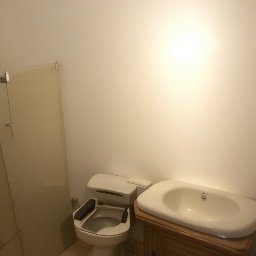} & 
\includegraphics{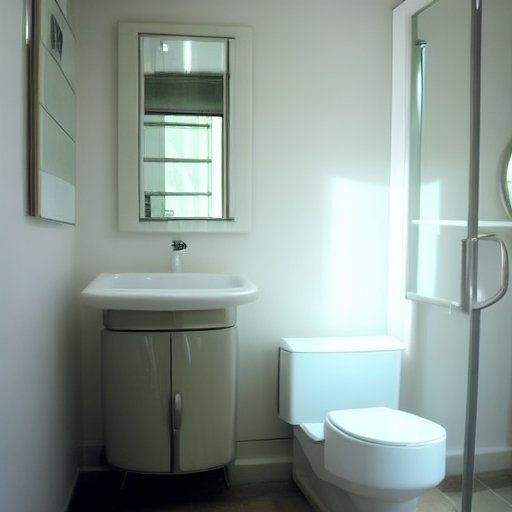} \\
{UMBRAE} & {NeuroPictor} & {NeuroVLA} & {UniBrain} & {STTM} & {MindTuner} & {BrainGuard} \\ [5pt]
\includegraphics{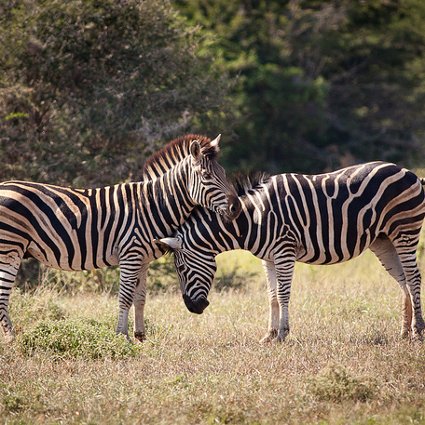} & 
\includegraphics{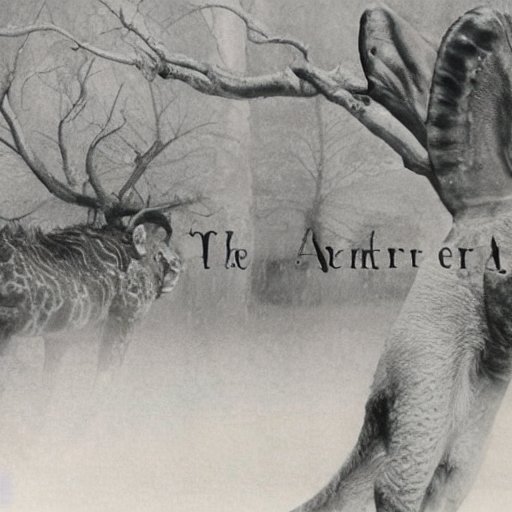} & 
\includegraphics{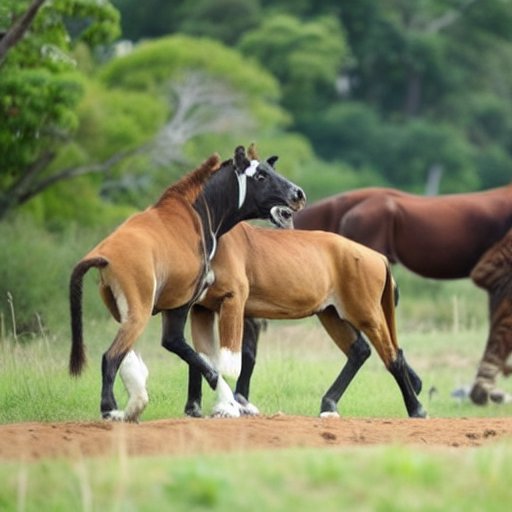} & 
\includegraphics{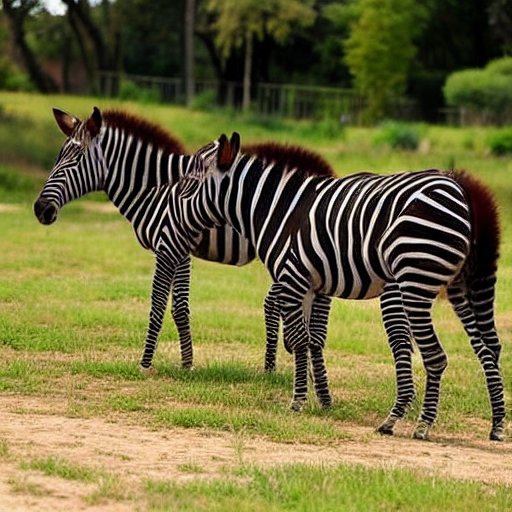} & 
\includegraphics{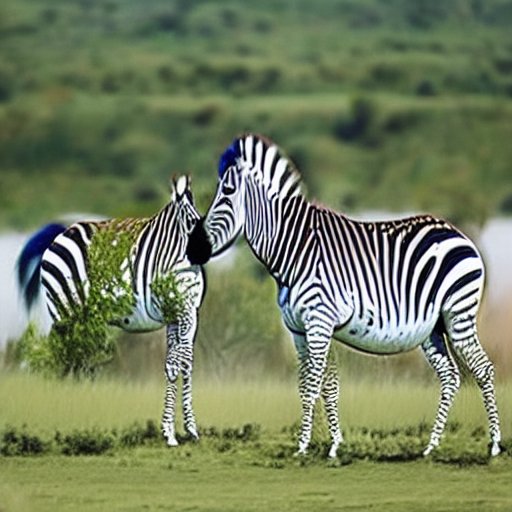} &
\includegraphics{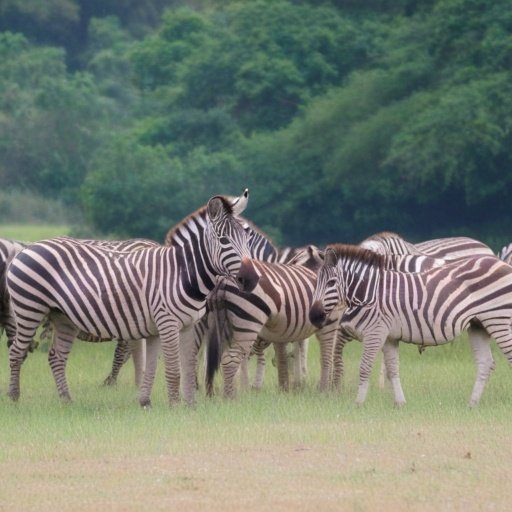} &
\includegraphics{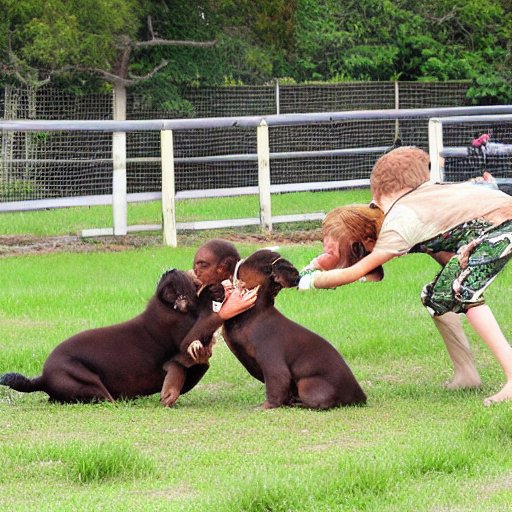} \\
{Reference} & {SDRecon} & {BrainDiffuser} & {MindEye} & {DREAM} & {MindEye2} & {MindBridge}\\
\includegraphics{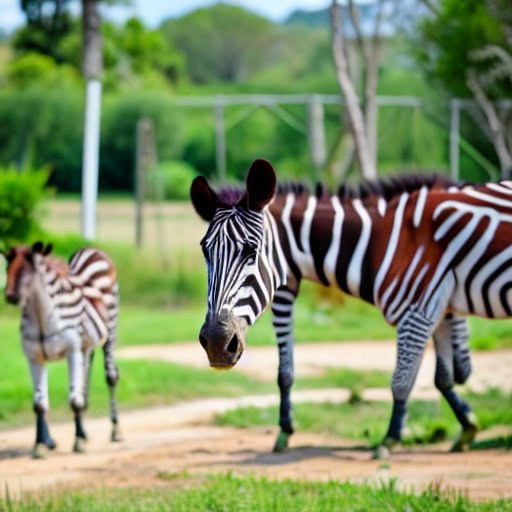} & \includegraphics{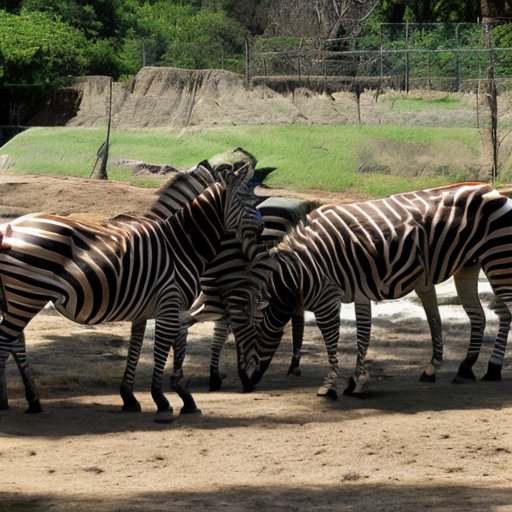} & 
\includegraphics{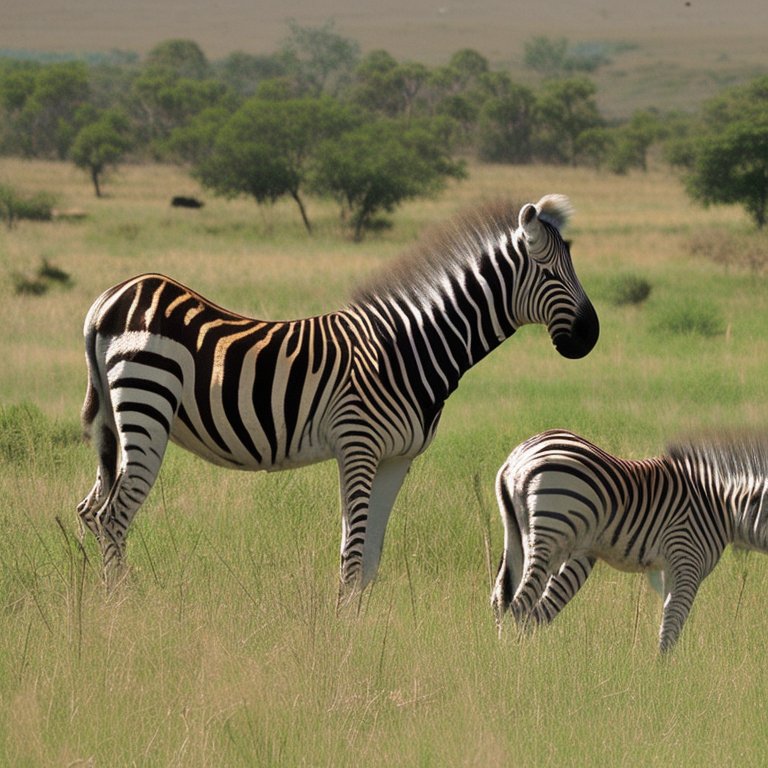} &
\includegraphics{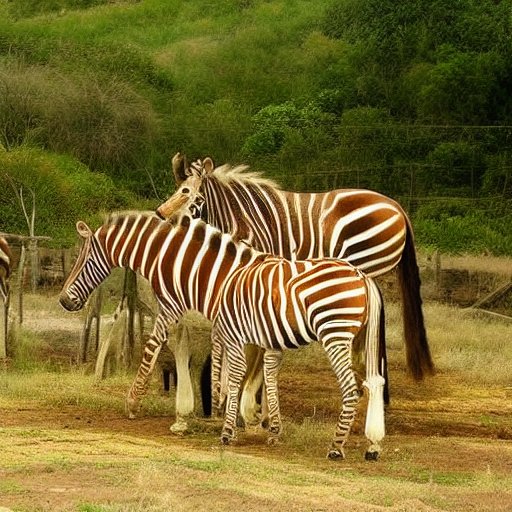} & 
\includegraphics{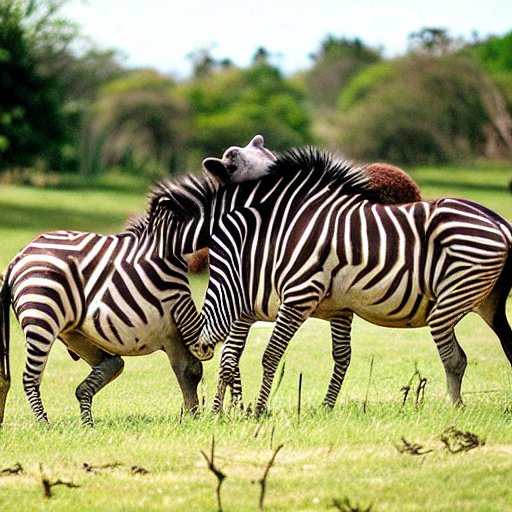}  & 
\includegraphics{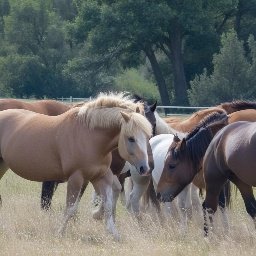} & 
\includegraphics{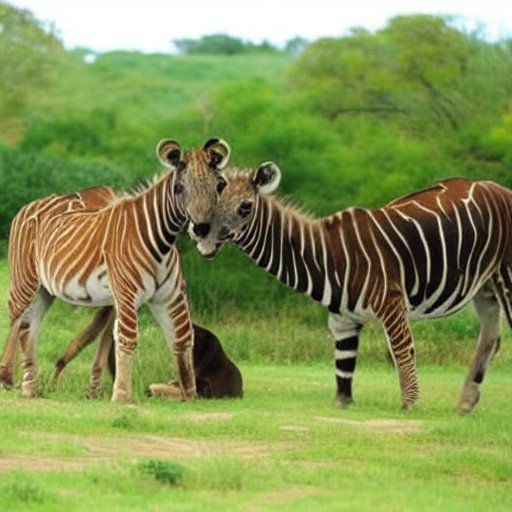} \\
{UMBRAE} & {NeuroPictor} & {NeuroVLA} & {UniBrain} & {STTM} & {MindTuner} & {BrainGuard} \\ [5pt]
\includegraphics{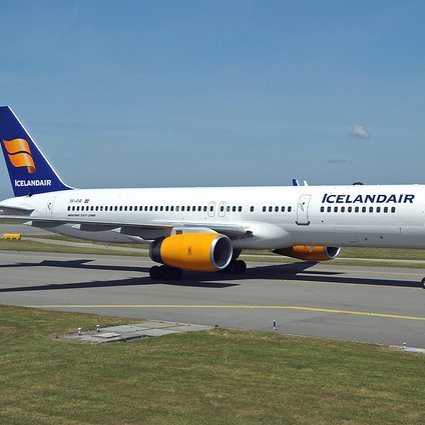} & 
\includegraphics{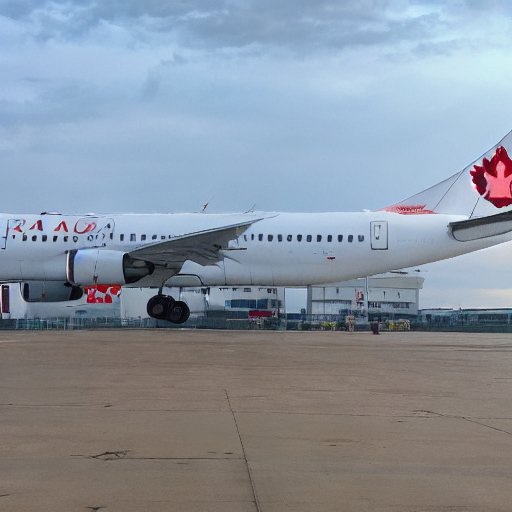} & 
\includegraphics{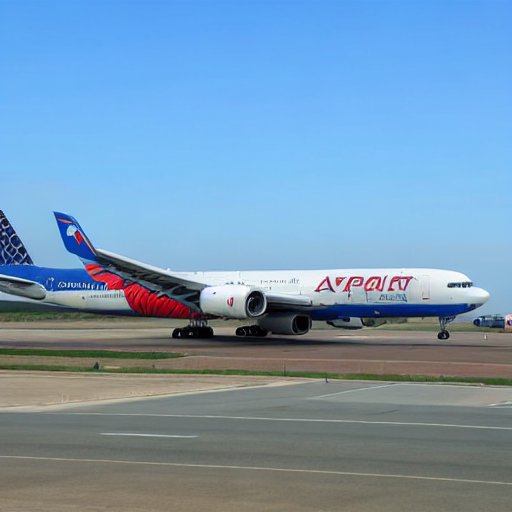} & 
\includegraphics{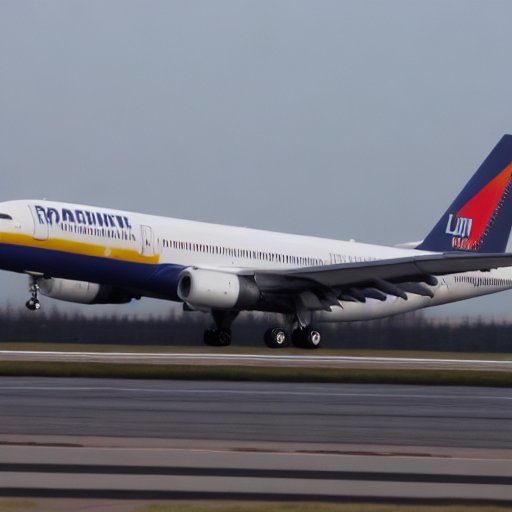} & 
\includegraphics{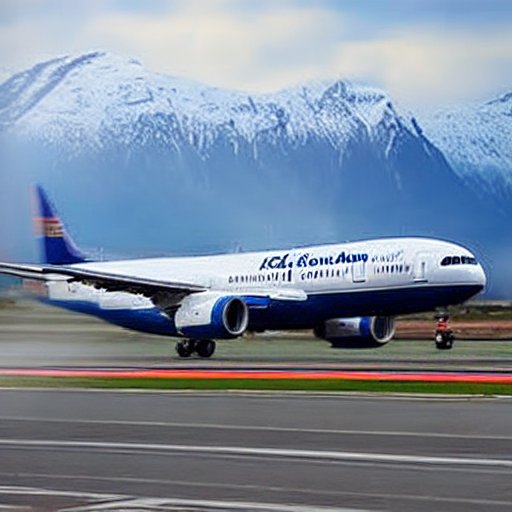} &
\includegraphics{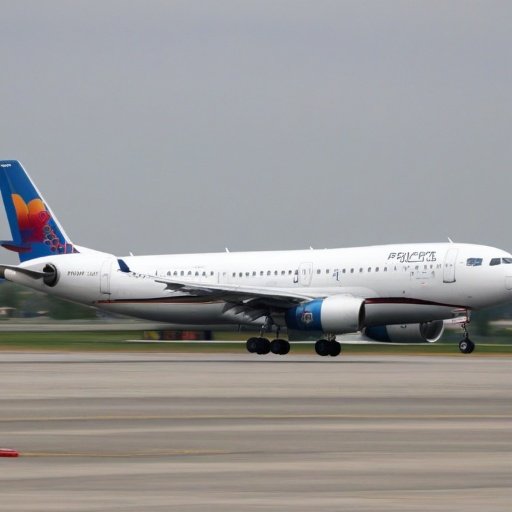} &
\includegraphics{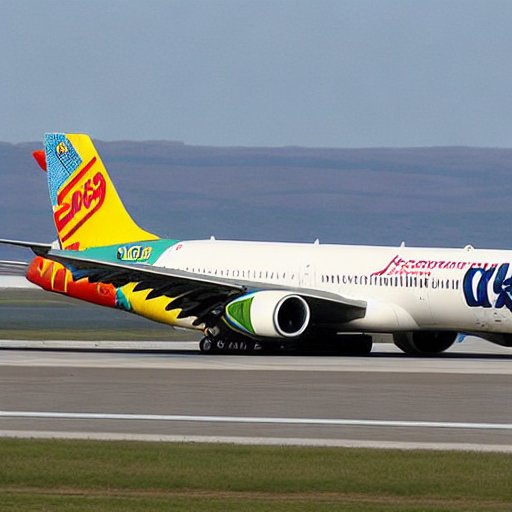} \\
{Reference} & {SDRecon} & {BrainDiffuser} & {MindEye} & {DREAM} & {MindEye2} & {MindBridge}\\
\includegraphics{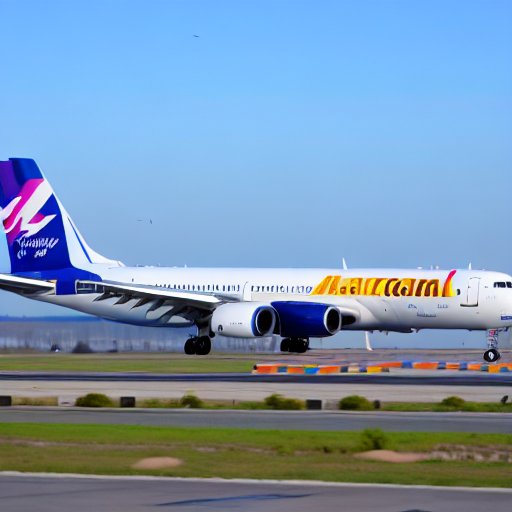} & \includegraphics{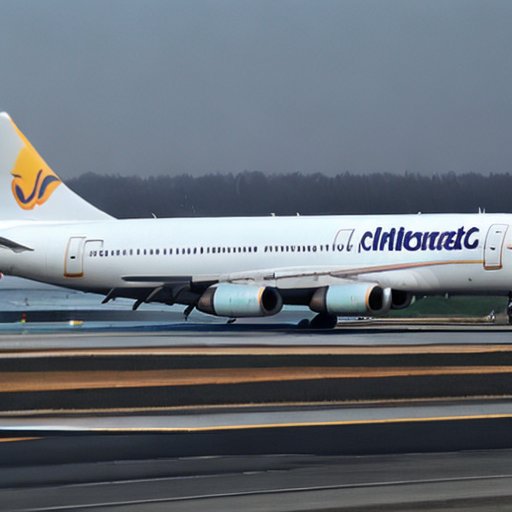} & 
\includegraphics{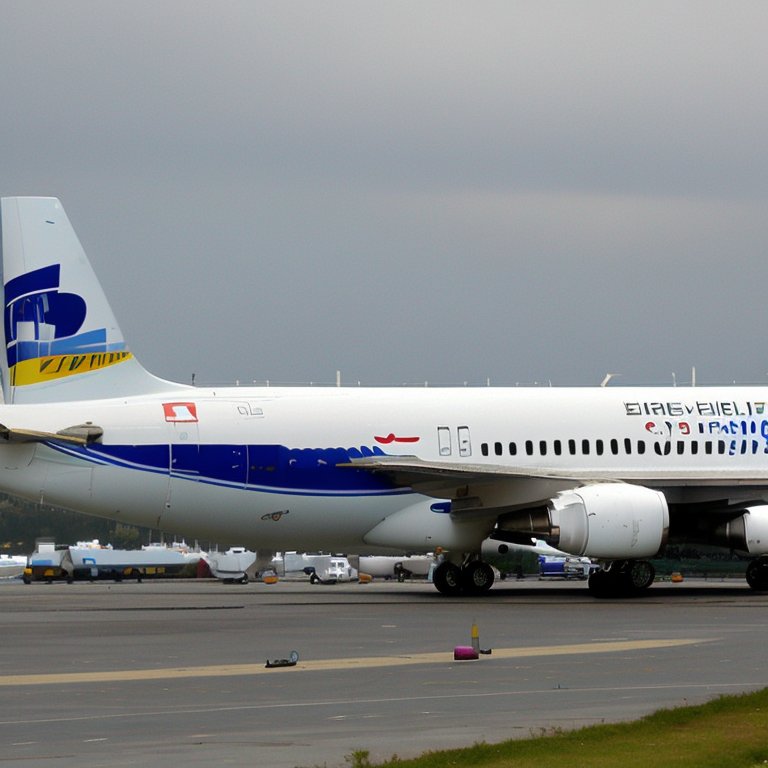} &
\includegraphics{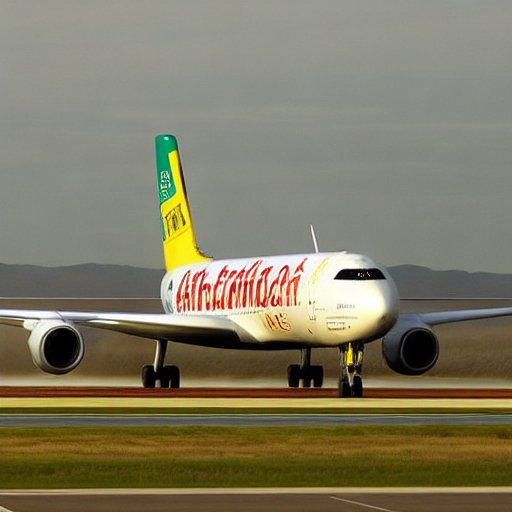} & 
\includegraphics{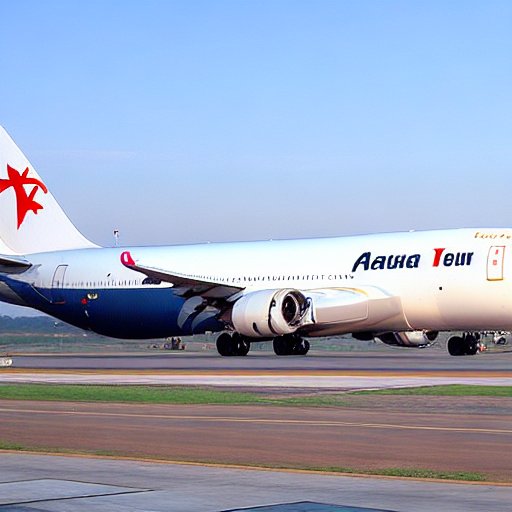}  & 
\includegraphics{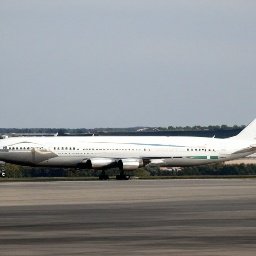} & 
\includegraphics{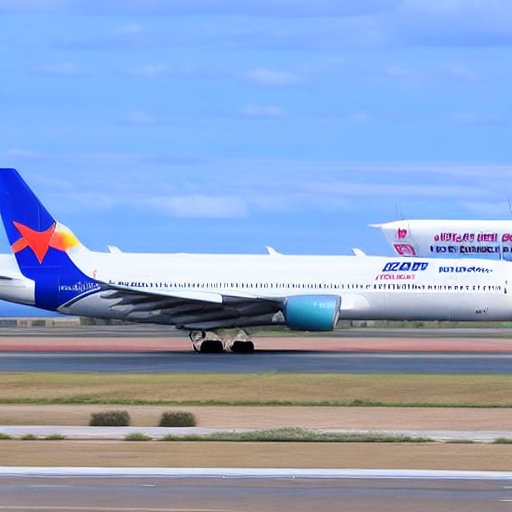} \\
{UMBRAE} & {NeuroPictor} & {NeuroVLA} & {UniBrain} & {STTM} & {MindTuner} & {BrainGuard} \\
\end{tabular}
}
\caption{Comparison of brain decoding reconstructions and corresponding references from NSD. 
}
\label{fig:supmat_nsd_recon}
}
\end{figure*}

\subsection{Detailed Captioning Comparison}
\label{subsec:supmat_detailed_caption}

Below is a comparison of detailed captioning results based on reconstructions generated by different brain decoding methods on NSD. The visual stimuli from the NSD test set, along with the corresponding decoded reconstructions for each image, are provided as reference in~\cref{fig:supmat_nsd_recon}. 
Due to page limitations, we only present the detailed captions for the first image here (``a boat at a lake dock'').

We highlight the \objbox{objects}, \atrbox{attributes}, and \relbox{relationships} extracted from the captions using the pipeline introduced in the main paper. These provide a structured and interpretable representation of the decoded contents. 
It should be noted that the annotations in the original captions shown below are intended for intuitive understanding and differ slightly from the parsed captions used for final evaluation, which are stored as structured data. 
The actual parsed results are below. %

\vspace{8pt}
\begin{itemize}[leftmargin=17.5mm]
\setlength{\itemsep}{2pt}
\item[\textbf{Method}] \textbf{Detailed Captioning}
\vspace{3pt}
\small{
\item[Reference] The image depicts a serene lakeside scene. In the \atrbox{natural} foreground, there is a \atrbox{calm} body of \objbox{water} \relbox{with} a few \objbox{ripples}, suggesting a gentle breeze. On the left side of the image, there is a \atrbox{wooden} \objbox{dock} \relbox{extend into} the \objbox{water}, with a \atrbox{small} \objbox{boat} moored at the end. The \objbox{dock} appears to be made of wooden planks and is \relbox{support by} wooden \atrbox{wooden} \objbox{pylons}.
In the middle of the image, there is a \atrbox{large} \objbox{boat} \relbox{dock at} a \atrbox{floating} \objbox{dock}. This \objbox{boat} is \atrbox{white} and has a \atrbox{blue} \objbox{stripe} \relbox{run along} its side. It has a covered area \relbox{with} a \objbox{roof}, and there are \objbox{windows} visible \relbox{on side of} the area, indicating it may be a cabin cruiser or a houseboat.
On the right side of the image, there is a \atrbox{grassy} \objbox{hillside} \relbox{with} \objbox{trees}, suggesting a wooded area. The \objbox{sky} is \atrbox{overcast}, with no visible sun or shadows, which gives the scene a \atrbox{diffused} light.
There are no visible {texts} {in} the image or distinctive brands. The style of the image is a \atrbox{straightforward}, \atrbox{color} \objbox{photograph}. The focus is on the \objbox{boats} and the surrounding natural environment, with no people or other significant \objbox{objects} in the immediate foreground.

\item[SDRecon] The image captures a bustling scene at an \objbox{terminal}. The \objbox{terminal} is characterized by a \atrbox{curved} \objbox{ceiling}, adorned with a pattern of \atrbox{diagonal} \objbox{lines} \relbox{on} the \objbox{ceiling} that add a sense of depth and dimension to the space. A \atrbox{large} \objbox{sign}, emblazoned with the \objbox{words} \relbox{on} the \objbox{sign}—"FLIGHTS" and "GATE"—\relbox{hang from} the \objbox{ceiling}, guiding travelers to their respective destinations.
The \objbox{terminal} is abuzz with activity. Numerous \objbox{people} \atrbox{walking} can be seen moving around, each engrossed in their own journey. Their movements create a \atrbox{dynamic} \objbox{tableau}, \relbox{add} life and \relbox{add} energy to the otherwise static environment.
The \objbox{terminal} is also home to a variety of \atrbox{scattered} \objbox{vehicles}, including cars and buses, which are \relbox{in} the \objbox{terminal} \relbox{alongside} the \objbox{people}. These \objbox{vehicles}, along with the \objbox{people}, contribute to the overall sense of movement and activity \relbox{in} the \objbox{terminal}.
Despite the monochrome palette of the image, the scene is rich in detail and texture. The interplay of light and shadow, the \atrbox{intricate} patterns \relbox{on} the \objbox{ceiling}, and the varied elements within the \objbox{terminal} all come together to create a \atrbox{vibrant}, \atrbox{engaging} image. The image is a testament to the constant motion and energy that characterizes life at an airport \objbox{terminal}.

\item[BrainDiffuser] The image is a {photograph} depicting a \atrbox{serene} scene \relbox{by} a body of \objbox{water}, which appears to be a lake or a \atrbox{calm} \objbox{river}. The \objbox{water} is reflecting the \atrbox{blue} of the \objbox{sky}, suggesting it is a \atrbox{clear} day. On the right side of the image, there is a \objbox{dock} or pier \relbox{extend into} the \objbox{water}, with several \objbox{boats} \relbox{moor along} its length. The \objbox{boats} vary in size and color, with one being particularly \atrbox{large} and \atrbox{blue}.
On the \objbox{dock}, there are a few \atrbox{visible} \objbox{people}, some of whom are \atrbox{standing} and others who might be seated or moving about. They \relbox{wear} \atrbox{casual} \objbox{attire}, and their presence adds a sense of activity to the otherwise tranquil setting.
In the background, there are \objbox{trees} and a \objbox{building} \relbox{with} a \atrbox{sloped} \objbox{roof}, which could be part of a \atrbox{residential} or waterfront \atrbox{area} \relbox{in} the background.
The \objbox{sky} is \atrbox{partly cloudy}, with patches of \atrbox{blue} visible, indicating fair weather conditions.

\item[MindEye] The image depicts a \atrbox{serene} \objbox{water} scene with a \atrbox{large} \atrbox{blue} and \atrbox{white} \objbox{boat} \relbox{float on} \objbox{calm} \objbox{water}. The \objbox{boat} has a \atrbox{white} \objbox{superstructure} \relbox{with} a \atrbox{blue} \objbox{hull} and is equipped with multiple levels of \objbox{seating} and a covered area \relbox{on} the upper deck. There are several \objbox{people} \relbox{on} the \objbox{boat}, suggesting it is in use. The \objbox{sky} is \atrbox{clear} \relbox{with} a few \objbox{clouds}, indicating fair weather. In the background, there are other \objbox{boats} \relbox{moor at} a \objbox{dock}, and the \objbox{shoreline} is \relbox{lined with} \objbox{greenery}. The style of the image is a \atrbox{realistic} \objbox{photograph} with natural colors and lighting. There are no visible texts or distinctive marks that provide additional context or information about the location or the boat's identity.

\item[DREAM] The image captures a \atrbox{serene} scene \relbox{at} a \objbox{marina} during sunset. The \objbox{sky}, painted in \atrbox{orange} and \atrbox{pink} \relbox{serve} as a beautiful backdrop to the \atrbox{tranquil} \objbox{water}. The \objbox{marina} is \relbox{filled with} numerous \atrbox{docked} and \atrbox{scattered} \objbox{boats} of various sizes, all bobbing gently \relbox{on} the \objbox{surface} of the \objbox{water}. The \objbox{boats} are scattered across the image, some closer to the foreground, while others are further away, creating a sense of depth. The \atrbox{calm} \objbox{water} \relbox{reflects} the vibrant colors of the \objbox{sky}, adding to the overall tranquility of the scene. The \objbox{image} \relbox{exudes} a sense of \objbox{peace} and \objbox{tranquility}, as if inviting the viewer to take a moment and appreciate the beauty of nature.

\item[MindEye2] The image depicts a \atrbox{serene} scene \relbox{on} a body of \objbox{water}, possibly a lake or a calm river. There are several \objbox{boats} visible, all of which appear to be \atrbox{small} and possibly used for recreational purposes. The \objbox{boats} are \atrbox{painted} in various colors, including shades of \atrbox{blue}, \atrbox{white}, and \atrbox{brown}.
In the foreground, there is a \objbox{boat} with a \atrbox{blue} \objbox{hull} and a \atrbox{white} \objbox{cabin}. This \objbox{boat} is closer to the viewer and is the most \atrbox{detailed} in the image. Behind it, there are two other \objbox{boats}. The one in the middle has a \atrbox{brown} \objbox{hull} and a \atrbox{white} \objbox{cabin}, while the one on the far right has a \atrbox{white} \objbox{hull} and a \atrbox{brown} \objbox{cabin} \relbox{on the right side of} the \objbox{hull}.
The \atrbox{calm}, \atrbox{still} \objbox{water} \relbox{reflects} the \objbox{light} in a way that suggests it might be a sunny day. The background is \relbox{filled with} \atrbox{lush} \objbox{greenery}, indicating that the location is likely near a forest or a heavily wooded area. There are no visible texts or distinctive markings that provide additional information about the location or the \objbox{boats}.
The style of the image is a \atrbox{standard}, \atrbox{non-stylized} \objbox{photograph} with a focus on the natural scenery and the \objbox{boats}. The perspective is from a distance, allowing for a view of the \objbox{boats} \relbox{in} the \objbox{photograph} and the surrounding environment. The image does not contain any people or animals, and the focus is on the stillness of the \objbox{water} and the tranquility of the setting.

\item[MindBridge] The image captures a \atrbox{serene} scene of a \objbox{river}, where several \objbox{boats} are \atrbox{docked}. The \objbox{river}, a \atrbox{deep shade}, \atrbox{blue}, is nestled amidst a backdrop of \atrbox{lush} \atrbox{green} \objbox{trees}. The \objbox{boats}, varying in size and color, are \relbox{docked along} the \objbox{riverbank}, their \objbox{hulls} \atrbox{glistening} under the sunlight.
In the foreground, a \atrbox{white} \objbox{boat} stands out, its \atrbox{yellow} \objbox{roof} adding a pop of color. Further down the \objbox{river}, a \atrbox{blue} \objbox{boat} \relbox{with} a \atrbox{white} \objbox{roof} can be seen, its vibrant hue contrasting with the \relbox{surrounding} \objbox{greenery}.
The \objbox{river} itself is dotted \relbox{with} several \objbox{buoys}, their \atrbox{bright} colors standing out against the \atrbox{blue} of the \objbox{water}. These \objbox{buoys}, \relbox{along} with the \atrbox{docked} \objbox{boats}, suggest a well-regulated \objbox{waterway}.
In the distance, beyond the \objbox{river}, a \objbox{city} \objbox{skyline} can be seen. The \objbox{city}'s \objbox{buildings}, though far away, add a touch of urban life to the otherwise natural landscape.
Overall, the image presents a harmonious blend of nature and urban life, with the \objbox{river} \relbox{serving as} a \atrbox{tranquil} \objbox{waterway} \relbox{amidst} the \objbox{city}'s bustling activity.

\item[UMBRAE] The image captures a \atrbox{serene} scene on a \objbox{river}, where two \objbox{boats} are gently \relbox{floating}. The \objbox{boat} on the left is a traditional \atrbox{wooden} \atrbox{brown} \objbox{boat}, painted in a \atrbox{warm} \atrbox{brown} color. It's adorned with a \atrbox{white} \objbox{roof}, \relbox{adding} a touch of \atrbox{rustic} \objbox{charm}.
\relbox{On the right}, there's a \atrbox{modern} \atrbox{blue} and \atrbox{white} \objbox{boat}, its \atrbox{sleek} \objbox{design} \relbox{contrasting} with the \atrbox{wooden} boat's \atrbox{rustic} appeal. The \objbox{boats} are not alone \relbox{in} the \objbox{water}; they are \relbox{accompanied of} a few \objbox{people} who are enjoying their time \relbox{on} the \objbox{river}.
In the distance, you can see a \atrbox{small} \objbox{island} with a \atrbox{tall} \objbox{lighthouse} \relbox{standing} tall. The \objbox{lighthouse}, a \atrbox{beacon} of \atrbox{safety} for sailors, adds a sense of depth and scale to the scene.
Above, the \objbox{sky} is a \atrbox{clear} \atrbox{blue}, dotted with \atrbox{fluffy} \atrbox{white} \objbox{clouds}. The \objbox{sun} is \atrbox{shining} \atrbox{brightly}, \relbox{casting} a \atrbox{soft} \objbox{glow} \relbox{on} the \objbox{water} and illuminating the scene with a \atrbox{soft} \objbox{light}.
Overall, the image paints a picture of a \atrbox{peaceful} \objbox{day} on the \objbox{river}, with \objbox{boats} gently \relbox{floating} and \objbox{people} enjoying their time \relbox{on} the \objbox{water}.

\item[NeuroPictor] The image depicts a \atrbox{serene} waterfront scene with two \objbox{boats} \relbox{docked at} a \objbox{pier}. The \objbox{boat} on the left is a \atrbox{white} \objbox{motorboat} with a \atrbox{blue} \objbox{stripe} \relbox{on side of} its side. It has a \atrbox{small} \objbox{cabin} and is equipped with a \atrbox{white} \atrbox{outboard} \objbox{motor}. The \objbox{boat} on the right is a \atrbox{large} vessel with a \atrbox{blue} \objbox{hull} and a \atrbox{white} \objbox{cabin}. It has a \atrbox{white} \atrbox{outboard} \objbox{motor} as well. Both \objbox{boats} are \relbox{secure with} \objbox{ropes}.
The \objbox{pier} itself is constructed of \atrbox{wood} and is situated in front of a backdrop of \atrbox{lush} \atrbox{green} \objbox{trees}. The \objbox{water} is \atrbox{calm}, reflecting the \atrbox{clear} \atrbox{blue} \objbox{sky} above. There are no visible \objbox{markings} \relbox{on} the \objbox{boats} or the \objbox{pier} that provide additional information. The style of the image is a \atrbox{realistic} \objbox{photograph} with \atrbox{natural} \objbox{lighting} and colors.

\item[NeuroVLA] The image captures a \atrbox{vibrant} \objbox{scene} at a \objbox{race track}. The \objbox{track}, a \atrbox{smooth} gray surface, is \relbox{bordered by} a \atrbox{grassy} \objbox{area} \relbox{on} one \relbox{side} and a line of \objbox{trees} on the other. A group of \objbox{people}, dressed in a variety of colors, are scattered around the \objbox{track}, some standing, others \relbox{walk around} the \objbox{track}.

On the \objbox{track}, two \objbox{individuals} are engaged in a race. One of them is leading, their \objbox{body} \atrbox{leaning} forward in anticipation. The other \objbox{racer} is close behind, their body language suggesting a fierce competition.
In the background, a \atrbox{blue} \objbox{building} \relbox{stand against} the \objbox{greenery}. A \atrbox{red} and \atrbox{white} \atrbox{striped} \objbox{awning} extends from it, providing a splash of color to the scene. The \objbox{building}'s location suggests it might be a facility \relbox{related to} the \objbox{race track}.
The image is a \atrbox{dynamic} \atrbox{snapshot} of a moment, \relbox{fill with} \objbox{color}. It's a testament to the excitement and energy of racing.

\item[UniBrain] The image depicts a coastal scene with a \atrbox{sandy} \objbox{beach} leading up to a body of water that appears to be a river or a lake. \relbox{On} the \objbox{beach}, there are several \objbox{boats} of different sizes, including a \atrbox{small} \objbox{motorboat} and a \atrbox{large} \objbox{sailboat}. The \objbox{boats} are \atrbox{moored} and appear to be unoccupied. A few \objbox{people} are \atrbox{visible}, but they are too \atrbox{far away} to discern any specific details.
\relbox{On the right side} of the \objbox{image}, there is a \objbox{road} or \objbox{path} \relbox{run parallel to} the water's \objbox{edge}. Along this path, there are various \objbox{vehicles} parked, including a \atrbox{large} \objbox{truck} with a \objbox{trailer} attached and a \atrbox{smaller} \objbox{truck}. The \objbox{vehicles} are \relbox{park in} a \objbox{line}, suggesting they might be used for transporting goods or equipment.
The \objbox{sky} is \atrbox{overcast}, and the \objbox{lighting} in the image is \atrbox{soft}, indicating either early morning or late afternoon. The \objbox{vegetation} \relbox{along} the water's \relbox{edge} includes a mix of grasses and \atrbox{small} \objbox{trees}. There are no visible texts or distinctive brands in the image. The style of the image is a \atrbox{standard}, \atrbox{non-stylized} \objbox{photograph} with a focus on the \atrbox{natural} and \atrbox{man-made} \objbox{elements} present in the scene.

\item[STTM] The image captures a \atrbox{serene} \objbox{scene} at a \objbox{marina}. Dominating the foreground is a large body of \objbox{water}, its \objbox{surface} \atrbox{rippling} gently. A \objbox{blue} \objbox{boat}, adorned with a \atrbox{white} \objbox{stripe}, is seen \relbox{glid across} the \objbox{water}, heading towards the right side of the image.
On the left side of the image, a \atrbox{wooden} \objbox{pier} \relbox{extend into} the \objbox{water}. It's a \objbox{hub} of activity with several \objbox{boats} \relbox{dock at} it. Among them, a \atrbox{white} \objbox{boat} stands out, its vibrant color \relbox{contrast with} the surrounding structures.
\relbox{In} the \objbox{background}, a cluster of \objbox{trees} can be seen, their \atrbox{lush} \atrbox{green} \objbox{foliage} adding a touch of nature to the scene. Above it all, the \objbox{sky} stretches out in a \atrbox{clear} \atrbox{blue} expanse, devoid of any clouds. The image beautifully encapsulates a day at the \objbox{marina}, \relbox{filled with} the \atrbox{tranquility} of the \objbox{water} and the bustle of boating \objbox{activity}.

\item[MindTuner] The image shows a \atrbox{large}, \atrbox{white} \objbox{boat} with a \atrbox{blue} \objbox{roof} and a \atrbox{red} \atrbox{stripe} \relbox{on the side}, \relbox{floating on} a body of \objbox{water}. The \objbox{boat} has multiple levels and appears to be a \objbox{passenger} or tourist vessel, as suggested by the presence of \objbox{windows} and what looks like a \atrbox{covered} \objbox{area} \relbox{on} the \atrbox{upper} \objbox{deck}. The \objbox{boat} is equipped with a \objbox{mast} and \objbox{sails}, indicating that it may be capable of sailing. The surrounding environment includes a \atrbox{forested} \objbox{area} with \atrbox{lush} \atrbox{green} \objbox{trees}, and the \objbox{water} reflects the \atrbox{overcast} \objbox{sky} above. 
The style of the image is a \atrbox{realistic} \objbox{photograph} with \atrbox{natural} \objbox{lighting} and colors.

\item[BrainGuard] The image depicts a serene \objbox{harbor} scene. There are several \objbox{boats} of \atrbox{varying sizes} and \atrbox{colors} \relbox{moored at} the \objbox{dock}. The \objbox{water} is \atrbox{calm}, reflecting the \atrbox{clear} \atrbox{blue} \objbox{sky} above. The \objbox{dock} is \relbox{made of} \atrbox{wood} and \relbox{extends into} the \objbox{water}, providing a place for the \objbox{boats} to rest. In the distance, you can see a \atrbox{lush} \atrbox{green} \objbox{hillside}.
The \objbox{boats} are securely \relbox{tied to} the \objbox{dock}, suggesting that they are not in motion. The overall \objbox{atmosphere} of the image is \atrbox{peaceful} and \atrbox{idyllic}.
}
\end{itemize}

\subsection{Parsed Descriptions}
\label{subsec:supmat_parsed_caption}

The parsed captions are stored in a structured dictionary format, containing information about \objbox{objects}, \atrbox{attributes}, and \relbox{relationships}. The identified objects are stored in a list, with attributes mapped to each object through key-value pairs. Relationships between objects are encoded in a directional scene graph, where edges capture the directed interactions or spatial relationships.%

\vspace{8pt}
\begin{itemize}[leftmargin=17.5mm]
\setlength{\itemsep}{2pt}
\item[\textbf{Method}] \textbf{Parsed Description}
\vspace{3pt}
\small{
\item[Reference] \objbox{objects}: boat, water, pylon, roof, ripple, window, stripe, tree, dock, sky, hillside

\atrbox{attributes}: \objbox{water}: calm; \objbox{boat}: large, white, small; \objbox{dock}: wooden, floating; \objbox{pylon}: wooden; \objbox{stripe}: blue; \objbox{area}: covered; \objbox{hillside}: grassy; \objbox{sky}: overcast, diffused; \objbox{photograph}: color, straightforward; \objbox{foreground}: natural 

\relbox{relations}: (\objbox{window}, on side of, \objbox{area}); (\objbox{dock}, extend into, \objbox{water}); (\objbox{boat}, dock at, \objbox{dock}); (\objbox{hillside}, with, \objbox{tree}); (\objbox{boat}, in, \objbox{foreground}); (\objbox{water}, with, \objbox{ripple}); (\objbox{area}, with, \objbox{roof}); (\objbox{stripe}, run along, \objbox{boat}); (\objbox{dock}, support by, \objbox{pylon})

\item[SDRecon] \objbox{objects}: line, vehicle, word, terminal, people, ceiling, sign

\atrbox{attributes}: \objbox{ceiling}: curved; \objbox{line}: diagonal; \objbox{sign}: large; \objbox{people}: walking; \objbox{tableau}: dynamic; \objbox{vehicle}: scattered; \objbox{pattern}: intricate; \objbox{terminal}: vibrant, engaging

\relbox{relations}: (\objbox{word}, on, \objbox{sign}); (\objbox{tableau}, add, \objbox{energy}); (\objbox{vehicle}, alongside, \objbox{people}); (\objbox{tableau}, add, \objbox{life}); (\objbox{vehicle}, in, \objbox{terminal}); (\objbox{terminal}, have, \objbox{ceiling}); (\objbox{scene}, have, \objbox{detail}); (\objbox{sign}, hang from, \objbox{ceiling}); (\objbox{pattern}, on, \objbox{ceiling}); (\objbox{scene}, have, \objbox{texture}); (\objbox{terminal}, have, \objbox{vehicle})

\item[BrainDiffuser] \objbox{objects}: water, building, people, boat, dock, river, sky, roof

\atrbox{attributes}: \objbox{river}: calm; \objbox{scene}: serene; \objbox{sky}: blue, partly cloudy, clear; \objbox{boat}: blue, large; \objbox{people}: visible, standing; \objbox{attire}: casual; \objbox{area}: waterfront, part, residential; \objbox{roof}: sloped; \objbox{photograph}: naturalistic; \objbox{element}: human-made; \objbox{environment}: natural

\relbox{relations}: (\objbox{people}, wear, \objbox{attire}); (\objbox{boat}, moor along, \objbox{water}); (\objbox{building}, in, \objbox{background}); (\objbox{dock}, extend into, \objbox{water}); (\objbox{scene}, by, \objbox{water}); (\objbox{people}, on, \objbox{dock}); (\objbox{building}, with, \objbox{roof})

\item[MindEye] \objbox{objects}: sky, shoreline, water, boat, superstructure, people, cloud, dock, hull, seating, greenery

\atrbox{attributes}: \objbox{boat}: blue, large, white; \objbox{water}: calm; \objbox{hull}: blue; \objbox{superstructure}: white; \objbox{sky}: clear; \objbox{photograph}: realistic

\relbox{relations}: (\objbox{superstructure}, with, \objbox{hull}); (\objbox{boat}, have, \objbox{seating}); (\objbox{boat}, moor at, \objbox{dock}); (\objbox{people}, on, \objbox{boat}); (\objbox{sky}, with, \objbox{cloud}); (\objbox{greenery}, line, \objbox{shoreline}); (\objbox{boat}, have, \objbox{hull}); (\objbox{boat}, float on, \objbox{water})

\item[DREAM] \objbox{objects}: boat, water, marina, peace, sky

\atrbox{attributes}: \objbox{scene}: serene; \objbox{sky}: vibrant, pink, orange; \objbox{water}: tranquil, calm; \objbox{boat}: docked, scattered

\relbox{relations}: (\objbox{image}, exude, \objbox{tranquility}); (\objbox{boat}, on, \objbox{surface}); (\objbox{image}, exude, \objbox{peace}); (\objbox{sky}, serve, \objbox{water}); (\objbox{marina}, fill with, \objbox{boat}); (\objbox{sky}, reflect in, \objbox{water}); (\objbox{water}, have, \objbox{surface}); (\objbox{scene}, at, \objbox{marina})

\item[MindEye2] \objbox{objects}: boat, water, cabin, hull, light, greenery

\atrbox{attributes}: \objbox{scene}: serene; \objbox{boat}: detailed, blue, small, white, painted; \objbox{hull}: blue, white, brown; \objbox{cabin}: white, brown; \objbox{water}: still, calm, tranquil; \objbox{greenery}: lush; \objbox{photograph}: non-stylized, standard

\relbox{relations}: (\objbox{boat}, in, \objbox{distance}); (\objbox{scene}, on, \objbox{water}); (\objbox{water}, reflect, \objbox{light}); (\objbox{background}, fill with, \objbox{greenery}); (\objbox{boat}, in, \objbox{photograph}); (\objbox{boat}, have, \objbox{hull}); (\objbox{cabin}, on the right side of, \objbox{hull})

\item[MindBridge] \objbox{objects}: boat, riverbank, water, roof, city, tree, waterway, buoy, hull, building, river, greenery, skyline

\atrbox{attributes}: \objbox{river}: blue, tranquil, deep shade; \objbox{tree}: lush, green; \objbox{hull}: glistening; \objbox{boat}: blue, white, docked; \objbox{roof}: white, yellow; \objbox{buoy}: bright; \objbox{water}: blue

\relbox{relations}: (\objbox{boat}, dock along, \objbox{riverbank}); (\objbox{buoy}, along, \objbox{boat}); (\objbox{boat}, with, \objbox{roof}); (\objbox{greenery}, surround, \objbox{boat}); (\objbox{city}, have, \objbox{building}); (\objbox{boat}, dock in, \objbox{river}); (\objbox{city}, have, \objbox{activity}); (\objbox{river}, have, \objbox{buoy}); (\objbox{boat}, have, \objbox{roof}); (\objbox{river}, in, \objbox{city}); (\objbox{river}, serve as, \objbox{waterway}); (\objbox{boat}, have, \objbox{hull}); (\objbox{city}, have, \objbox{skyline})

\item[UMBRAE] \objbox{objects}: sky, boat, water, roof, island, cloud, lighthouse, sun, light, people, river

\atrbox{attributes}: \objbox{boat}: warm, brown, blue, wooden, modern, rustic, white; \objbox{charm}: rustic; \objbox{roof}: white; \objbox{design}: sleek; \objbox{island}: small; \objbox{lighthouse}: beacon, tall, safety; \objbox{cloud}: fluffy, white; \objbox{sky}: blue, clear; \objbox{light}: soft; \objbox{sun}: shining, bright; \objbox{day}: day, peaceful

\relbox{relations}: (\objbox{people}, on, \objbox{water}); (\objbox{boat}, float on, \objbox{river}); (\objbox{roof}, add, \objbox{charm}); (\objbox{glow}, on, \objbox{water}); (\objbox{people}, on, \objbox{river}); (\objbox{island}, with, \objbox{lighthouse}); (\objbox{sun}, cast, \objbox{glow}); (\objbox{design}, on the right side of, \objbox{boat}); (\objbox{boat}, in, \objbox{water}); (\objbox{boat}, float on, \objbox{water}); (\objbox{boat}, accompanie of, \objbox{people})

\item[NeuroPictor] \objbox{objects}: pier, boat, water, motorboat, cabin, lighting, tree, stripe, hull, motor, sky, rope

\atrbox{attributes}: \objbox{motorboat}: white; \objbox{stripe}: blue; \objbox{cabin}: small, white; \objbox{motor}: outboard, white; \objbox{boat}: large; \objbox{hull}: blue; \objbox{pier}: wooden; \objbox{tree}: lush, green; \objbox{sky}: blue, clear; \objbox{water}: calm; \objbox{lighting}: natural; \objbox{photograph}: realistic

\relbox{relations}: (\objbox{marking}, on, \objbox{boat}); (\objbox{boat}, secure with, \objbox{rope}); (\objbox{boat}, dock at, \objbox{pier}); (\objbox{cabin}, have, \objbox{motor}); (\objbox{stripe}, on side of, \objbox{motorboat}); (\objbox{photograph}, have, \objbox{lighting}); (\objbox{boat}, have, \objbox{cabin}); (\objbox{pier}, in front of, \objbox{tree}); (\objbox{boat}, on, \objbox{pier}); (\objbox{boat}, have, \objbox{hull})

\item[NeuroVLA] \objbox{objects}: building, people, line, greenery, racer, tree, body, race track

\atrbox{attributes}: \objbox{scene}: vibrant, dynamic; \objbox{area}: grassy; \objbox{track}: smooth; \objbox{body}: leaning; \objbox{building}: blue

\relbox{relations}: (\objbox{individual}, on, \objbox{track}); (\objbox{scene}, fill with, \objbox{color}); (\objbox{tree}, in, \objbox{line}); (\objbox{tree}, on side of, \objbox{track}); (\objbox{racer}, have, \objbox{body language}); (\objbox{people}, walk around, \objbox{track}); (\objbox{scene}, at, \objbox{race track}); (\objbox{area}, border, \objbox{track}); (\objbox{building}, stand against, \objbox{greenery}); (\objbox{building}, link to, \objbox{race track})

\item[UniBrain] \objbox{objects}: sailboat, water, path, road, people, trailer, boat, motorboat, line, beach, truck, vehicle, tree, sky, lighting, vegetation

\atrbox{attributes}: \objbox{beach}: sandy; \objbox{motorboat}: small; \objbox{sailboat}: large; \objbox{boat}: moored; \objbox{people}: visible, far away; \objbox{truck}: large, small; \objbox{lighting}: soft; \objbox{sky}: overcast; \objbox{tree}: small; \objbox{element}: natural, man-made; \objbox{photograph}: non-stylized, standard

\relbox{relations}: (\objbox{vehicle}, have, \objbox{trailer}); (\objbox{vegetation}, have, \objbox{tree}); (\objbox{road}, on the right side of, \objbox{image}); (\objbox{motorboat}, on, \objbox{beach}); (\objbox{vegetation}, on edge of, \objbox{water}); (\objbox{beach}, lead up to, \objbox{water}); (\objbox{vehicle}, park in, \objbox{line}); (\objbox{vehicle}, park along, \objbox{path}); (\objbox{text}, in, \objbox{image}); (\objbox{road}, run parallel to, \objbox{edge})

\item[STTM] \objbox{objects}: pier, boat, water, marina, hub, tree, stripe, foliage, sky

\atrbox{attributes}: \objbox{scene}: serene; \objbox{surface}: rippling; \objbox{boat}: vibrant, blue, white; \objbox{stripe}: white; \objbox{pier}: wooden; \objbox{foliage}: lush, green; \objbox{sky}: blue, clear; \objbox{water}: tranquil

\relbox{relations}: (\objbox{tree}, have, \objbox{foliage}); (\objbox{boat}, glid across, \objbox{water}); (\objbox{pier}, extend into, \objbox{water}); (\objbox{boat}, dock at, \objbox{hub}); (\objbox{boat}, have, \objbox{stripe}); (\objbox{marina}, fill with, \objbox{activity}); (\objbox{boat}, contrast with, \objbox{structure}); (\objbox{marina}, fill with, \objbox{water}); (\objbox{tree}, in, \objbox{background}); (\objbox{water}, have, \objbox{surface})

\item[MindTuner] \objbox{objects}: water, boat, mast, roof, window, tree, stripe, deck, sky, sail, lighting

\atrbox{attributes}: \objbox{boat}: large, white, passenger; \objbox{roof}: blue; \objbox{stripe}: red; \objbox{area}: green, covered, forested; \objbox{deck}: upper; \objbox{tree}: lush; \objbox{sky}: overcast; \objbox{lighting}: natural; \objbox{photograph}: realistic

\relbox{relations}: (\objbox{area}, with, \objbox{tree}); (\objbox{boat}, have, \objbox{sail}); (\objbox{boat}, have, \objbox{mast}); (\objbox{photograph}, have, \objbox{lighting}); (\objbox{window}, on, \objbox{deck}); (\objbox{stripe}, on side of, \objbox{boat}); (\objbox{text}, on, \objbox{boat}); (\objbox{boat}, have, \objbox{window}); (\objbox{boat}, float on, \objbox{water})

\item[BrainGuard] \objbox{objects}: water, boat, harbor, dock, wood, sky, hillside

\atrbox{attributes}: \objbox{boat}: colors, varying sizes; \objbox{sky}: blue, clear; \objbox{water}: calm; \objbox{hillside}: lush, green; \objbox{atmosphere}: idyllic, peaceful

\relbox{relations}: (\objbox{boat}, at, \objbox{dock}); (\objbox{dock}, make of, \objbox{wood}); (\objbox{boat}, tie to, \objbox{dock}); (\objbox{dock}, extend into, \objbox{water})
}
\end{itemize}

\subsection{Captioning with Different MLLMs}
\label{subsec:supmat_ablation_mllm}

The following two subsections present ablation examples evaluating the impact of different MLLMs and prompting strategies on detailed captioning. For clarity and brevity, we report only the parsed descriptions of~\objbox{objects},~\atrbox{attributes}, and~\relbox{relationships} in the structured format. 
Despite variations in the generated captions across different MLLMs and prompts, the evaluation scores and method ranking remain consistent, as shown in~\cref{fig:ablation}, with more detailed scores in the quantitative evaluation (see ~\cref{tab:supmat_ab_prompt} and~\cref{tab:supmat_ab_mllm}
).

\vspace{8pt}
\begin{itemize}[leftmargin=25mm]
\setlength{\itemsep}{2pt}
\item[\textbf{MLLM}] \textbf{Parsed Description}
\vspace{3pt}
\small{
\item[LLaVA-1.5-7B] \objbox{objects}: lake, people, boat, tree, car, person, water, pier

\atrbox{attributes}: 
\objbox{boat}: small, large, few; 
\objbox{people}: scattered; 
\objbox{car}: parked

\relbox{relations}: 
(\objbox{pier}, on, \objbox{lake}); 
(\objbox{car}, park on the right side of, \objbox{scene}); 
(\objbox{boat}, in, \objbox{center}); 
(\objbox{person}, stand on edge of, \objbox{water}); 
(\objbox{boat}, in, \objbox{background}); 
(\objbox{people}, near, \objbox{boat}); 
(\objbox{tree}, surround, \objbox{boat})

\item[LLaVA-1.5-13B] \objbox{objects}: dock, ripple, hillside, stripe, boat, tree, sky, water, roof, pylon, window

\atrbox{attributes}: 
\objbox{water}: calm; 
\objbox{boat}: large, white, small; 
\objbox{dock}: wooden, floating; 
\objbox{pylon}: wooden; 
\objbox{stripe}: blue; 
\objbox{area}: covered; 
\objbox{hillside}: grassy; 
\objbox{sky}: diffused, overcast; 
\objbox{photograph}: straightforward, color; 
\objbox{foreground}: natural

\relbox{relations}: 
(\objbox{stripe}, run along, \objbox{boat}); 
(\objbox{window}, on side of, \objbox{area}); 
(\objbox{hillside}, with, \objbox{tree}); 
(\objbox{dock}, support by, \objbox{pylon}); 
(\objbox{dock}, extend into, \objbox{water}); 
(\objbox{boat}, in, \objbox{foreground}); 
(\objbox{boat}, dock at, \objbox{dock}); 
(\objbox{area}, with, \objbox{roof}); 
(\objbox{water}, with, \objbox{ripple})

\item[LLaVA-1.6-7B] \objbox{objects}: awning, marina, boat, tree, island, water, pier

\atrbox{attributes}: \objbox{scene}: serene; \objbox{boat}: black, white, back; \objbox{tree}: green, lush; \objbox{island}: spotted, small; \objbox{water}: calm; \objbox{view}: picturesque

\relbox{relations}: (\objbox{boat}, dock at, \objbox{pier}); (\objbox{island}, in, \objbox{marina}); (\objbox{boat}, have, \objbox{awning}); (\objbox{boat}, in, \objbox{marina}); (\objbox{marina}, nest among, \objbox{tree})
\item[LLaVA-1.6-13B] \objbox{objects}: boat, water, pylon, roof, ripple, window, stripe, tree, dock, sky, hillside

\atrbox{attributes}: \objbox{water}: calm; \objbox{boat}: large, white, small; \objbox{dock}: wooden, floating; \objbox{pylon}: wooden; \objbox{stripe}: blue; \objbox{area}: covered; \objbox{hillside}: grassy; \objbox{sky}: overcast, diffused; \objbox{photograph}: color, straightforward; \objbox{foreground}: natural 

\relbox{relations}: (\objbox{window}, on side of, \objbox{area}); (\objbox{dock}, extend into, \objbox{water}); (\objbox{boat}, dock at, \objbox{dock}); (\objbox{hillside}, with, \objbox{tree}); (\objbox{boat}, in, \objbox{foreground}); (\objbox{water}, with, \objbox{ripple}); (\objbox{area}, with, \objbox{roof}); (\objbox{stripe}, run along, \objbox{boat}); (\objbox{dock}, support by, \objbox{pylon})

\item[LLaVA-1.6-34B] \objbox{objects}: tree, boat, hillside, dock, pier, stripe, waterfront, sky, water 

\atrbox{attributes}: \objbox{waterfront}: serene; \objbox{pier}: wooden; \objbox{boat}: small, blue, white; \objbox{stripe}: blue; \objbox{area}: covered; \objbox{water}: calm; \objbox{hillside}: lush, green, verdant; \objbox{sky}: gray, overcast; \objbox{dock}: peaceful; \objbox{tree}: green; \objbox{object}: relative; \objbox{landscape}: natural; \objbox{scene}: waterfront, realistic 

\relbox{relations}: (\objbox{stripe}, on side of, \objbox{boat}); (\objbox{boat}, have, \objbox{stripe}); (\objbox{boat}, in, \objbox{image}); (\objbox{tree}, cover, \objbox{hillside}); (\objbox{object}, dock at, \objbox{pier}); (\objbox{boat}, dock at, \objbox{pier}); (\objbox{boat}, on, \objbox{water}); (\objbox{pier}, extend into, \objbox{water}); (\objbox{boat}, against, \objbox{sky}); (\objbox{boat}, on, \objbox{dock})

\item[GPT-4o] \objbox{objects}: yacht, lake, shoreline, forest, dock, boat, houseboat, hill, water, pier, shore

\atrbox{attributes}: 
\objbox{forest}: green, dense; 
\objbox{lake}: calm; 
\objbox{boat}: large, small; 
\objbox{motor yacht}: white; 
\objbox{top}: flat, sleek; 
\objbox{dock}: metal, wooden; 
\objbox{water}: windless, calm; 
\objbox{hill}: forested; 
\objbox{scene}: tranquil, scenic

\relbox{relations}: 
(\objbox{boat}, dock at, \objbox{pier}); 
(\objbox{forest}, surround, \objbox{lake}); 
(\objbox{hill}, in, \objbox{background}); 
(\objbox{dock}, connect to, \objbox{boat}); 
(\objbox{dock}, anchor to, \objbox{shore}); 
(\objbox{forest}, line, \objbox{shoreline}); 
(\objbox{yacht}, have, \objbox{top}); 
(\objbox{boat}, on, \objbox{lake})
}
\end{itemize}

\subsection{Captioning with Different Prompts}
\label{subsec:supmat_ablation_prompt}

This section presents a comparison of different \textit{description prompts} and their effects on the  detailed caption generation, using LLaVA-1.6-7B. %
Specifically, we compare three description prompts: (1) ``Describe the image in detail.'', (2) ``Describe the image in detail, focusing on visible objects.'', and (3) ``Describe the image in detail, focusing on visible objects and their relationships.''
Below are the parsed descriptions of \objbox{objects}, \atrbox{attributes}, and \relbox{relationships} from the obtained captions. 

\vspace{8pt}
\begin{itemize}[leftmargin=17.5mm]
\setlength{\itemsep}{2pt}
\item[\textbf{Prompt}] \textbf{Parsed Description}
\vspace{3pt}
\small{
\item[ID \#1] \objbox{objects}: awning, marina, boat, tree, island, water, pier

\atrbox{attributes}: \objbox{scene}: serene; \objbox{boat}: black, white, back; \objbox{tree}: green, lush; \objbox{island}: spotted, small; \objbox{water}: calm; \objbox{view}: picturesque

\relbox{relations}: (\objbox{boat}, dock at, \objbox{pier}); (\objbox{island}, in, \objbox{marina}); (\objbox{boat}, have, \objbox{awning}); (\objbox{boat}, in, \objbox{marina}); (\objbox{marina}, nest among, \objbox{tree})

\item[ID \#2] \objbox{objects}: lake, greenery, marina, sky, boat, tree, island, water, hull, pier

\atrbox{attributes}: 
\objbox{boat}: large, white, small; 
\objbox{design}: sleek; 
\objbox{pier}: wooden; 
\objbox{water}: calm, blue; 
\objbox{hull}: white; 
\objbox{tree}: green, lush; 
\objbox{island}: small; 
\objbox{sky}: clear, blue; 
\objbox{marvel}: man-made

\relbox{relations}: 
(\objbox{tree}, on either side of, \objbox{lake}); 
(\objbox{boat}, in, \objbox{water}); 
(\objbox{boat}, dock at, \objbox{pier}); 
(\objbox{marvel}, in, \objbox{setting}); 
(\objbox{island}, punctuate, \objbox{horizon}); 
(\objbox{boat}, have, \objbox{hull}); 
(\objbox{island}, with, \objbox{greenery}); 
(\objbox{pier}, extend into, \objbox{water}); 
(\objbox{lake}, nest among, \objbox{tree}); 
(\objbox{tree}, in, \objbox{scene}); 
(\objbox{boat}, have, \objbox{design}); 
(\objbox{boat}, mirror, \objbox{boat})

\item[ID \#3] \objbox{objects}: lake, light, marina, sky, boat, foliage, tree, water, pier

\atrbox{attributes}: 
\objbox{boat}: modern, large, white, blue; 
\objbox{design}: sleek; 
\objbox{pier}: weathered, wooden; 
\objbox{water}: calm; 
\objbox{tranquility}: broken; 
\objbox{form}: small; 
\objbox{foliage}: green; 
\objbox{tree}: lush; 
\objbox{light}: soft; 
\objbox{scene}: peaceful; 
\objbox{sky}: overcast; 
\objbox{marina}: beautiful

\relbox{relations}: 
(\objbox{light}, over, \objbox{scene}); 
(\objbox{form}, scatter across, \objbox{water}); 
(\objbox{pier}, extend into, \objbox{water}); 
(\objbox{lake}, nest among, \objbox{tree}); 
(\objbox{boat}, have, \objbox{design}); 
(\objbox{boat}, in, \objbox{background}); 
(\objbox{pier}, contrast with, \objbox{boat}); 
(\objbox{sky}, have, \objbox{light}); 
(\objbox{marina}, have, \objbox{backdrop}); 
(\objbox{lake}, have, \objbox{foliage})
}
\end{itemize}

\subsection{Comparison of Multigranular Segmentation Results}

\cref{fig:supmat_sam_comparison} shows the comparison of multigranular segmentation results between reference images and reconstructions. 
The segmentations are visualized at multiple levels, including semantic, instance, and part granularity. Unlike previous metrics that evaluate only low-level consistency with the reference using a single black-box score, our segmentation-based multigranular metric provides more interpretable insights. IoU and AP scores are higher when the reconstructions are structurally aligned with the reference. 
Taking examples in~\cref{fig:supmat_sam_comparison}, reconstructions where an airplane or zebra appears in the correct position and orientation receive higher scores—a property that holds across different granularities. For instance, BrainDiffuser~\cite{ozcelik2023brain} yields lower scores due to incorrect airplane orientation, while MindEye~\cite{scotti2023reconstructing} experiences score drops resulting from perspective shifts caused by camera zoom.
Overall, most methods produce reconstructions that preserve the high-level semantic structure of the original images, as reflected in the alignment of major object categories. However, finer-grained details, such as part boundaries or object instances in crowded scenes, are often less accurately recovered, suggesting limitations in capturing precise spatial layouts and textures.

\begin{figure*}[thbp]
\centering
\renewcommand{\arraystretch}{0.6}
\setkeys{Gin}{width=0.16\linewidth}
\setlength{\tabcolsep}{1.2pt}
\footnotesize
{
\resizebox{\textwidth}{!}{
\begin{tabular}[t]{ccccccc}
\parbox[t]{2mm}{\multirow{1}{*}{\rotatebox[origin=c]{90}{Reference\hspace{-7em}}}} & \includegraphics{images/supmat/sam/reference/foreground/361.jpg} & \includegraphics{images/supmat/sam/reference/background/361.jpg} & \includegraphics{images/supmat/sam/reference/part/361.jpg} & \includegraphics{images/supmat/sam/reference/foreground/893.jpg} & \includegraphics{images/supmat/sam/reference/background/893.jpg} & \includegraphics{images/supmat/sam/reference/part/893.jpg} \\
\parbox[t]{2mm}{\multirow{1}{*}{\rotatebox[origin=c]{90}{BrainDiffuser\hspace{-7em}}}} & \includegraphics{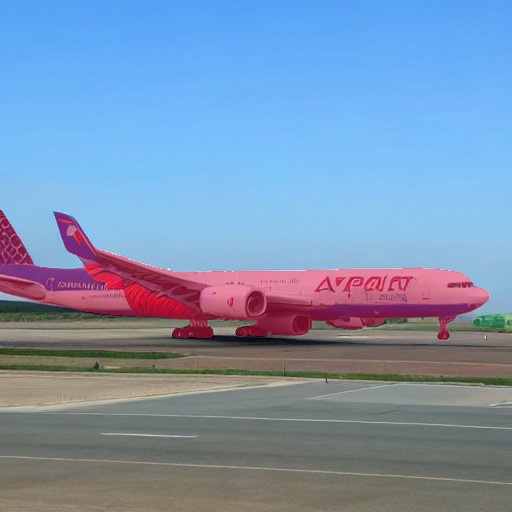} & \includegraphics{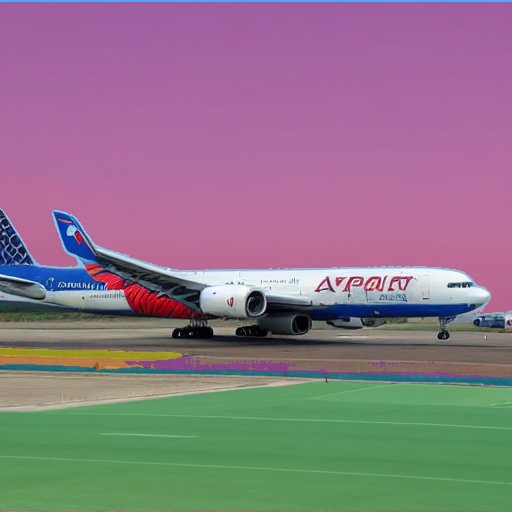} & \includegraphics{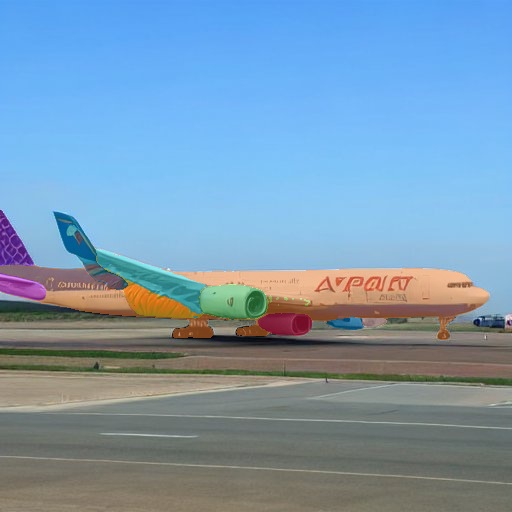} & \includegraphics{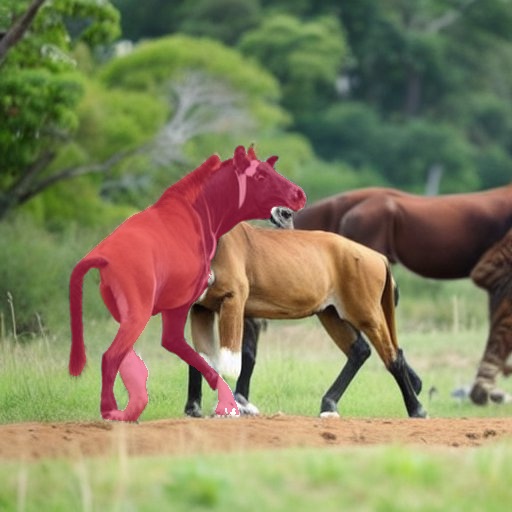} & \includegraphics{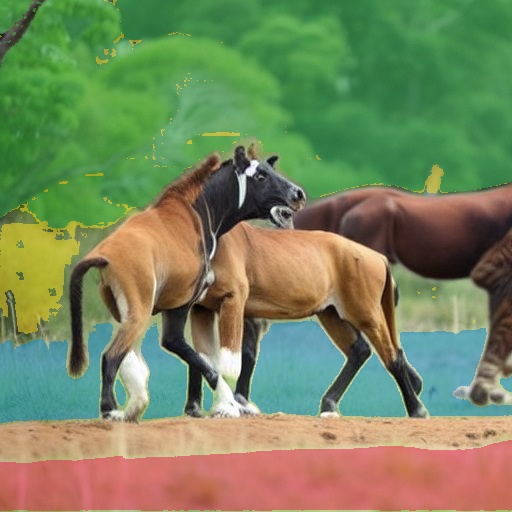} & \includegraphics{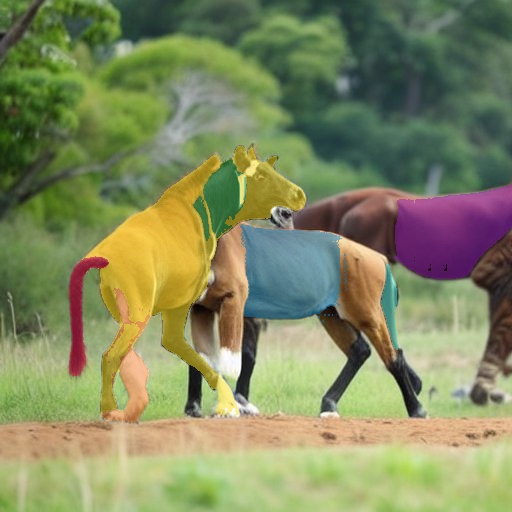} \\
\parbox[t]{2mm}{\multirow{1}{*}{\rotatebox[origin=c]{90}{MindEye\hspace{-7em}}}} & \includegraphics{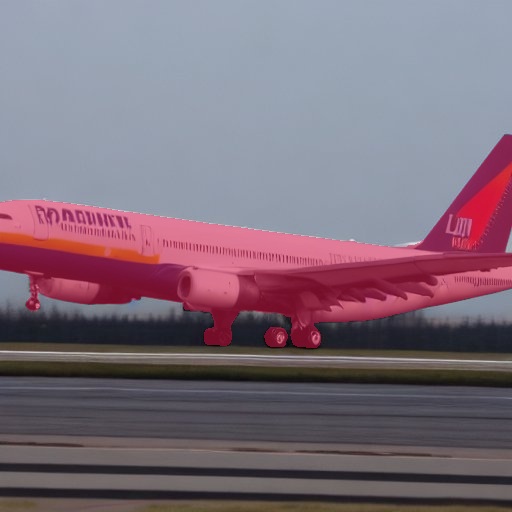} & \includegraphics{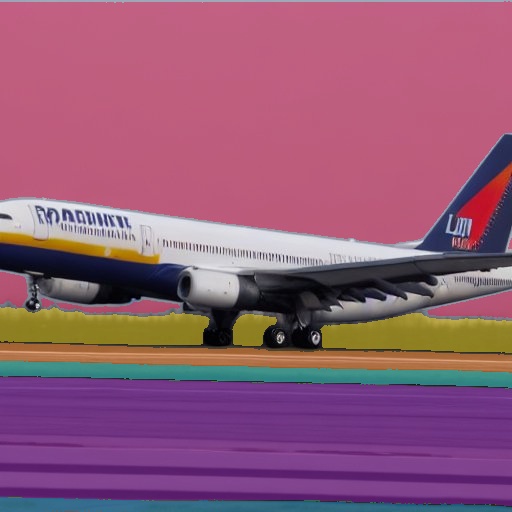} & \includegraphics{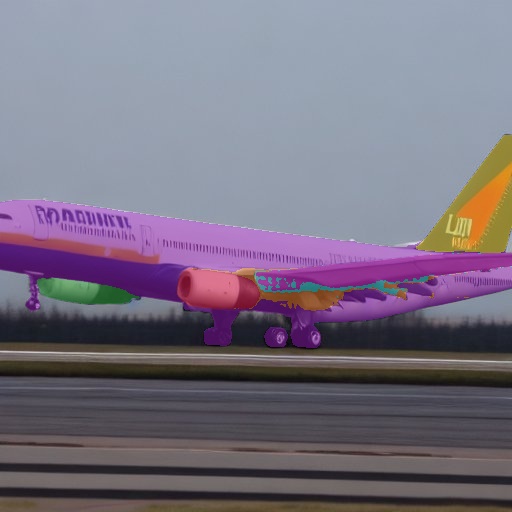} & \includegraphics{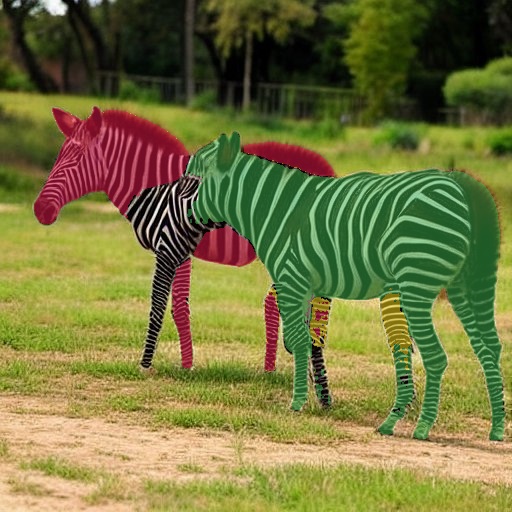} & \includegraphics{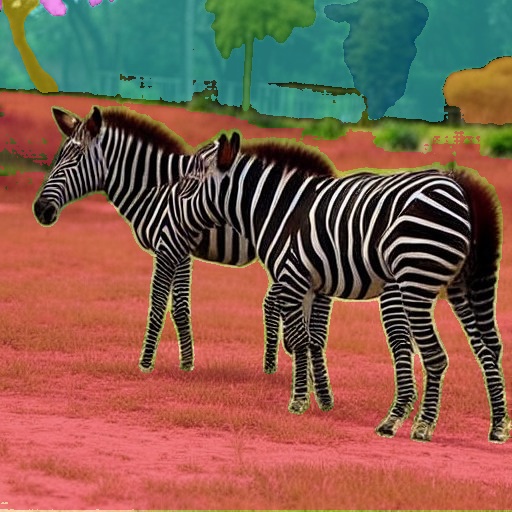} & \includegraphics{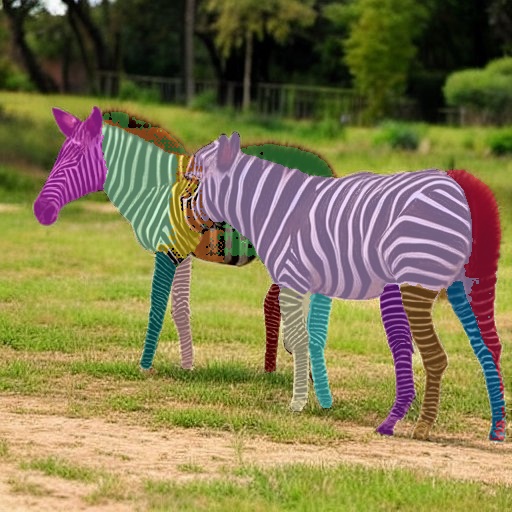} \\
\parbox[t]{2mm}{\multirow{1}{*}{\rotatebox[origin=c]{90}{MindEye2\hspace{-7em}}}} & \includegraphics{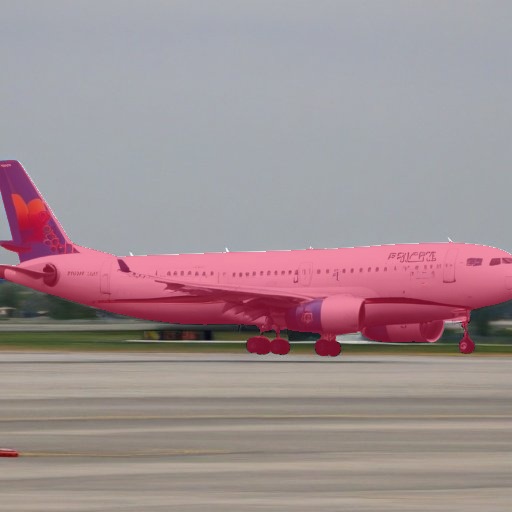} & \includegraphics{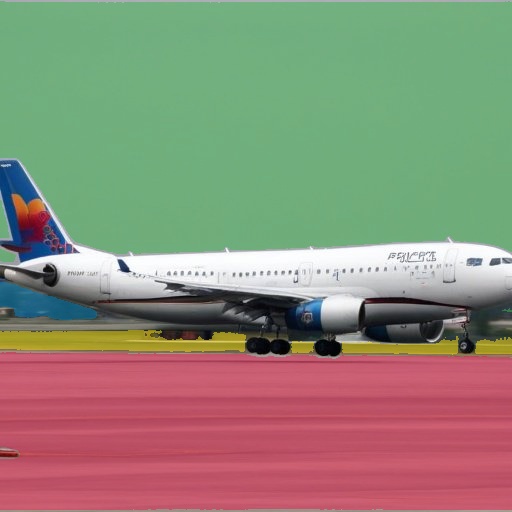} & \includegraphics{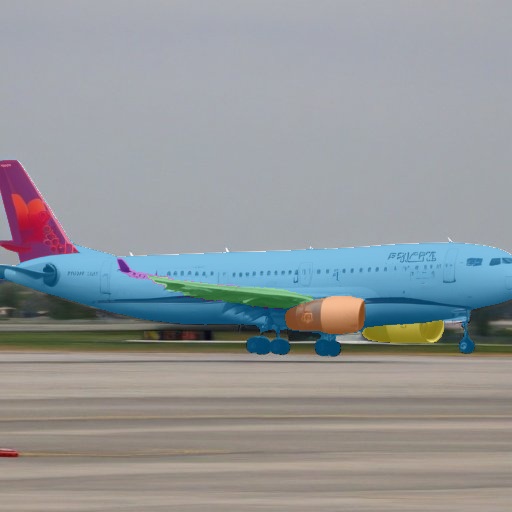} & \includegraphics{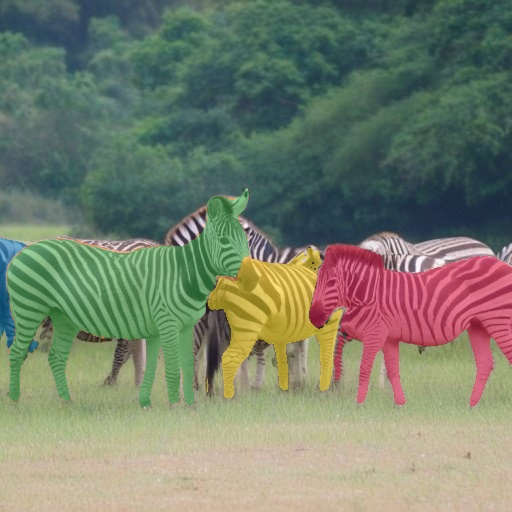} & \includegraphics{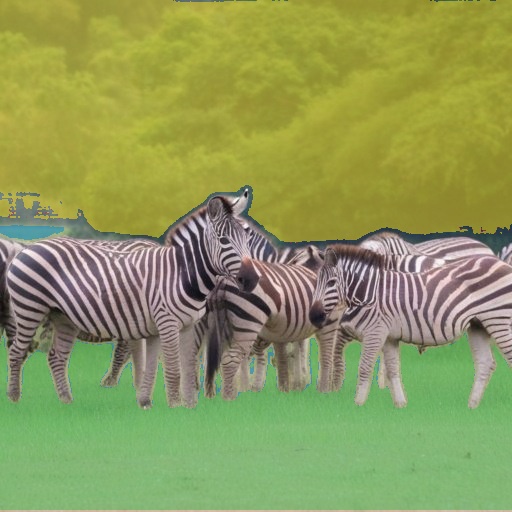} & \includegraphics{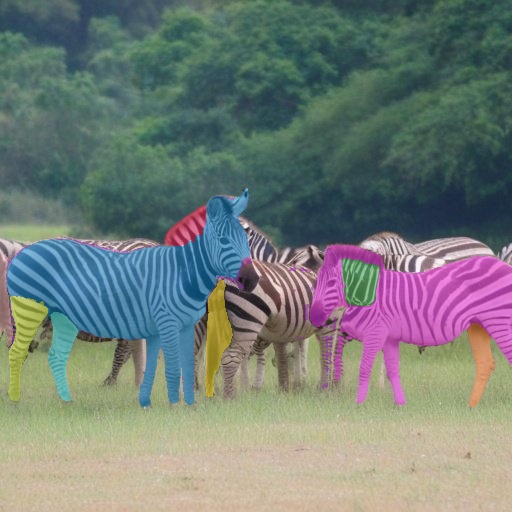} \\
\parbox[t]{2mm}{\multirow{1}{*}{\rotatebox[origin=c]{90}{NeuroPictor\hspace{-7em}}}} & \includegraphics{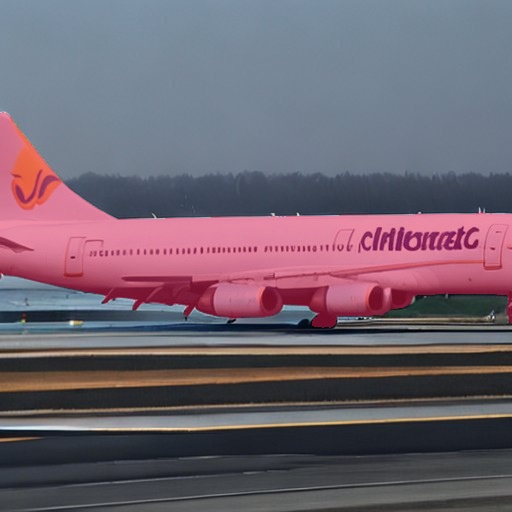} & \includegraphics{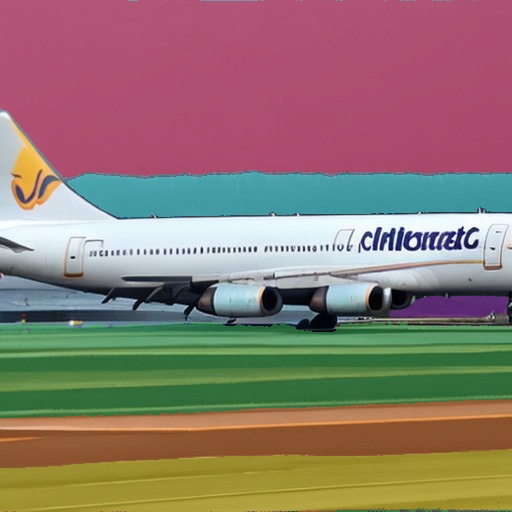} & \includegraphics{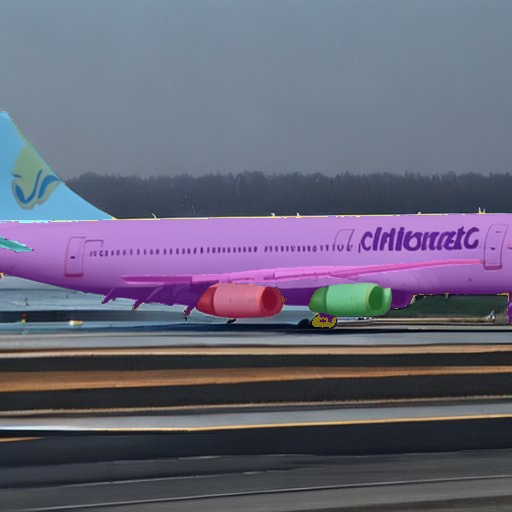} & \includegraphics{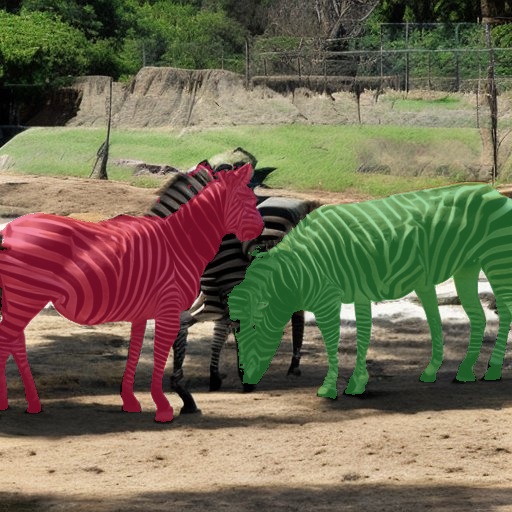} & \includegraphics{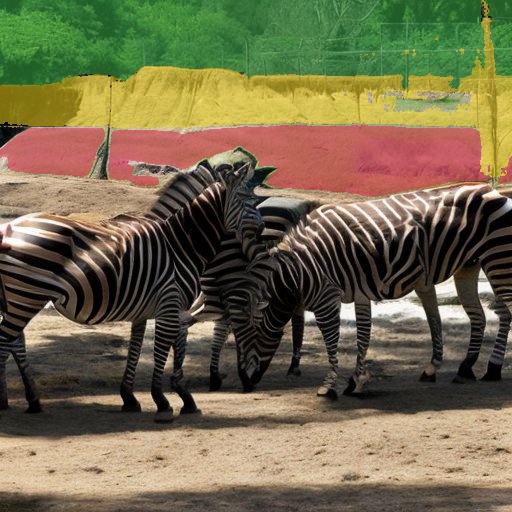} & \includegraphics{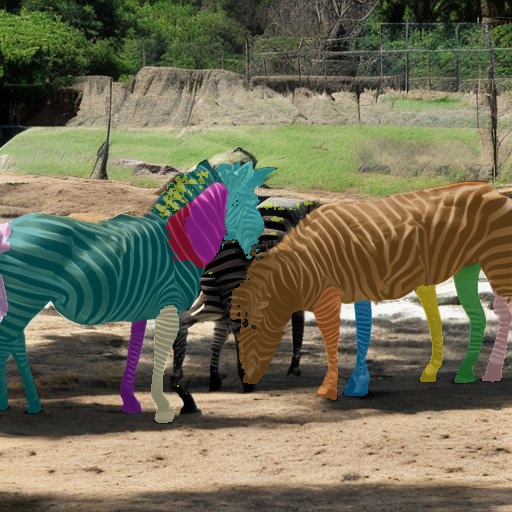} \\
\parbox[t]{2mm}{\multirow{1}{*}{\rotatebox[origin=c]{90}{DREAM\hspace{-7em}}}} & \includegraphics{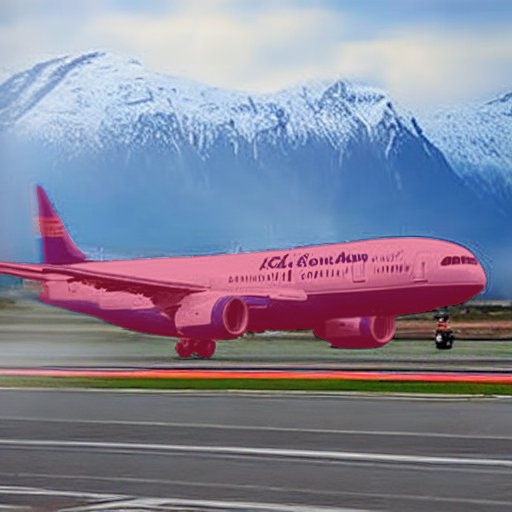} & \includegraphics{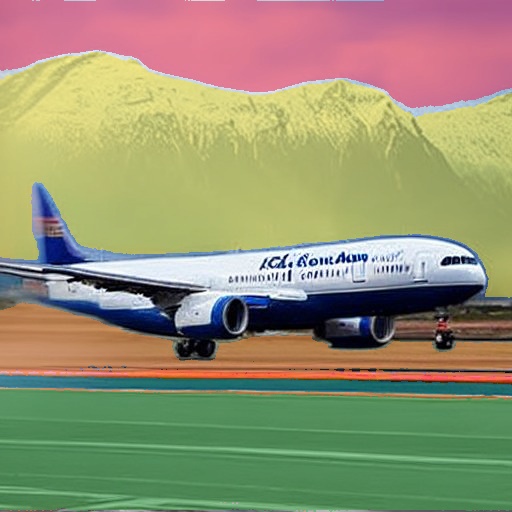} & \includegraphics{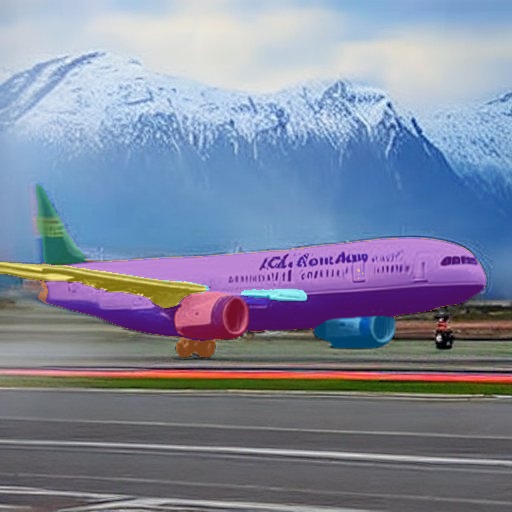} & \includegraphics{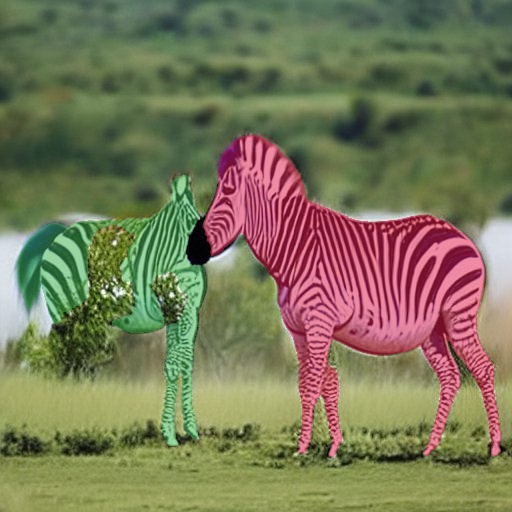} & \includegraphics{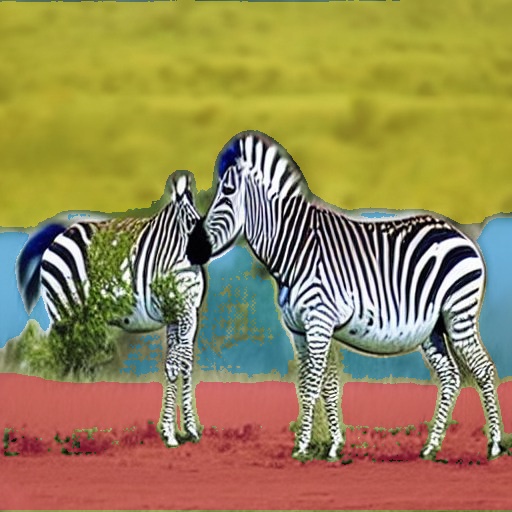} & \includegraphics{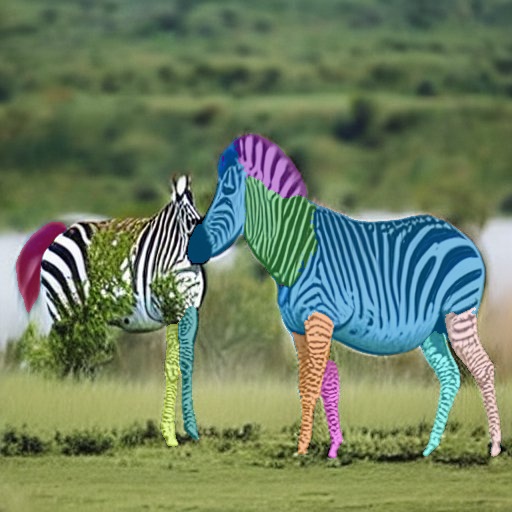} \\
\parbox[t]{2mm}{\multirow{1}{*}{\rotatebox[origin=c]{90}{UMBRAE\hspace{-7em}}}} & \includegraphics{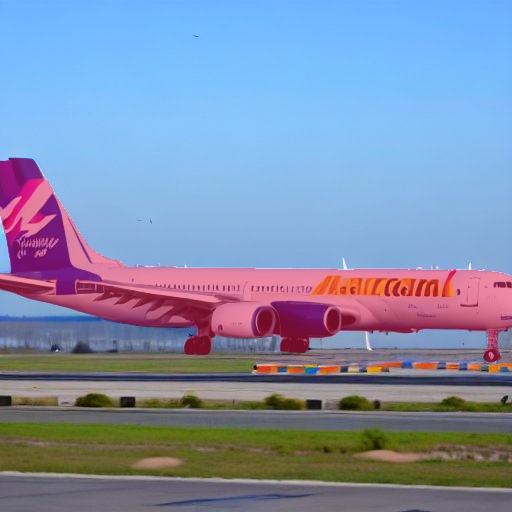} & \includegraphics{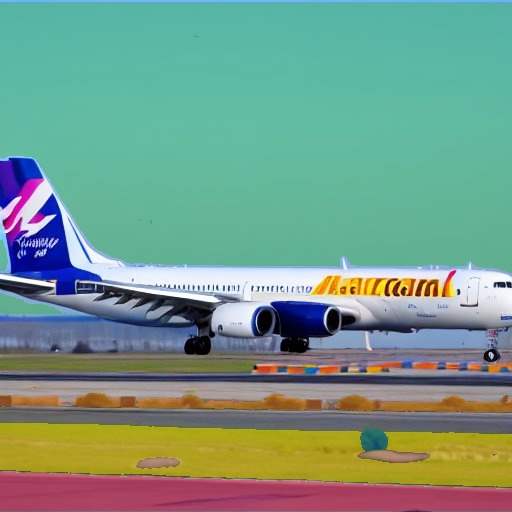} & \includegraphics{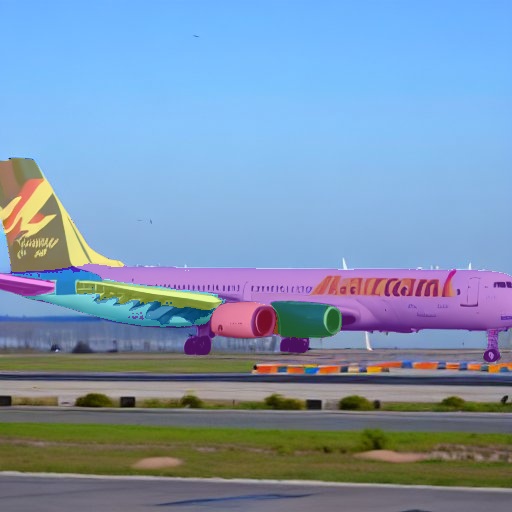} & \includegraphics{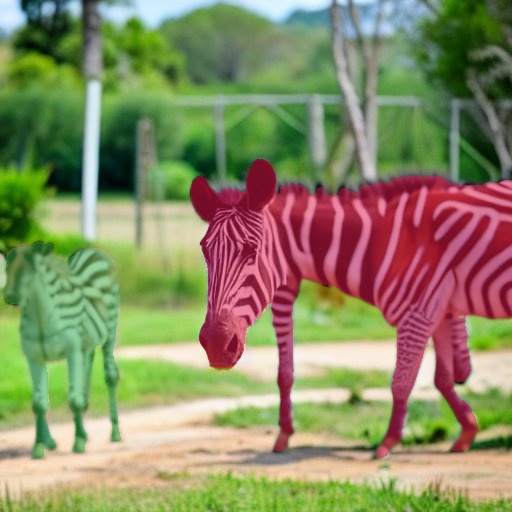} & \includegraphics{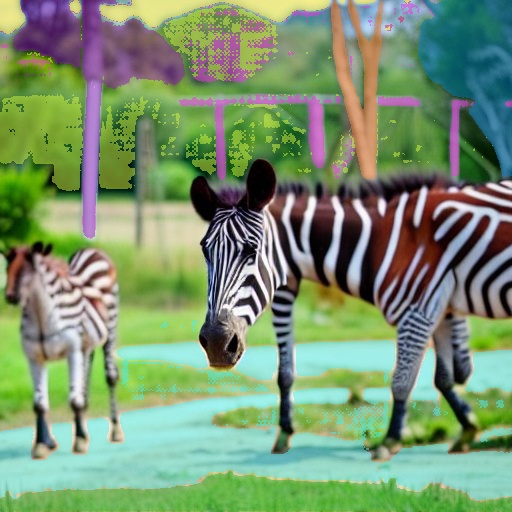} & \includegraphics{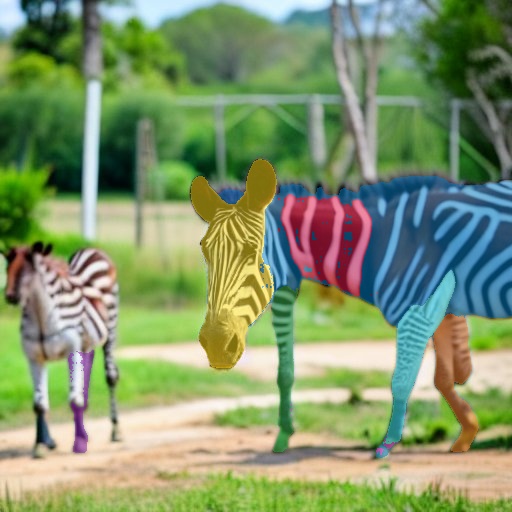} \\
\parbox[t]{2mm}{\multirow{1}{*}{\rotatebox[origin=c]{90}{BrainGuard\hspace{-7em}}}} & \includegraphics{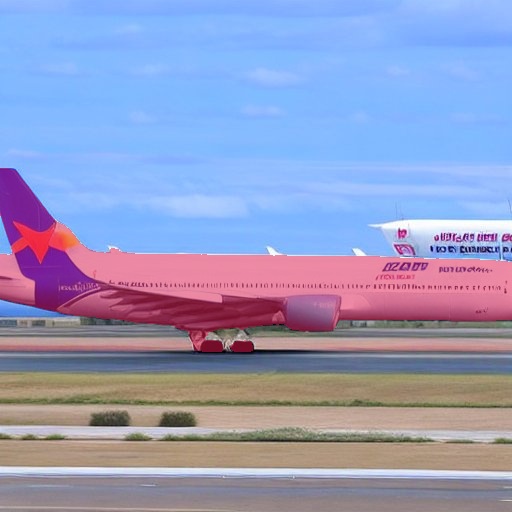} & \includegraphics{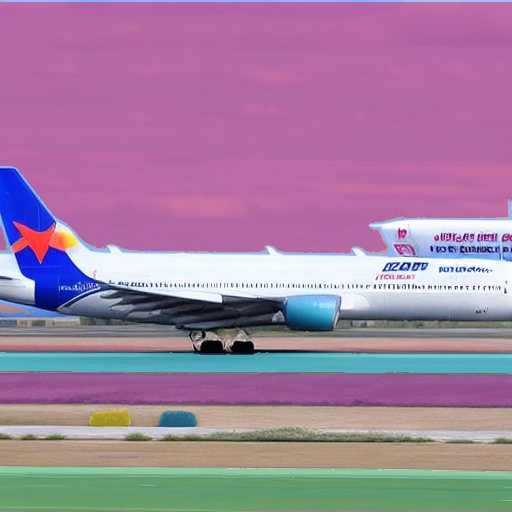} & \includegraphics{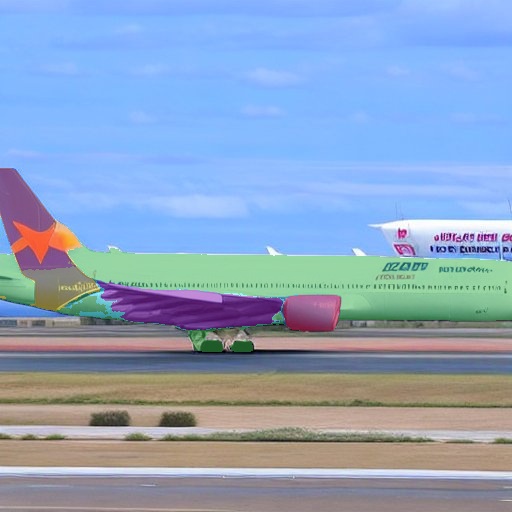} & \includegraphics{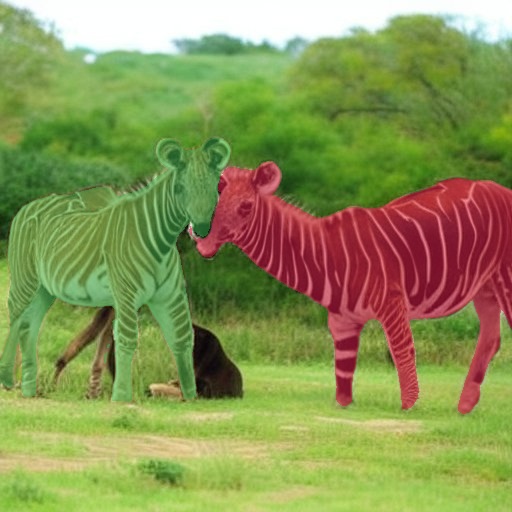} & \includegraphics{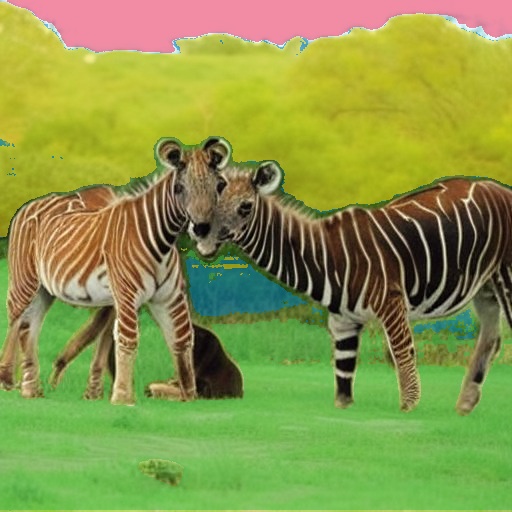} & \includegraphics{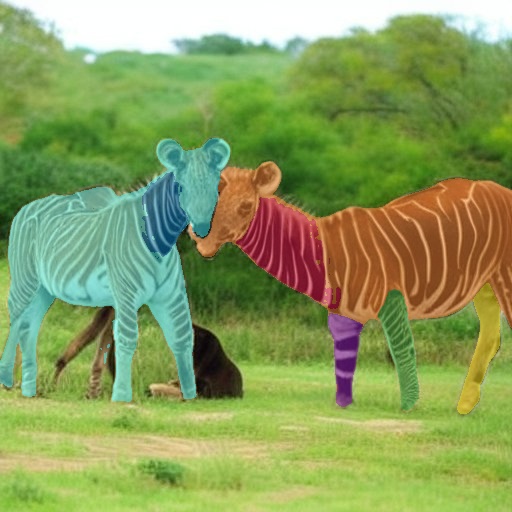} \\
& foreground & background & component & foreground & background & component \\
\end{tabular}
}
\caption{Comparison of multigranular segmentations between reference images and reconstructions.
}
\label{fig:supmat_sam_comparison}
}
\end{figure*}

\setlength{\tabcolsep}{4pt}
\setlength{\fboxrule}{0pt} 
\setlength{\fboxsep}{2pt}
\begin{table*}[t!]
\centering
\caption{Evaluation results using different prompt strategies. Three prompts are considered:  (1) ``Describe the image in detail.'', (2) ``Describe the image in detail, focusing on visible objects.'', and (3) ``Describe the image in detail, focusing on visible objects and their relationships.'' 
The evaluation scores remain stable and the ranking of methods is consistent despite numerical fluctuations across prompts, indicating that our metric is robust to moderate prompt variations. 
}
\label{tab:supmat_ab_prompt}
\resizebox{0.85\textwidth}{!}
{
\begin{tabular}{c|l|ccc|ccc|ccc|c|c}
\toprule
~ & \multirow{2}{*}{Method} & \multicolumn{3}{c|}{Object}  & \multicolumn{3}{c|}{Attribute}  & \multicolumn{3}{c|}{Relation} & \multirow{2}{*}{BASIC-H} & \multirow{2}{*}{Rank} \\
~ & ~ & P & R & F1 & P & R & F1 & P & R & F1 & ~ & ~\\
\midrule
\parbox[t]{3mm}{\multirow{7}{*}{\rotatebox[origin=c]{90}{Prompt \#1}}} & BrainDiffuser & 57.68 &  58.61 &  57.77 &  14.70 &  47.55 &  20.97 &  42.57 &  43.50 &  42.69 &  40.04 & 7\\
~ & UMBRAE    & 61.12 &  62.48 &  61.37 &  18.17 &  53.93 &  25.61 &  47.47 &  48.96 &  47.84 &  44.36 & 6 \\
~ & NeuroPictor   &  62.31 &  62.21 &  61.68 &  18.38 &  53.63 &  25.43 &  48.62 &  49.49 &  48.67 &  44.58 & 5 \\
~ & MindEye   & 62.76 &  60.86 &  61.27 &  19.87 &  53.13 &  27.11 &  49.09 &  48.01 &  48.17 &  44.98 & 4 \\
~ & MindEye2  & 62.09 &  63.39 &  62.21 &  18.93 &  53.30 &  26.10 &  49.15 &  50.36 &  49.40 &  45.20 & 3 \\
~ & BrainGuard & 63.29 &  63.18 &  62.68 &  19.69 &  54.61 &  27.18 &  50.34 &  51.75 &  50.62 &  46.07 & 2 \\
~ & DREAM & 64.97 &  64.27 &  63.94 &  19.52 &  52.98 &  26.94 &  52.73 &  54.52 &  53.18 &  46.99  & 1 \\
\midrule
\parbox[t]{3mm}{\multirow{7}{*}{\rotatebox[origin=c]{90}{Prompt \#2}}} & BrainDiffuser & 57.75 &  58.42 &  57.57 &  13.77 &  48.75 &  20.03 &  43.11 &  43.91 &  43.19 &  39.68 & 7 \\
~ & UMBRAE &  61.81 &  62.94 &  61.89 &  17.81 &  52.95 &  25.07 &  48.67 &  48.90 &  48.46 &  44.48 & 6 \\
~ & NeuroPictor  & 62.58 &  61.63 &  61.43 &  18.67 &  51.34 &  25.74 &  49.50 &  48.98 &  48.88 &  44.65 & 5 \\
~ & MindEye  &  62.99 &  63.27 &  62.60 &  18.46 &  50.78 &  25.43 &  50.53 &  50.20 &  50.03 &  45.22  & 4 \\
~ & MindEye2  &  63.45 &  61.59 &  61.99 &  19.42 &  51.99 &  26.44 &  50.32 &  48.95 &  49.27 &  45.23  & 3 \\
~ & BrainGuard  &  63.67 &  62.72 &  62.65 &  19.80 &  54.07 &  27.26 &  51.08 &  50.91 &  50.62 &  46.09 & 2  \\
~ & DREAM  &  64.91 &  63.83 &  63.55 &  19.45 &  52.58 &  26.78 &  52.97 &  53.47 &  52.80 & 46.69 & 1 \\
\midrule
\parbox[t]{3mm}{\multirow{7}{*}{\rotatebox[origin=c]{90}{Prompt \#3}}} & BrainDiffuser &  57.19 &  58.69 &  57.48 &  13.11 &  46.73 &  19.07 &  44.35 &  45.24 &  44.53 &  39.53 & 7 \\
~ & UMBRAE &  61.06 &  62.21 &  61.17 &  17.42 &  52.04 &  24.40 &  49.69 &  49.78 &  49.48 &  44.12 & 6 \\
~ & NeuroPictor &  62.57 &  61.46 &  61.34 &  17.40 &  50.27 &  24.13 &  50.86 &  50.39 &  50.32 &  44.25 & 5 \\
~ & MindEye &  62.45 &  60.87 &  61.09 &  18.27 &  50.55 &  25.15 &  51.20 &  49.81 &  50.17 &  44.53 & 4 \\
~ & MindEye2  &  62.49 &  62.68 &  61.94 &  17.93 &  50.59 &  24.74 &  51.12 &  51.33 &  50.94 &  44.86 & 3 \\
~ & BrainGuard   &  63.75 &  62.40 &  62.46 &  19.21 &  52.17 &  26.43 &  52.66 &  52.14 &  52.12 &  45.98 & 2 \\
~ & DREAM   &  64.59 &  63.88 &  63.39 &  19.32 &  51.77 &  26.57 &  54.60 &  54.62 &  54.28 &  46.84 & 1 \\
\bottomrule
\end{tabular}
}
\end{table*}

\section{Quantitative Comparative Analysis}
\label{sec:supmat_quantitative_results}

\subsection{Evaluating NSD with the BASIC}

This section presents a performance comparison of visual decoding methods on NSD~\cite{allen2022massive} using the proposed BASIC metrics, which capture both structural (BASIC-L) and semantic (BASIC-H) alignment. 
The final BASIC score is computed as the average of BASIC-L and BASIC-H, reflecting the equal importance of low-level structural and high-level semantic alignment in brain visual decoding evaluation.
\cref{fig:supmat_basic_nsd_barchart} presents the results, with the values of the top-5 best performing methods annotated for each dimension.
For structural alignment NeuroPictor~\cite{huo2024neuropictor}, STTM~\cite{liu2025see}, MindEye2~\cite{scotti2024mindeye2}, BrainGuard~\cite{tian2025brainguard}, and DREAM~\cite{xia2024dream} achieve the highest BASIC-L scores, reflecting better spatial reconstruction. In contrast, semantic alignment (BASIC-H) is best captured by NeuroVLA~\cite{shen2024neuro}, DREAM~\cite{xia2024dream}, STTM~\cite{liu2025see}, BrainGuard~\cite{tian2025brainguard}, and MindTuner~\cite{gong2025mindtuner}, indicating superior semantic preservation.
NeuroPictor~\cite{huo2024neuropictor} ranks highest in the composite BASIC score, followed by STTM~\cite{liu2025see}, BrainGuard~\cite{tian2025brainguard}, MindEye2~\cite{scotti2024mindeye2}, and DREAM~\cite{xia2024dream}, reflecting more balanced performance across both perceptual dimensions. While NeuroVLA~\cite{shen2024neuro} excels in semantic similarity, it underperforms in structural precision, suggesting a trade-off between high-level conceptual encoding and spatial accuracy. 
These results demonstrate the complementary nature of BASIC-L and BASIC-H and underscore the importance of multigranular evaluation for benchmarking brain visual decoding.

\begin{figure*}[t!]
    \centering
    \includegraphics[width=0.85\linewidth]{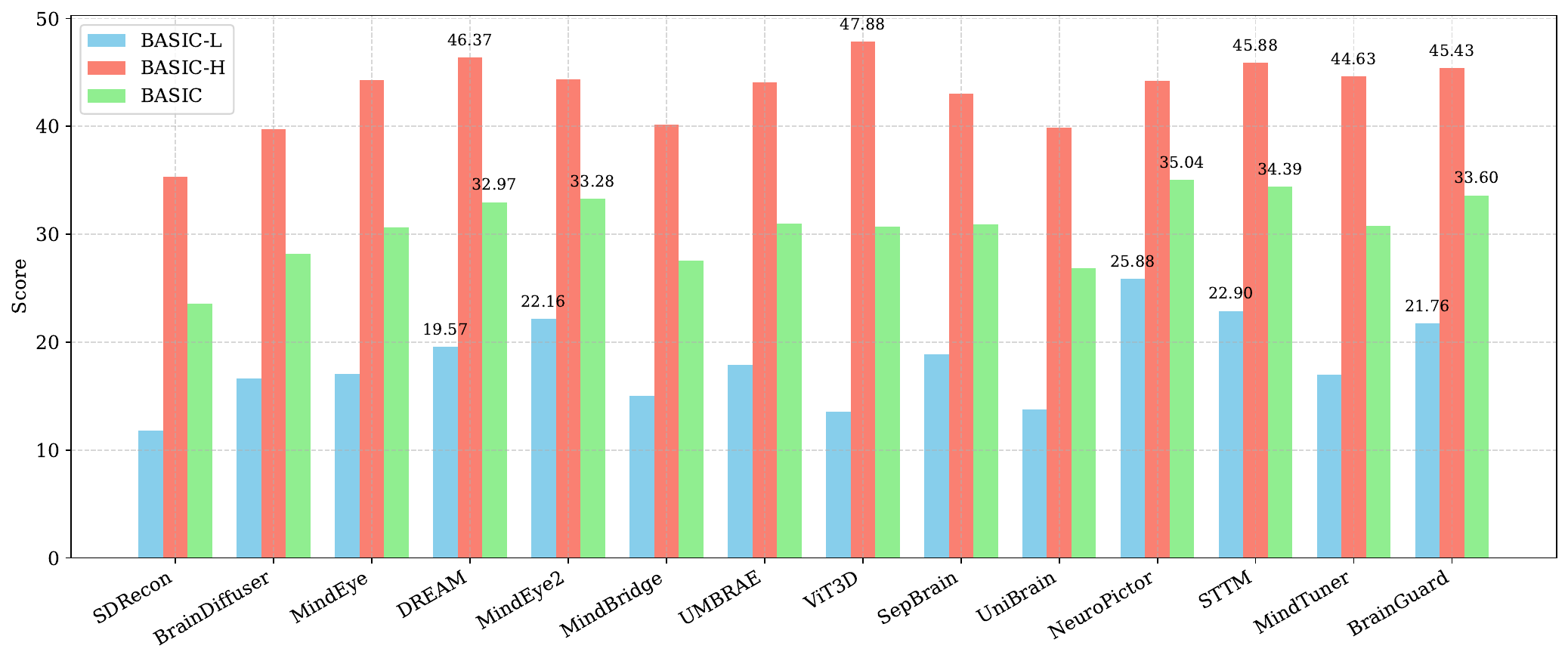}
    \caption{BASIC results for NSD evaluation.}
    \label{fig:supmat_basic_nsd_barchart}
\end{figure*}

\subsection{Stability across Evaluation Configurations}
\label{subsec:supmat_quantitative_comparison}

This section evaluates the robustness and consistency of our metric across multiple factors, including MLLM choices and prompt strategies in BASIC-H, as well as box and text threshold configurations in BASIC-L. The metric scores remain relatively stable and largely insensitive to such setup variations.

\cref{tab:supmat_ab_prompt} presents the quantitative results obtained using different captioning prompts. Three prompt variants, same as above
are considered: (1) ``Describe the image in detail.'', (2) ``Describe the image in detail, focusing on visible objects.'', and (3) ``Describe the image in detail, focusing on visible objects and their relationships.'' Despite slight differences in phrasing, the evaluation scores remain consistent across prompts, indicating that our metric is robust to moderate prompt variations and does not overly rely on specific wording. The MLLM used here is LLaVA-1.6-7B~\cite{liu2023visual}. The metric remains consistent and discriminative even with very close scores (refer to the demonstrated scores of UMBRAE~\cite{xia2024umbrae}, NeuroPictor~\cite{huo2024neuropictor}, MindEye~\cite{scotti2023reconstructing}, and MindEye2~\cite{scotti2024mindeye2}).

\cref{tab:supmat_ab_mllm} presents the quantitative evaluation results obtained using different MLLMs, including LLaVA-1.5-7B, LLaVA-1.5-13B, LLaVA-1.6-7B, LLaVA-1.6-13B, LLaVA-1.6-34B, and GPT-4o. Despite variations in model version and scale, the overall metric scores remain stable, and the relative method rankings are mostly preserved. This consistency across MLLMs demonstrates the robustness of our evaluation framework with respect to the choice of captioning model.

\cref{tab:supmat_ab_cross_evaluation} presents a cross-evaluation, where reconstruction captions from each method are generated using one MLLM and assessed against captions produced by different MLLMs as references. 
Since no \textit{human-ranked scores} are available for relevance analysis (i.e., consistency with human expert judgments using metrics such as Pearson correlation, coefficient of determination, or Kendall rank correlation), and obtaining reliable human-annotated relative rankings for each method is challenging, we instead adopt a cross-model evaluation strategy. Specifically, we assess performance by varying the MLLM models used as the producer (i.e., generating captions for the reconstructed images) and the evaluator (generating reference captions). This cross-model setup reveals consistency across MLLMs, suggesting that reliance on human-written ``ground truth'' captions is not strictly necessary for evaluation.
\cref{tab:supmat_ab_cross_evaluation_braindiffuser} presents the detailed subscores for BrainDiffuser in the cross-model evaluation.

\cref{tab:supmat_ab_threshold} summarizes the effect of varying box and text thresholds on the multigranular segmentation evaluation. While threshold changes influence individual sub-scores, the relative rankings of methods remain consistent within each configuration. The final BASIC-L score demonstrates notable stability across all threshold settings, despite fluctuations in component metrics.

\setlength{\tabcolsep}{4pt}
\setlength{\fboxrule}{0pt} 
\setlength{\fboxsep}{2pt}
\begin{table*}[t!]
\centering
\caption{Evaluation results with different MLLM choices. The evaluation scores remain stable and the method ranking is mostly consistent despite numerical fluctuations across MLLM choices.
}
\label{tab:supmat_ab_mllm}
{
\begin{tabular}{c|l|ccc|ccc|ccc|c|c}
\toprule
~ & \multirow{2}{*}{Method} & \multicolumn{3}{c|}{Object}  & \multicolumn{3}{c|}{Attribute}  & \multicolumn{3}{c|}{Relation} & \multirow{2}{*}{BASIC-H} & \multirow{2}{*}{Rank} \\
~ & ~ & P & R & F1 & P & R & F1 & P & R & F1 & ~ & ~ \\
\midrule
\parbox[t]{3mm}{\multirow{7}{*}{\rotatebox[origin=c]{90}{LLaVA-1.5-7B}}} & BrainDiffuser & 56.96 & 59.04 & 57.26 & 11.84 & 23.65 & 14.60 & 45.32 & 46.76 & 45.65 & 37.87 & 7 \\ 
~ & UMBRAE    & 62.51 & 62.94 & 61.65 & 15.04 & 30.50 & 18.70 & 52.68 & 52.55 & 52.10 & 42.56 & 6 \\ 
~ & NeuroPictor   & 63.79 & 61.94 & 61.71 & 16.26 & 31.12 & 19.88 & 53.40 & 52.70 & 52.59 & 43.15 & 5 \\ 
~ & MindEye   & 63.93 & 61.98 & 61.88 & 16.08 & 32.43 & 20.16 & 53.51 & 52.35 & 52.41 & 43.30 & 4 \\ 
~ & MindEye2  & 64.17 & 63.62 & 62.71 & 17.51 & 34.49 & 21.58 & 53.90 & 53.81 & 53.39 & 44.39 & 3 \\ 
~ & BrainGuard    & 65.26 & 64.84 & 64.04 & 18.36 & 35.71 & 22.72 & 55.85 & 55.22 & 55.00 & 45.70 & 2 \\ 
~ & DREAM & 68.76 & 65.62 & 65.54 & 18.52 & 36.81 & 23.06 & 60.02 & 59.18 & 59.01 & 47.24 & 1 \\ 
\midrule
\parbox[t]{3mm}{\multirow{7}{*}{\rotatebox[origin=c]{90}{LLaVA-1.5-13B}}} & BrainDiffuser & 57.34 & 59.53 & 57.66 & 11.66 & 23.69 & 14.26 & 45.45 & 47.18 & 45.90 & 37.95 & 7 \\ 
~ & UMBRAE    & 62.32 & 63.20 & 61.72 & 15.42 & 31.11 & 19.00 & 52.34 & 52.64 & 52.05 & 42.70 & 6 \\ 
~ & NeuroPictor   & 63.97 & 62.91 & 62.32 & 16.17 & 32.08 & 19.99 & 52.98 & 53.20 & 52.66 & 43.46 & 4 \\ 
~ & MindEye   & 63.46 & 61.90 & 61.59 & 16.51 & 33.25 & 20.49 & 53.40 & 52.06 & 52.29 & 43.29 & 5 \\ 
~ & MindEye2  & 63.78 & 64.63 & 63.12 & 17.30 & 33.97 & 21.24 & 53.45 & 54.03 & 53.32 & 44.41 & 3 \\ 
~ & BrainGuard    & 65.21 & 64.75 & 63.85 & 18.52 & 35.15 & 22.63 & 55.59 & 55.90 & 55.30 & 45.65 & 2 \\ 
~ & DREAM & 69.06 & 66.02 & 66.01 & 20.53 & 40.73 & 25.34 & 60.17 & 59.15 & 59.17 & 48.37 & 1 \\ 
\midrule
\parbox[t]{3mm}{\multirow{7}{*}{\rotatebox[origin=c]{90}{LLaVA-1.6-7B}}} & BrainDiffuser & 57.68 & 58.61 & 57.77 & 14.70 & 47.55 & 20.97 & 42.57 & 43.50 & 42.69 & 40.04 & 7 \\ 
~ & UMBRAE    & 61.12 & 62.48 & 61.37 & 18.17 & 53.93 & 25.61 & 47.47 & 48.96 & 47.84 & 44.36 & 6 \\ 
~ & NeuroPictor   & 62.31 & 62.21 & 61.68 & 18.38 & 53.63 & 25.43 & 48.62 & 49.49 & 48.67 & 44.58 & 5 \\ 
~ & MindEye   & 62.76 & 60.86 & 61.27 & 19.87 & 53.13 & 27.11 & 49.09 & 48.01 & 48.17 & 44.98 & 4  \\ 
~ & MindEye2  & 62.09 & 63.39 & 62.21 & 18.93 & 53.30 & 26.10 & 49.15 & 50.36 & 49.40 & 45.20 & 3\\ 
~ & BrainGuard    & 63.29 & 63.18 & 62.68 & 19.69 & 54.61 & 27.18 & 50.34 & 51.75 & 50.62 & 46.07 & 2\\ 
~ & DREAM & 64.97 & 64.27 & 63.94 & 19.52 & 52.98 & 26.94 & 52.73 & 54.52 & 53.18 & 46.99 & 1 \\ 
\midrule
\parbox[t]{3mm}{\multirow{7}{*}{\rotatebox[origin=c]{90}{LLaVA-1.6-13B}}} & BrainDiffuser & 57.86 & 59.09 & 58.07 & 13.31 & 45.83 & 19.37 & 43.20 & 44.41 & 43.49 & 39.68 & 7 \\ 
~ & UMBRAE    & 61.59 & 62.09 & 61.32 & 17.50 & 50.64 & 24.39 & 49.08 & 49.08 & 48.75 & 44.03 & 6 \\ 
~ & NeuroPictor   & 62.99 & 61.06 & 61.38 & 17.88 & 50.00 & 24.60 & 49.62 & 49.08 & 48.98 & 44.19 & 5 \\ 
~ & MindEye   & 62.95 & 60.65 & 61.27 & 18.16 & 51.19 & 25.09 & 49.98 & 48.42 & 48.84 & 44.31 & 4 \\ 
~ & MindEye2  & 62.57 & 62.13 & 61.73 & 17.87 & 50.14 & 24.73 & 49.72 & 49.17 & 49.07 & 44.39 & 3 \\ 
~ & BrainGuard    & 63.64 & 62.37 & 62.44 & 18.67 & 52.41 & 25.86 & 51.25 & 50.72 & 50.60 & 45.44 & 2 \\ 
~ & DREAM & 65.62 & 63.06 & 63.56 & 18.97 & 50.69 & 25.91 & 53.46 & 53.22 & 52.92 & 46.38 & 1 \\ 
\midrule
\parbox[t]{3mm}{\multirow{7}{*}{\rotatebox[origin=c]{90}{LLaVA-1.6-34B}}} & BrainDiffuser & 57.55 & 58.70 & 57.67 & 13.43 & 46.59 & 19.50 & 42.98 & 44.47 & 43.40 & 39.55 & 7 \\ 
~ & UMBRAE   & 62.75 & 60.57 & 61.12 & 17.45 & 50.59 & 24.30 & 49.21 & 48.12 & 48.31 & 43.83 & 6 \\ 
~ & NeuroPictor  & 61.28 & 61.98 & 61.16 & 17.39 & 51.77 & 24.37 & 48.50 & 48.82 & 48.34 & 43.88 & 5 \\ 
~ & MindEye   & 62.85 & 60.96 & 61.23 & 17.60 & 50.84 & 24.33 & 48.91 & 48.66 & 48.45 & 43.91 & 4 \\ 
~ & MindEye2  & 62.35 & 62.24 & 61.75 & 17.28 & 49.96 & 24.15 & 49.20 & 49.13 & 48.82 & 44.13 & 3 \\ 
~ & BrainGuard    & 63.29 & 62.54 & 62.32 & 18.49 & 52.60 & 25.66 & 50.62 & 50.27 & 50.08 & 45.21 & 2 \\ 
~ & DREAM & 65.30 & 63.13 & 63.48 & 18.82 & 50.78 & 25.67 & 52.48 & 52.53 & 52.12 & 46.09 & 1 \\ 
\midrule
\parbox[t]{3mm}{\multirow{7}{*}{\rotatebox[origin=c]{90}{GPT-4o}}} & BrainDiffuser & 57.23 & 58.78 & 57.55 & 13.22 & 45.50 & 19.06 & 44.12 & 44.41 & 43.99 & 39.44 &7 \\ 
~ & UMBRAE   & 62.04 & 60.49 & 60.68 & 16.72 & 48.84 & 23.26 & 50.39 & 47.78 & 48.75 & 43.32 & 6 \\ 
~ & NeuroPictor  & 60.76 & 61.79 & 60.74 & 16.94 & 49.13 & 23.60 & 49.64 & 48.67 & 48.87 & 43.51 & 4\\ 
~ & MindEye   & 62.27 & 60.82 & 60.87 & 16.83 & 48.50 & 23.29 & 50.20 & 48.57 & 49.04 & 43.47 & 5 \\ 
~ & MindEye2  & 61.78 & 61.95 & 61.31 & 16.49 & 47.68 & 22.97 & 50.08 & 48.84 & 49.14 & 43.54 & 3 \\ 
~ & BrainGuard    & 62.90 & 62.24 & 61.96 & 18.34 & 51.29 & 25.22 & 51.76 & 49.96 & 50.54 & 44.98 & 2 \\ 
~ & DREAM & 64.72 & 63.01 & 63.09 & 18.64 & 49.89 & 25.38 & 53.81 & 52.71 & 52.91 & 45.97 & 1 \\ 
\bottomrule
\end{tabular}
}
\end{table*}

\setlength{\tabcolsep}{4pt}
\setlength{\fboxrule}{0pt} 
\setlength{\fboxsep}{2pt}
\begin{table*}[t!]
\centering
\caption{Cross-model evaluation. Each method's reconstruction captions (\textit{recon}) are generated using one MLLM and evaluated against reference captions (\textit{stimuli}) produced by another different MLLMs. This setup assesses the impact of using various MLLMs as the caption producer and the evaluator.}
\label{tab:supmat_ab_cross_evaluation}
\vspace{-2.5pt}
{
\begin{tabular}{l|l|cccccccccc}
\toprule
Stimuli & Recon & BrainDiffuser & MindEye & MindEye2 & DREAM & UMBRAE & NeuroPictor & BrainGuard  \\
\midrule
1.5-7B  & 1.5-7B  & 37.87     & 42.56  & 43.15   & 43.30   & 44.39    & 45.70  & 47.24 \\
1.5-7B  & 1.5-13B & 37.82     & 42.70  & 42.90   & 42.93   & 44.41    & 45.59  & 47.42 \\
1.5-7B  & 1.6-7B  & 35.05     & 39.68  & 39.51   & 40.70   & 40.59    & 41.73  & 43.79 \\
1.5-7B  & 1.6-13B & 34.90     & 39.29  & 38.85   & 40.50   & 40.50    & 41.63  & 43.03 \\
1.5-13B & 1.5-7B  & 37.80     & 42.34  & 42.95   & 43.13   & 44.44    & 45.33  & 47.45 \\
1.5-13B & 1.5-13B & 37.95     & 42.70  & 43.46   & 43.29   & 44.41    & 45.65  & 48.37 \\
1.5-13B & 1.6-7B  & 35.43     & 39.94  & 39.88   & 41.05   & 41.28    & 42.14  & 43.71 \\
1.5-13B & 1.6-13B & 34.85     & 39.86  & 39.32   & 40.54   & 40.64    & 41.75  & 43.38 \\
1.6-7B  & 1.5-7B  & 32.65     & 36.02  & 36.11   & 36.23   & 36.28    & 37.50  & 38.49 \\
1.6-7B  & 1.5-13B & 32.81     & 35.92  & 36.14   & 36.16   & 36.46    & 37.80  & 38.94 \\
1.6-7B  & 1.6-7B  & 40.04     & 44.36  & 44.58   & 44.98   & 45.20    & 46.07  & 46.99 \\
1.6-7B  & 1.6-13B & 39.78     & 44.15  & 44.24   & 44.82   & 44.73    & 45.63  & 46.68 \\
1.6-13B & 1.5-7B  & 32.58     & 35.44  & 35.43   & 35.25   & 35.52    & 36.55  & 37.83 \\
1.6-13B & 1.5-13B & 32.69     & 35.40  & 35.48   & 35.26   & 35.86    & 36.76  & 38.21 \\
1.6-13B & 1.6-7B  & 39.08     & 43.44  & 43.40   & 43.51   & 43.50    & 44.39  & 45.48 \\
1.6-13B & 1.6-13B & 39.68     & 44.03  & 44.19   & 44.31   & 44.39    & 45.44  & 46.37 \\
\bottomrule
\end{tabular}
}
\end{table*}

\vspace{-2.5pt}
\setlength{\tabcolsep}{4pt}
\setlength{\fboxrule}{0pt} 
\setlength{\fboxsep}{2pt}
\begin{table*}[t!]
\centering
\caption{Cross-model evaluation results for BrainDiffuser.}
\label{tab:supmat_ab_cross_evaluation_braindiffuser}
\vspace{-2.5pt}
{
\begin{tabular}{c|ll|ccc|ccc|ccc|c}
\toprule
~ & \multicolumn{2}{c|}{Setup}  & \multicolumn{3}{c|}{Object}  & \multicolumn{3}{c|}{Attribute}  & \multicolumn{3}{c|}{Relation} & \multirow{2}{*}{BASIC-H} \\
~ & Stimuli & Reon & P & R & F1 & P & R & F1 & P & R & F1 & \\
\midrule
\parbox[t]{3mm}{\multirow{16}{*}{\rotatebox[origin=c]{90}{BrainDiffuser}}} & 1.5-7B & 1.5-7B & 56.96 & 59.04 & 57.26 & 11.84 & 23.65 & 14.60 & 45.32 & 46.75 & 45.65 & 37.87 \\
~ & 1.5-7B  & 1.5-13B & 57.14 & 58.77 & 57.20 & 11.86 & 23.88 & 14.65 & 45.08 & 46.54 & 45.40 & 37.82 \\
~ & 1.5-7B  & 1.6-7B  & 57.81 & 49.81 & 52.96 & 11.97 & 21.48 & 14.21 & 42.82 & 39.86 & 40.93 & 35.05 \\
~ & 1.5-7B  & 1.6-13B & 58.40 & 48.89 & 52.63 & 11.95 & 22.23 & 14.33 & 43.42 & 38.75 & 40.59 & 34.90 \\
~ & 1.5-13B & 1.5-7B  & 56.82 & 59.46 & 57.37 & 11.53 & 23.66 & 14.21 & 45.42 & 47.10 & 45.84 & 37.80 \\
~ & 1.5-13B & 1.5-13B & 57.34 & 59.53 & 57.66 & 11.66 & 23.69 & 14.26 & 45.45 & 47.18 & 45.90 & 37.95 \\
~ & 1.5-13B & 1.6-7B  & 57.97 & 50.36 & 53.41 & 12.25 & 22.27 & 14.56 & 42.93 & 40.27 & 41.21 & 35.43 \\
~ & 1.5-13B & 1.6-13B & 58.65 & 49.43 & 53.13 & 11.55 & 20.93 & 13.55 & 43.59 & 39.19 & 40.90 & 34.85 \\
~ & 1.6-7B  & 1.5-7B  & 48.88 & 59.95 & 53.24 & 5.15  & 25.94 & 8.17  & 38.26 & 43.67 & 40.41 & 32.65 \\
~ & 1.6-7B  & 1.5-13B & 49.23 & 60.02 & 53.41 & 5.30  & 26.95 & 8.40  & 38.33 & 43.63 & 40.43 & 32.81 \\
~ & 1.6-7B  & 1.6-7B  & 57.68 & 58.61 & 57.77 & 14.70 & 47.55 & 20.97 & 42.57 & 43.50 & 42.69 & 40.04 \\
~ & 1.6-7B  & 1.6-13B & 58.29 & 57.58 & 57.59 & 14.56 & 45.41 & 20.65 & 43.07 & 42.38 & 42.41 & 39.78 \\
~ & 1.6-13B & 1.5-7B  & 48.25 & 60.82 & 53.10 & 4.79  & 28.37 & 7.82  & 38.10 & 45.36 & 41.05 & 32.58 \\
~ & 1.6-13B & 1.5-13B & 48.49 & 60.99 & 53.29 & 4.91  & 27.94 & 7.94  & 38.07 & 45.19 & 40.96 & 32.69 \\
~ & 1.6-13B & 1.6-7B  & 56.44 & 59.24 & 57.39 & 12.91 & 47.22 & 18.89 & 41.84 & 44.54 & 42.83 & 39.08 \\
~ & 1.6-13B & 1.6-13B & 57.86 & 59.09 & 58.07 & 13.31 & 45.83 & 19.37 & 43.19 & 44.41 & 43.49 & 39.68 \\
\bottomrule
\end{tabular}
}
\end{table*}

\setlength{\tabcolsep}{4pt}
\setlength{\fboxrule}{0pt} 
\setlength{\fboxsep}{2pt}
\begin{table*}[t!]
\centering
\caption{Evaluation results under different box and text thresholds.}
\label{tab:supmat_ab_threshold}
\resizebox{0.75\textwidth}{!}{%
\begin{tabular}{c|cc|cc|cc|cc|cc|cc|c}
\toprule
~ & \multicolumn{2}{c|}{Threshold} & \multicolumn{2}{c|}{Salient} & \multicolumn{2}{c|}{Binary} & \multicolumn{2}{c|}{Semantic} & \multicolumn{2}{c|}{Instance} & \multicolumn{2}{c|}{Part} & \multirow{2}{*}{BASIC-L} \\
\cmidrule(r){2-13} %
~ & Box & Text & IoU  & AP  & IoU  & AP  & IoU  & AP  & IoU  & AP  & IoU  & AP  &  \\
\midrule
\parbox[t]{3mm}{\multirow{9}{*}{\rotatebox[origin=c]{90}{BrainDiffuser}}} & 0.25 & 0.30 & 17.96 & 20.21 & 38.98 & 45.85 & 18.66 & 20.78 & 20.09 & 1.94 & 7.86 & 7.82 & 16.65 \\
~ & 0.25 & 0.35 & 16.67 & 18.50 & 32.48 & 39.27 & 22.23 & 24.73 & 17.34 & 1.72 & 8.25 & 8.22 & 16.54 \\
~ & 0.25 & 0.40 & 14.60 & 16.10 & 25.64 & 31.26 & 25.76 & 28.55 & 15.99 & 1.54 & 8.76 & 8.67 & 16.57 \\
~ & 0.30 & 0.30 & 6.52 & 7.34 & 22.72 & 26.26 & 18.65 & 20.77 & 28.66 & 2.97 & 8.01 & 7.66 & 15.39 \\
~ & 0.30 & 0.35 & 5.96 & 6.65 & 19.36 & 22.85 & 22.24 & 24.74 & 23.83 & 2.52 & 8.76 & 8.45 & 15.06 \\
~ & 0.30 & 0.40 & 5.49 & 6.02 & 15.50 & 18.39 & 25.76 & 28.56 & 19.70 & 2.18 & 8.91 & 8.54 & 14.79 \\
~ & 0.35 & 0.30 & 6.26 & 6.97 & 20.14 & 23.87 & 22.18 & 24.64 & 32.71 & 3.73 & 8.48 & 8.06 & 17.30 \\
~ & 0.35 & 0.35 & 6.25 & 6.98 & 20.30 & 23.93 & 22.24 & 24.74 & 32.64 & 3.77 & 8.71 & 8.41 & 17.34 \\
~ & 0.35 & 0.40 & 5.75 & 6.31 & 16.29 & 19.33 & 25.76 & 28.56 & 27.50 & 2.99 & 8.91 & 8.54 & 16.82 \\
\midrule
\parbox[t]{3mm}{\multirow{9}{*}{\rotatebox[origin=c]{90}{MindEye}}} & 0.25 & 0.30 & 19.20 & 22.79 & 45.53 & 55.17 & 18.73 & 21.94 & 20.36 & 1.99 & 7.49 & 8.26 & 17.03 \\
~ & 0.25 & 0.35 & 18.27 & 21.49 & 39.75 & 49.61 & 21.64 & 25.08 & 18.01 & 1.90 & 7.47 & 8.14 & 16.89 \\
~ & 0.25 & 0.40 & 16.65 & 19.60 & 32.86 & 41.91 & 25.27 & 29.03 & 16.76 & 1.73 & 7.83 & 8.41 & 17.07 \\
~ & 0.30 & 0.30 & 7.55 & 8.76 & 26.09 & 30.97 & 18.72 & 21.94 & 28.12 & 3.04 & 7.74 & 8.14 & 15.53 \\
~ & 0.30 & 0.35 & 7.15 & 8.34 & 22.87 & 27.74 & 21.64 & 25.08 & 24.38 & 2.80 & 7.63 & 7.93 & 15.17 \\
~ & 0.30 & 0.40 & 6.43 & 7.40 & 19.00 & 23.39 & 25.27 & 29.03 & 20.63 & 2.45 & 8.58 & 8.65 & 15.12 \\
~ & 0.35 & 0.30 & 7.23 & 8.35 & 22.96 & 28.04 & 21.51 & 24.96 & 31.56 & 3.97 & 7.84 & 8.05 & 17.00 \\
~ & 0.35 & 0.35 & 7.36 & 8.58 & 23.61 & 28.59 & 21.65 & 25.09 & 31.75 & 4.12 & 7.78 & 8.08 & 17.11 \\
~ & 0.35 & 0.40 & 6.62 & 7.62 & 19.62 & 24.11 & 25.28 & 29.04 & 27.90 & 3.42 & 8.57 & 8.65 & 16.99 \\
\midrule
\parbox[t]{3mm}{\multirow{9}{*}{\rotatebox[origin=c]{90}{UMBRAE}}} & 0.25 & 0.30 & 21.33 & 24.62 & 40.96 & 48.53 & 18.94 & 21.16 & 20.29 & 2.15 & 8.43 & 8.47 & 17.89 \\
~ & 0.25 & 0.35 & 20.34 & 23.49 & 35.99 & 43.24 & 22.57 & 25.03 & 17.91 & 1.99 & 9.03 & 8.99 & 18.03 \\
~ & 0.25 & 0.40 & 18.00 & 20.78 & 29.67 & 36.01 & 25.84 & 28.72 & 16.70 & 1.80 & 8.87 & 8.79 & 17.81 \\
~ & 0.30 & 0.30 & 8.38  & 9.70  & 23.68 & 27.43 & 18.95 & 21.17 & 28.32 & 3.25 & 8.78 & 8.46 & 16.09 \\
~ & 0.30 & 0.35 & 7.95  & 9.24  & 21.38 & 24.91 & 22.58 & 25.05 & 24.37 & 2.88 & 9.43 & 9.07 & 16.01 \\
~ & 0.30 & 0.40 & 6.73  & 7.74  & 17.21 & 20.30 & 25.86 & 28.74 & 20.35 & 2.46 & 9.41 & 8.89 & 15.50 \\
~ & 0.35 & 0.30 & 8.03  & 9.32  & 21.70 & 25.34 & 22.47 & 24.93 & 32.58 & 4.30 & 9.67 & 9.31 & 18.11 \\
~ & 0.35 & 0.35 & 8.19  & 9.53  & 22.15 & 25.77 & 22.58 & 25.05 & 32.59 & 4.39 & 9.64 & 9.28 & 18.18 \\
~ & 0.35 & 0.40 & 6.95  & 7.99  & 17.78 & 20.95 & 25.85 & 28.74 & 28.31 & 3.56 & 9.41 & 8.89 & 17.50 \\
\midrule
\parbox[t]{3mm}{\multirow{9}{*}{\rotatebox[origin=c]{90}{DREAM}}} & 0.25 & 0.30 & 23.62 & 26.10 & 46.03 & 57.11 & 21.15 & 24.13 & 21.41 & 2.32 & 9.22  & 8.79  & 19.57 \\
~ & 0.25 & 0.35 & 22.74 & 24.72 & 40.18 & 50.40 & 24.92 & 27.69 & 18.88 & 2.23 & 10.43 & 9.88  & 19.86 \\
~ & 0.25 & 0.40 & 19.69 & 21.10 & 32.22 & 41.52 & 27.37 & 29.81 & 17.34 & 2.03 & 10.46 & 9.74  & 19.17 \\
~ & 0.30 & 0.30 & 8.83  & 9.95  & 24.72 & 30.18 & 21.14 & 24.12 & 29.79 & 3.52 & 9.80  & 9.10  & 17.34 \\
~ & 0.30 & 0.35 & 8.54  & 9.36  & 21.72 & 26.76 & 24.91 & 27.68 & 25.78 & 3.22 & 11.03 & 10.10 & 17.44 \\
~ & 0.30 & 0.40 & 7.60  & 8.15  & 17.70 & 21.94 & 27.39 & 29.83 & 21.51 & 2.80 & 10.83 & 9.77  & 16.67 \\
~ & 0.35 & 0.30 & 8.61  & 9.47  & 21.66 & 26.96 & 24.62 & 27.42 & 32.95 & 4.52 & 10.89 & 9.89  & 19.15 \\
~ & 0.35 & 0.35 & 8.72  & 9.56  & 22.23 & 27.36 & 24.91 & 27.68 & 33.22 & 4.70 & 11.01 & 10.08 & 19.35 \\
~ & 0.35 & 0.40 & 7.76  & 8.32  & 17.68 & 22.40 & 27.39 & 29.83 & 28.99 & 3.74 & 10.83 & 9.78  & 18.59 \\
\midrule
\parbox[t]{3mm}{\multirow{9}{*}{\rotatebox[origin=c]{90}{BrainGuard}}} & 0.25 & 0.30 & 25.90 & 28.07 & 49.22 & 58.99 & 23.34 & 25.21 & 23.98 & 3.29 & 10.82 & 10.32 & 21.77 \\
~ & 0.25 & 0.35 & 24.43 & 26.52 & 42.69 & 51.86 & 27.09 & 29.01 & 20.68 & 3.03 & 11.31 & 10.71 & 21.53 \\
~ & 0.25 & 0.40 & 22.12 & 23.86 & 34.67 & 42.80 & 30.62 & 32.75 & 19.04 & 2.75 & 11.87 & 11.06 & 21.42 \\
~ & 0.30 & 0.30 & 10.45 & 11.35 & 29.24 & 34.15 & 23.34 & 25.20 & 32.67 & 4.88 & 11.39 & 10.51 & 19.42 \\
~ & 0.30 & 0.35 & 9.71  & 10.57 & 25.45 & 29.86 & 27.07 & 28.99 & 28.03 & 4.34 & 13.11 & 11.96 & 19.31 \\
~ & 0.30 & 0.40 & 8.58  & 9.32  & 20.49 & 24.49 & 30.61 & 32.75 & 23.87 & 3.78 & 13.91 & 12.61 & 18.97 \\
~ & 0.35 & 0.30 & 9.93  & 10.80 & 26.03 & 30.81 & 26.95 & 28.84 & 36.23 & 6.21 & 12.97 & 11.84 & 21.37 \\
~ & 0.35 & 0.35 & 10.04 & 10.93 & 26.41 & 30.98 & 27.08 & 29.01 & 36.26 & 6.34 & 13.26 & 12.13 & 21.50 \\
~ & 0.35 & 0.40 & 8.88  & 9.64  & 21.12 & 25.24 & 30.61 & 32.75 & 31.72 & 5.12 & 14.14 & 12.89 & 21.07 \\
\midrule
\parbox[t]{3mm}{\multirow{9}{*}{\rotatebox[origin=c]{90}{MindEye2}}} & 0.25 & 0.30 & 25.29 & 26.27 & 47.93 & 57.52 & 24.33 & 25.68 & 24.09 & 3.45 & 12.32 & 11.03 & 22.16 \\
~ & 0.25 & 0.35 & 24.20 & 25.17 & 41.83 & 51.74 & 27.89 & 29.26 & 21.03 & 3.21 & 13.38 & 11.95 & 22.17 \\
~ & 0.25 & 0.40 & 22.27 & 22.90 & 35.64 & 44.47 & 31.47 & 32.84 & 19.50 & 3.06 & 14.26 & 12.64 & 22.27 \\
~ & 0.30 & 0.30 & 9.79 & 10.22 & 27.75 & 32.74 & 24.35 & 25.69 & 32.93 & 5.10 & 12.80 & 11.28 & 19.82 \\
~ & 0.30 & 0.35 & 9.32 & 9.73 & 24.53 & 29.56 & 27.91 & 29.27 & 28.51 & 4.53 & 14.09 & 12.51 & 19.72 \\
~ & 0.30 & 0.40 & 8.85 & 9.18 & 20.70 & 25.12 & 31.48 & 32.84 & 24.21 & 4.04 & 14.20 & 12.54 & 19.41 \\
~ & 0.35 & 0.30 & 9.49 & 9.85 & 24.73 & 29.79 & 27.65 & 28.98 & 36.58 & 6.45 & 14.09 & 12.54 & 21.72 \\
~ & 0.35 & 0.35 & 9.51 & 9.92 & 25.05 & 30.14 & 27.90 & 29.26 & 36.79 & 6.56 & 14.19 & 12.65 & 21.86 \\
~ & 0.35 & 0.40 & 9.00 & 9.34 & 21.13 & 25.59 & 31.48 & 32.84 & 32.51 & 5.43 & 14.18 & 12.52 & 21.53 \\
\midrule
\parbox[t]{3mm}{\multirow{9}{*}{\rotatebox[origin=c]{90}{NeuroPictor}}} & 0.25 & 0.30 & 
{29.45} & 31.29 & 47.79 & 56.48 & 27.97 & 29.57 & 26.47 & 4.08 & 17.17 & 15.84 & 25.88 \\
~ & 0.25 & 0.35 & 27.64 & 28.96 & 41.30 & 49.20 & 32.34 & 33.68 & 23.04 & 3.79 & 18.36 & 16.67 & 25.81 \\
~ & 0.25 & 0.40 & 24.45 & 25.65 & 33.91 & 41.12 & 37.05 & 38.35 & 21.30 & 3.59 & 21.01 & 19.13 & 26.13 \\
~ & 0.30 & 0.30 & 11.65 & 12.26 & 27.75 & 31.91 & 27.97 & 29.57 & 36.13 & 6.03 & 18.06 & 16.17 & 23.13 \\
~ & 0.30 & 0.35 & 10.84 & 11.42 & 24.56 & 28.30 & 32.34 & 33.69 & 31.31 & 5.31 & 19.79 & 17.43 & 23.12 \\
~ & 0.30 & 0.40 & 9.47 & 9.97 & 20.04 & 23.37 & 37.06 & 38.37 & 26.68 & 4.78 & 21.61 & 19.30 & 23.10 \\
~ & 0.35 & 0.30 & 11.22 & 11.78 & 25.29 & 29.16 & 32.37 & 33.76 & 40.78 & 7.59 & 19.85 & 17.52 & 25.62 \\
~ & 0.35 & 0.35 & 11.17 & 11.77 & 25.44 & 29.24 & 32.34 & 33.68 & 40.75 & 7.64 & 19.90 & 17.54 & 25.61 \\
~ & 0.35 & 0.40 & 9.77 & 10.28 & 20.81 & 24.20 & 37.06 & 38.36 & 35.88 & 6.29 & 21.72 & 19.40 & 25.51 \\
\bottomrule
\end{tabular}
}
\end{table*}

\vspace{-2.5pt}
\section{Limitation and Broader Impact}
\label{sec:supmat_discussion}

\paragraph{Limitations.} Despite our efforts in developing a versatile and robust evaluation pipeline, there are still challenges in capturing nuances in brain visual decoding, such as exploring additional cognitive dimensions like emotional or attentional states, to enable a more holistic evaluation. This limitation stems from current datasets lacking sufficient diversity in cognitive states or subject demographics. Furthermore, our method targets recent decoding techniques that can generate clear and semantically relevant images. The aspect of image quality is therefore not considered a key criterion. This makes the framework less suitable for assessing early-stage visual coding work that often contains artifacts, blurriness, or difficult-to-recognize semantics. The reliance on large-scale MLLMs
also presents challenges in terms of computational cost and susceptibility to hallucination.

\paragraph{Broader Impacts.} 

This presented framework provides a more detailed, interpretable, and standardized method for assessing visual decoding models. These enhanced evaluation capabilities facilitates a deeper understanding of how models interpret and reconstruct visual information, which is crucial for advancing brain-inspired models, brain-computer interfaces, and assistive technologies. 
To address privacy and security concerns, particularly those surrounding neurodata and potential misuse, it is essential to rigorously assess ethical implications
and implement 
safeguards for participant data before applying the evaluation.

\fi

\end{document}